\documentclass[11pt]{article}

\usepackage[final]{acl}
\usepackage{microtype}
\usepackage{graphicx}
\usepackage{subcaption}
\usepackage{tabularx}

\usepackage{booktabs} 
\usepackage{booktabs}
\usepackage{multirow}

\usepackage{pifont}
\usepackage{pgfplots}
\usepackage{tikz}
\pgfplotsset{compat=1.18}
\usepgfplotslibrary{groupplots}
\usepackage{booktabs}
\usepackage{makecell}   % for 2-line cells
\usepackage{graphicx}   % for \resizebox (optional)
\usepackage{enumitem}
\definecolor{tableheader}{RGB}{230,240,255}
\definecolor{tablerow}{RGB}{245,248,255}
\usepackage{pgfplotstable}
\usepackage{placeins}    % for \FloatBarrier (optional)
\definecolor{linkblue}{RGB}{0,0,255}
\usepackage{caption}
\usepackage{tikz}
\usepackage{subcaption}
\usepackage{booktabs}
\usepackage{makecell}
\usepackage[table]{xcolor} % only if you keep row colors elsewhere
\definecolor{gain}{RGB}{0,120,0}
\definecolor{loss}{RGB}{180,0,0}
\definecolor{headergray}{RGB}{235,235,235}
\definecolor{sectiongray}{RGB}{245,245,245}
\definecolor{stagegray}{RGB}{240,240,240}
\definecolor{datablue}{RGB}{20,90,160}
\definecolor{trainpurple}{RGB}{120,40,140}
\definecolor{resultgreen}{RGB}{0,120,60}
\definecolor{datablue}{RGB}{20,90,160}
\definecolor{trainpurple}{RGB}{120,40,140}
\definecolor{resultgreen}{RGB}{0,120,60}

\usepackage{graphicx}
\usepackage{subcaption}
\usepackage{algorithm}
\usepackage{algpseudocode} % from algorithmicx
\usepackage{colortbl}
\usepackage{xcolor}
\usepackage{amsmath}
\usepackage{amssymb}
\usepackage{mathtools}
\usepackage{amsthm}
\usepackage{xcolor}
\usepackage{booktabs}
\usepackage{makecell}
\definecolor{gain}{RGB}{0,120,0}
\definecolor{headergray}{RGB}{235,235,235}
\definecolor{gain}{RGB}{0,120,0}
\usepackage[capitalize,noabbrev]{cleveref}
\theoremstyle{plain}

\theoremstyle{definition}

\theoremstyle{remark}

\usepackage{times}
\usepackage{latexsym}
\usepackage[textsize=tiny]{todonotes}
\usepackage[most]{tcolorbox}
\usepackage{xcolor}

\usepackage[T1]{fontenc}
\PassOptionsToPackage{breaklinks,colorlinks=true,linkcolor=blue}{hyperref}
\usepackage[utf8]{inputenc}

\usepackage{microtype}

\usepackage{inconsolata}

\usepackage{graphicx}
\usepackage[small,compact]{titlesec}
\usepackage{flafter}
\newcommand{\squishlist}{
   \begin{list}{$\bullet$}
      { \setlength{\itemsep}{0pt} 
        \setlength{\parsep}{0pt} 
        \setlength{\topsep}{0pt} 
        \setlength{\partopsep}{0pt} 
        \setlength{\leftmargin}{1.5em} 
        \setlength{\labelwidth}{1em} 
        \setlength{\labelsep}{0.5em} } 
}

\title{\textbf{LiFT:} How to Enable In-Context Learning for Longitudinal Modelling}

\author{
Iqra Ali\textsuperscript{1},
Talia Tseriotou\textsuperscript{1},
Mahmud Elahi Akhter\textsuperscript{1}, \\
\textbf{Yuxiang Zhou}\textsuperscript{1},
\textbf{Maria Liakata}\textsuperscript{1,2} \\
\textsuperscript{1}Queen Mary University of London (UK) \\
\textsuperscript{2}The Alan Turing Institute (UK) \\
\texttt{\{iqra.ali,t.tseriotou,m.liakata\}@qmul.ac.uk}
}

\begin{document}
\maketitle
\begin{abstract}
Longitudinal NLP tasks such as mental health monitoring and stance evolution require modeling temporally ordered text to track persistence and detect change. Such tasks also suffer from data scarcity, often involving rare events and sparsely annotated data. Large Language Models (LLMs) can learn from small amounts of data through in-context learning (ICL) which is particularly important in low data resource scenarios. However, when it comes to tracking evolving interactions or identifying rare events or changes, LLMs fall short in both zero- and few-shot ICL. To address this limitation, we introduce LiFT, a model-agnostic longitudinal Instruction Fine-Tuning framework that is able to induce ICL behaviour in LLMs for longitudinal tasks. LiFT unifies diverse tasks through a shared instruction schema that leverages sequential dependencies, curriculum learning, and temporal conditioning. We evaluate LiFT on five longitudinal datasets under zero-, one-, and three-shot settings demonstrating that LiFT delivers superior performance to IFT models under identical ICL conditions.
Furthermore, we demonstrate that LiFT yields larger gains for minority classes, meaningfully attends to recent demonstration history and develops transferable longitudinal modeling capabilities beyond its training tasks\footnote{Our code will be released upon acceptance.}.
\end{abstract}
% \begin{figure}[t]
%     \centering
%     \includegraphics[width=\columnwidth]{plot4.pdf}
% \caption{Base and LiFT Macro-F1 across datasets and 0-1-3-shot settings. Light segments denote Base; dark segments denote LiFT. See \hyperref[fig:plot1]{\underline{\textcolor{blue}{Fig.~3}}}, and \hyperref[sec:fig]{\underline{\textcolor{blue}{Sec.~G}}} for details.}
% \label{fig:plot4}\end{figure}

\section{Introduction}
Many real-world applications require understanding how language, behaviour, and latent user states evolve over time. Such problems fall under \emph{longitudinal modeling} \cite{wang2025posts,ganesan2026word}, where observations are analyzed as temporally ordered sequences rather than independent instances \cite{bamler2017dynamic,rudolph2018dynamic,tseriotou-etal-2025-overview,ali-etal-2026-overview}. Unlike static NLP tasks that treat inputs independently, longitudinal tasks require modeling persistence, temporal dependencies, and state transitions across sequences of observations \cite{wang2019long,cui2025timer,kruse-etal-2025-large}. Longitudinal modeling arises in many NLP settings, including stance change detection \cite{kochkina2018allinonemultitasklearningrumour,kumar-carley-2019-tree}, mental health monitoring \cite{tsakalidis2022identifying, tsakalidis-etal-2022-overview,tseriotou-etal-2023-sequential,tseriotou2024tempoformer, hills2023creation, hills2024exciting}, multi-session dialogue reasoning \cite{liu-etal-2017-using-context,he2021speakerturnmodelingdialogue,havrilla2024teachinglargelanguagemodels,ge2025tremu}, timeline summarization \cite{song2025temporal,wu2025unfolding}, longitudinal clinical prediction and summarization \cite{cui2025timer,kruse-etal-2025-large,keerthana2025denselongitudinalprogressnote,cheng2025leveraging,chen2025reflections}, and behavioral forecasting from experiential timelines \cite{zheng2025promind,hayat2025context,hayat2025three}. Recent datasets and models increasingly focus on capturing temporal dependencies and evolving user behavior \cite{tsakalidis2022identifying, yang2023textittimetextitgraphrelativetimepretraining,xu2024mental,wang2025effectivetimeawarelanguagerepresentation,islakoglu2025chronosense}. 

Despite strong general capabilities, LLMs remain unreliable on temporal \cite{tan2023towards,fatemi2024test,chu2024timebench} and context-based user-level tasks \citep{Xuhai2024mentalllm} under zero- and few-shot prompting. Such limitations are particularly
pronounced in high-stakes domains longitudinal tasks but even on less complex non-longitudinal affective
computing tasks where fine-tuned
smaller models outperform LLMs \cite{amin2023will, yang-etal-2023-towards, Xuhai2024mentalllm}. Specifically, work on longitudinal classification shows that smaller fine-tuned models
outperform LLMs on both short \citep{wenzel2024temporal} and long \citep{tseriotou2024tempoformer} timelines, with specialised fine-tuning remaining superior and prompting relying on domain-specific prompt engineering \cite{zhu2024prompting}. Temporal benchmarks similarly reveal LLMs consistently underperfom compared to specialised models and humans, with ICL and chain-of-thought offering only partial gains \cite{qiu2024large}. Prompting alone does not resolve these limitations: chain-of-thought (CoT) and program-aided prompting yield inconsistent temporal-QA gains \cite{li2023unlocking}, excessive context can harm interval prediction \cite{cao2026more}, and demonstrations may convey format or label-space cues rather than the intended mapping \cite{workrethinking,yan2024understanding}. These challenges intensify in longitudinal tasks, where sparse evidence is distributed across ordered histories and long-context models remain vulnerable to distraction, recency bias, and position effects \cite{liu2023lost,xu2024stress}. Effective prediction therefore requires integrating evidence across observations and tracking evolving trajectories \cite{wang2019long,tseriotou-etal-2023-sequential,cui2025timer}. This motivates explicit longitudinal modeling that goes beyond generic ICL. Moreover,  modeling of user-generated longitudinal data should not compromise privacy preservation. %is usually privacy se such adaptation should hinder information sharing and user de-identification. %Privacy-sensitive longitudinal datasets, particularly in mental health, may contain sensitive and potentially identifying user information. 
We therefore opt for open-weight models that can be trained and evaluated locally. %This is also necessary for LiFT, which requires parameter access for LoRA, temporal--label conditioning, auxiliary objectives, and mechanistic analyses. 
In this light, we introduce \textbf{\textit{LiFT}}, a framework for adapting LLMs to temporally structured tasks. Across the evaluated settings, LiFT outperforms the corresponding instruction fine-tuned (IFT) baselines and generalises to unseen longitudinal datasets and task formulations. We contribute:
\begin{itemize}[leftmargin=*,nosep,labelsep=1em]
\item \textbf{Framework:} We introduce LiFT, combining curriculum-based few-shot training, temporal--label conditioning, and history-aware objectives to teach models to use temporally distributed evidence, enabling stronger ICL that transfers to unseen longitudinal datasets and tasks (\hyperref[sec:method]{\underline{\textcolor{blue}{Sec.~\ref*{sec:method}}}}).
\item \textbf{Comprehensive experiments:} Across OLMo (1B/7B), LLaMA-8B, and Qwen (14B/32B), LiFT enables superior zero- and few-shot longitudinal prediction, whereas IFT models benefit non-monotonically from increased demonstrations. Importantly, LiFT yields larger gains for minority classes and is shown to generalise to domains and task formulations distinct from training, demonstrating its capacity to learn transferable longitudinal patterns (\hyperref[sec:results]{\underline{\textcolor{blue}{Sec.~\ref*{sec:results}}}}).
\item \textbf{Mechanistic evidence and statistical reliability:} Probing and activation analyses show LiFT increases reliance on temporally structured historical context. Tests across seeds, datasets, and few-shot settings confirm consistence and statistical significance of performance gains (\hyperref[sec:mechanistic]{\underline{\textcolor{blue}{Sec.~\ref*{sec:mechanistic}}}}).
%that gains are statistically significant and directionally consistent.
%\item \textbf{OOD generalisation and minority-class performance:} LiFT generalises on two datasets excluded from training with distinct domains and task formulations, demonstrating its capacity to learn transferable longitudinal patterns. % of longitudinal change rather than dataset-specific decision boundaries. 
\end{itemize}

\begin{figure*}[t]
    \centering
    \includegraphics[
        width=0.98\textwidth,
        trim=8 10 10 7,
        clip
    ]{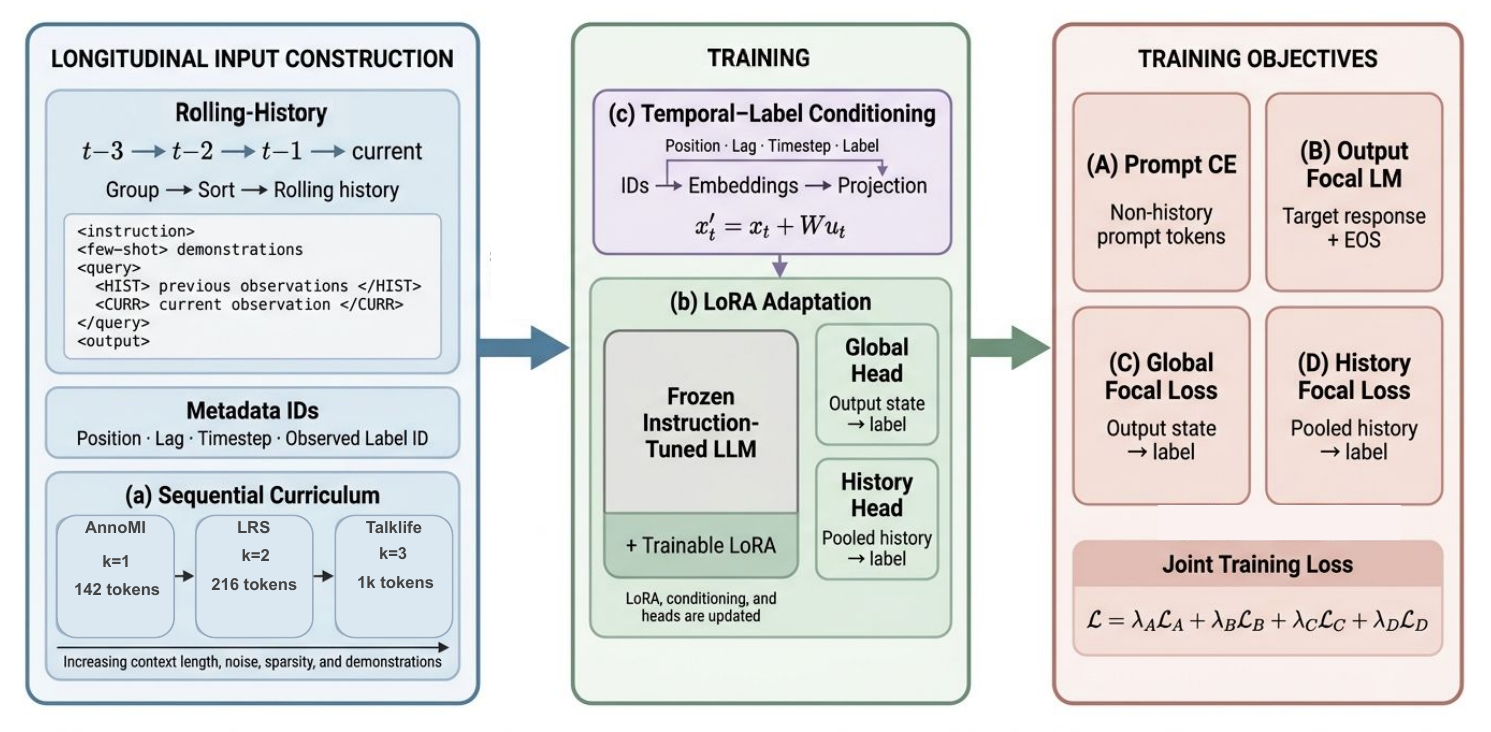}
\caption{\textit{Overview of LiFT:}
The longitudinal input construction stage (blue) groups observations by sequence, orders them chronologically, forms rolling histories, and converts them into structure  with temporal and label metadata IDs. Training (green) follows a  curriculum of increasing context length and difficulty: AnnoMI with (k{=}1) and a mean context token length of 142, LRS with (k{=}2) and 216, and TalkLife with (k{=}3) and (1.0\text{K}). The metadata IDs are embedded, projected, and added to the token representations through temporal--label conditioning. The instruction-tuned LLM is adapted using LoRA, together with a global classification head and a history-only auxiliary head. The joint objective (red) combines (A) prompt cross-entropy, (B) focal language-model loss, (C) global focal loss and (D) history-only focal loss.}
\label{fig:icl_ift}
\end{figure*}

\section{Related Work}
\noindent \textbf{LLMs for Longitudinal Tasks}
%LLMs have been applied to mental-health timelines, longitudinal Electronic Health Records (EHRs), event streams, and multi-session dialogue. 
have been applied in healthcare for trajectory summarisation, prediction, and temporal reasoning over multi-visit Electronic Health Records (EHRs) \cite{ li2025clicare,kruse-etal-2025-large,keerthana2025denselongitudinalprogressnote,almannaa2025investigating,fang2025betterehrreasoningllms}. LLMs struggle with long temporal progression \cite{kruse-etal-2025-large}, while \citet{moghaddam-etal-2026-reprompt} preserve longitudinal dependencies by recurrent integration of prior-visit states. %recurrently integrates prior-visit representations with structured EHR encodings). 
Towards longitudinal mental health monitoring, LLMs have been used %in aggregate evidence across posts, 
to summarise timelines, predict conditions, and model evolving psychological states \cite{song2024combininghierachicalvaesllms,lan-etal-2025-depression,song2025temporal,zheng2025promind,hayat2025context}. Other works have focused on iterative timeline summarisation \cite{hu-etal-2024-moments,wu2025unfolding}. %and long-term multi-session conversations \cite{maharana-etal-2024-evaluating,ge2025tremu}
Nevertheless, existing methods are largely tailored to individual domains or tasks \cite{kruse-etal-2025-large,wang-etal-2025-posts,hu-etal-2024-moments,maharana-etal-2024-evaluating}, often serialise records into a single input, leaving temporal dependencies and state changes to be inferred implicitly. % \cite{maharana-etal-2024-evaluating,kruse-etal-2025-large}.
%LiFT addresses this gap by explicitly training LLMs to integrate temporally distributed evidence and generalise to unseen longitudinal tasks.
\\
\noindent \textbf{Longitudinal Classification  Modeling}  represents observations as ordered trajectories and uses sequence-aware architectures to capture temporal dependencies \cite{wang2019long, tseriotou-etal-2023-sequential, hills2023creation}. Such methods have been applied to rumour detection \cite{kochkina2018allinonemultitasklearningrumour,kumar-carley-2019-tree}, dialogue classification \cite{liu-etal-2017-using-context}, and clinical outcome prediction from longitudinal records \cite{kruse-etal-2025-large,cui2025timer}. In mental health, longitudinal timeline classification is employed to detect behavioral or mood changes in users \cite{tsakalidis2022identifying,tseriotou-etal-2025-overview,wang2025posts}. Leveraging LLMs still remains challenging, with prompting-based approaches concatenating longitudinal context \cite{kruse-etal-2025-large,lan-etal-2025-depression,wang-etal-2025-posts} and under-performing smaller fine-tuned models \cite{amin2023will,xu2024mental,ali-etal-2026-overview, hills-etal-2026-multistream}. This motivates longitudinal model adaptation to improve few-shot LLM prompting. LiFT addresses this gap by explicitly training LLMs to integrate temporally distributed evidence, generalising in unseen longitudinal tasks. %Prompting-based approaches often serialize longitudinal observations into a single context, leaving temporal dependencies to be inferred implicitly \cite{kruse-etal-2025-large,lan-etal-2025-depression,wang-etal-2025-posts,maharana-etal-2024-evaluating}. Recent work shows that such methods under-perform smaller supervised models on longitudinal classification tasks \cite{hosokawa2024temporal,wenzel2024temporal,yang2023towards,xu2024mental}. This motivates longitudinal model adaptation for few-shot prompting.

\noindent \textbf{Instruction Fine-Tuning (IFT) Strategies} adapt LLMs to follow natural-language instructions using instruction--response datasets \cite{wei2022finetunedlanguagemodelszeroshot,ouyang2022traininglanguagemodelsfollow,lu2024emergent}. Frameworks such as FLAN and FLAN-T5 demonstrate that training on diverse instruction-based tasks improves zero-shot and cross-task generalisation \cite{wei2021finetuned,chung2024scaling}. Instruction datasets such as Super-NaturalInstructions and Self-Instruct further expand task diversity through human-authored and synthetically generated instructions \cite{wang2022super,wang2023self}. More recent work has explored competence-aware curricula that dynamically adapt data selection and training schedules to models' evolving capabilities \cite{li-etal-2025-teaching}. 
% Instruction-data selection has also been extended to multi-turn settings, where complete dialogue trajectories are evaluated rather than treating individual turns as independent training instances \cite{li-etal-2026-data}.
Parameter-efficient methods such as LoRA and QLoRA enable instruction tuning while updating only a small proportion of model parameters \cite{hu2021loralowrankadaptationlarge,dettmers2023qloraefficientfinetuningquantized}. LiFT %builds on these developments by combining 
combines curriculum-based instruction tuning, parameter-efficient adaptation, temporal--label conditioning, and history-aware objectives to support transferable longitudinal ICL.% on unseen tasks.

\section{Methodology}
\label{sec:method}
% \vspace{-0.8mm}
We introduce \textbf{LiFT} (Longitudinal Instruction Fine-Tuning), a framework for improving in-context learning of LLMs over temporally ordered sequences. As illustrated in
\hyperref[fig:icl_ift]{\underline{\textcolor{blue}{Fig.~\ref*{fig:icl_ift}}}},
LiFT combines:
(i) stage-wise curriculum training over builder-generated $k$-shot prompts
(\hyperref[fig:icl_ift]{\underline{\textcolor{blue}{Fig.~\ref*{fig:icl_ift}(a)}}}),
(ii) adaptation through LoRA
(\hyperref[fig:icl_ift]{\underline{\textcolor{blue}{Fig.~\ref*{fig:icl_ift}(b)}}}),
and
(iii) temporal and observed-label conditioning
(\hyperref[fig:icl_ift]{\underline{\textcolor{blue}{Fig.~\ref*{fig:icl_ift}(c)}}}).
The Trainer jointly optimizes four objectives:
(A) next-token cross-entropy over prompt-prefix tokens, excluding history spans
(\hyperref[fig:icl_ift]{\underline{\textcolor{blue}{Fig.~\ref*{fig:icl_ift}(A)}}});
(B) focal language-modeling loss over the target-output region
(\hyperref[fig:icl_ift]{\underline{\textcolor{blue}{Fig.~\ref*{fig:icl_ift}(B)}}});
(C) focal classification loss over a shared global label space
(\hyperref[fig:icl_ift]{\underline{\textcolor{blue}{Fig.~\ref*{fig:icl_ift}(C)}}});
and
(D) auxiliary history-only classification loss for subsequent prediction
(\hyperref[fig:icl_ift]{\underline{\textcolor{blue}{Fig.~\ref*{fig:icl_ift}(D)}}}). The remainder of this section presents the problem formulation
(\hyperref[sec:problem]{\underline{\textcolor{blue}{Sec.~\ref*{sec:problem}}}}),
the rolling-history formulation
(\hyperref[sec:seq_prompt]{\underline{\textcolor{blue}{Sec.~\ref*{sec:seq_prompt}}}}),
and curriculum-based instruction tuning
(\hyperref[sec:curriculum_ift]{\underline{\textcolor{blue}{Sec.~\ref*{sec:curriculum_ift}}}}).
It then describes model training
(\hyperref[sec:Longitudinal_Model_Training]{\underline{\textcolor{blue}{Sec.~\ref*{sec:Longitudinal_Model_Training}}}}), model adaptation
(\hyperref[sec:model_adaptation]{\underline{\textcolor{blue}{Sec.~\ref*{sec:model_adaptation}}}})
and temporal--label conditioning
(\hyperref[sec:rte]{\underline{\textcolor{blue}{Sec.~\ref*{sec:rte}}}}),
followed by the training objective
(\hyperref[sec:training]{\underline{\textcolor{blue}{Sec.~\ref*{sec:training}}}}).
\hyperref[alg:lift_builder]{\underline{\textcolor{blue}{Algorithm~\ref*{alg:lift_builder}}}}
and
\hyperref[alg:lift_trainer]{\underline{\textcolor{blue}{Algorithm~\ref*{alg:lift_trainer}}}}
detail the longitudinal input-construction and training workflows, respectively.

\subsection{Longitudinal Input Construction}
\label{sec:problem}

Let the history of a timeline be
\begin{equation}
\footnotesize
H_m=\bigl((x_1,t_1),\ldots,(x_m,t_m)\bigr),
\label{eq:problem_formulation1}
\end{equation}
where $x_i$ is the $i$-th observation and $t_i$ its timestamp, with
$t_1\leq\cdots\leq t_m\leq t_{m+1}$.
Given the current observation--timestamp pair $(x_{m+1},t_{m+1})$, LiFT conditions on the preceding history $H_m$ to produce a textual label $\hat{z}$ and a global class prediction $\hat{y}$
\begin{equation}
\footnotesize
\begin{aligned}
\hat{z} &= f\left(H_m,(x_{m+1},t_{m+1})\right),\\
\hat{y} &= g\left(H_m,(x_{m+1},t_{m+1})\right).\end{aligned}\label{eq}\end{equation}
where $\hat{z}$ is the textually generated label and $\hat{y}$ is the
prediction from the global classification head. Both represent the same
classification target. LiFT has not yet been evaluated for future-observation generation or longitudinal sequence forecasting.

\subsubsection{Rolling-History Sequential Structure}
\label{sec:seq_prompt}

For each dataset, observations are grouped using the relevant sequence
key, such as a timeline ID, and are then
sorted chronologically. For a current observation $(x_j,t_j)$, its
associated history is
\begin{equation}
\footnotesize
H_j =
\bigl(
(x_1,t_1),\ldots,(x_{j-1},t_{j-1})
\bigr),
\label{eq:rolling_history}
\end{equation}
which contains only historical (and not future) observations. The history and current observations are separated, preceded by \texttt{<hist>} and \texttt{<curr>} respectively:
\begin{equation}
\footnotesize
X_j =
\texttt{<hist>}H_j\texttt{</hist>}
\oplus
\texttt{<curr>}(x_j,t_j)\texttt{</curr>}.
\label{eq:query_structure}
\end{equation}
Given a fixed token budget with a portion reserved for output, the oldest history observations in $H_j$ are dropped iteratively until the prompt fits within the budget, while the instruction and current observation are always retained. Historical observations are prefixed with relative indices (\texttt{t-1}, \texttt{t-2}, \ldots).

%Let $T$ denote the tokenizer, $B$ the context budget, $B_{\mathrm{out}}$ the reserved output budget, and $\oplus$ sequence concatenation. After inserting $X_j$ into $Q_k$ in\hyperref[eq:prompt_template]{\underline{\textcolor{blue}{Eq.~\ref*{eq:prompt_template}}}}, the Builder enforces $\operatorname{TokLen}_{T}(Q_k) + B_{\mathrm{out}} \leq B$. If the prompt exceeds the budget, the oldest query-history observations are removed iteratively, while the instruction, current observation, output prefix, and $B_{\mathrm{out}}$-token allowance are retained. Historical observations are prefixed with relative indices such as \texttt{t-3}, \texttt{t-2}, and \texttt{t-1}.

\subsubsection{Curriculum-Based Few-Shot Instruction Tuning}
\label{sec:curriculum_ift}

Each training instance is represented using a common instruction schema:
\begin{equation}
\footnotesize
\begin{aligned}
Q_k={}&
\texttt{<instruction>}I\texttt{</instruction>}\\[-1pt]
&\oplus
\texttt{<few-shot>}\mathcal{D}_k
\texttt{</few-shot>}\\[-1pt]
&\oplus
\texttt{<query>}X_j\texttt{</query>}\\[-1pt]
&\oplus
\texttt{<output>}.
\end{aligned}
\label{eq:prompt_template}
\end{equation}
Here, $I$ is the natural-language task instruction,
$\mathcal{D}_k$ is the set of $k$ labelled in-context demonstrations,
and $X_j$ is the query as defined in \hyperref[tab:query_structure]{\underline{\textcolor{blue}{Eq.~\ref*{eq:query_structure}}}}. The target response $Y$ is not included in $Q_k$ and is
appended only during supervised training. %, as defined in \hyperref[eq:training_input]{\underline{\textcolor{blue}{Eq.~\ref*{eq:training_input}}}}. 
As illustrated in
\hyperref[fig:icl_ift]{\underline{\textcolor{blue}{Fig.~\ref*{fig:icl_ift}(a)}}},
training follows a three-stage curriculum ordered by increasing
contextual complexity (k timelines) and event sparsity
(see
\hyperref[tab:dataset_stats]{\underline{\textcolor{blue}{Table~\ref*{tab:dataset_stats}}}}
for the corresponding dataset statistics).
Stage~1 uses staged structured dialogue with relatively short turns (AnnoMI: motivational interviews; $k=1$; average context window (a.c.w.)=142 tokens) for stable initial supervision \cite{wu2022anno}. %AnnoMI with $k=1$ and an average context window of 142 tokens. Its structured motivational-interviewing
%dialogues provide stable initial
%supervision. 
Stage~2 introduces longer and noisier threads and evolving labels based on dynamic sequential context (LRS: evolving stance over social-media threads; $k=2$; a.c.w.=216 tokens) \cite{tseriotou-etal-2024-sig}. %LRS with $k=2$ and an average context window
%of 216 tokens, introducing noisier social-media language
%and evolving stance relationships. 
Stage~3 advances to even longer timelines containing sparse and implicit changes (TalkLife: mood changes over social media timelines; $k=3$; a.c.w.=1K tokens) \cite{tsakalidis-etal-2022-overview}.
%and contexts of 1.0K tokens, requiring the model to detect
%sparse and often implicit changes across longer mental-health timelines.
%Stage~3 uses TalkLife with $k=3$
%and contexts of 1.0K tokens, requiring the model to detect
%sparse and often implicit changes across longer mental-health timelines.
This progression increases simultaneously the difficulty of temporal modelling and
the context window size/number of examples.

\subsection{Longitudinal Model Training}
\label{sec:Longitudinal_Model_Training}
\subsubsection{Tokenisation and LoRA Adaptation}\label{sec:model_adaptation}

LiFT is built on top of instruction-tuned, open-weight decoder-only LLMs. The tokenizer vocabulary is extended with paired structural markers for
\texttt{instruction}, \texttt{few-shot}, \texttt{query}, \texttt{output},
\texttt{hist}, and \texttt{curr} regions.
These markers identify the functional regions of the prompt and allow
the same formatting to be used during training. %As shown in\hyperref[fig:icl_ift]{\underline{\textcolor{blue}{Fig.~\ref*{fig:icl_ift}(b)}}} 
The underlying language-model parameters are frozen, and LoRA modules are inserted into the attention and multilayer-perceptron projection layers.
Training is sequential across the curriculum stages, with each
stage continuing from the adapter parameters learned in the preceding
one.%The LoRA rank, learning-rate, optimization steps, demonstration format, and all other settings are held constant across baselines and ablations.

\subsubsection{Temporal--Label Conditioning}
\label{sec:rte}

The temporal--label conditioning module enriches each token embedding
with four signals (\hyperref[tab:query_structure]{\underline{\textcolor{blue}{Eq.~\ref*{eq:temporal_embeddings}}}}): absolute prompt position $p_t$, observation lag $r_t$,
local timestep $\tau_t$, and global label identifier $z_t$\footnote{Thus $r_t$ encodes distance from the current observation while $\tau_t$ encodes absolute timeline position.}. %For a token in historical observation $x_i$ relative to current observation $x_j$, $r_t=j-i$ and $\tau_t=i$; current-observation tokens use $r_t=0$ and $\tau_t=j$.}. 
%Each post within demonstrations has independent local coordinates, while other prompt elements have null temporal identifiers. 
Let $z_{\varnothing}$ denote the null-label identifier: only visible few-shot label tokens receive non-null identifiers, whereas all query and generated-output positions use $z_t=z_{\varnothing}$ during training and inference. %The gold query label is used only as a loss target and is never embedded, preventing target leakage. 
The four signals are embedded separately and concatenated at token position $t$:
\begin{equation}
\footnotesize
\mathbf{e}_t=
\left[
E_{\mathrm{abs}}(p_t);\,
E_{\mathrm{rel}}(r_t);\,
E_{\mathrm{lbl}}(z_t);\,
E_{\mathrm{time}}(\tau_t)
\right].
\label{eq:temporal_embeddings}
\end{equation}
$E_{\mathrm{abs}}$, $E_{\mathrm{rel}}$, $E_{\mathrm{lbl}}$, and $E_{\mathrm{time}}$ are learned embeddings, and $[\,;\,]$ denotes concatenation. The result is projected and added to the original token embedding:
\begin{equation}
\footnotesize
\mathbf{u}_t = \operatorname{MLP}(\mathbf{e}_t), \qquad
\mathbf{x}'_t = \mathbf{x}_t + W\mathbf{u}_t.
\label{eq:conditioned_embeddings}
\end{equation}
Here, $\mathbf{x}_t$ is the original token embedding, $\mathbf{u}_t$ the projected conditioning vector, and $W$ its mapping to the model dimension. A shared global label space assigns unique identifiers to dataset-specific labels, enabling one auxiliary classifier across datasets.

\subsection{Joint Training Objective}
\label{sec:training}

For a structured prompt prefix $Q_k$ and target response $Y$, the
complete training sequence is
\begin{equation}
\footnotesize
\begin{aligned}
\texttt{input\_ids}
={}&
\operatorname{tok}(Q_k)
\oplus
\operatorname{tok}(Y)\\[-1pt]
&\oplus
\operatorname{tok}(\texttt{</output>})
\oplus
[\texttt{eos}].
\end{aligned}
\label{eq:training_input}
\end{equation}
Here, $\operatorname{tok}(\cdot)$ denotes tokenization, $\oplus$ concatenation, and \texttt{eos} the end-of-sequence token. Let $\mathcal{I}^{(q)}_{H}$ be the query \texttt{<hist>} token positions and $\mathbf{h}_t$ the final hidden state at position $t$. The pooled query-history representation is
{\setlength{\abovedisplayskip}{1pt}
\setlength{\belowdisplayskip}{1pt}
\begin{equation}
\footnotesize
\mathbf{h}_{H}
=
\frac{1}{|\mathcal{I}^{(q)}_{H}|}
\sum_{t\in\mathcal{I}^{(q)}_{H}}
\mathbf{h}_t.
\label{eq:history_pooling}
\end{equation}}For the shared global classifier, let $t_Q$ be the opening \texttt{<output>} position and $\mathbf{h}_Q=\mathbf{h}_{t_Q}$ its final hidden state. %Since $t_Q$ precedes the target under causal attention, $\mathbf{h}_Q$ cannot access target tokens. 
The classifier computes:
\begin{equation}
\footnotesize
\mathbf{p}_{\mathrm{cls}}
=
\operatorname{softmax}
\left(
W_{\mathrm{cls}}\mathbf{h}_{Q}
+
\mathbf{b}_{\mathrm{cls}}
\right).
\label{eq:global_classifier}
\end{equation}
Let $\mathcal{I}_{\mathrm{CE}}$ contain prompt-prefix positions outside \texttt{<hist>} spans, and $\mathcal{I}_{\mathrm{out}}$ the target response, \texttt{</output>}, and \texttt{eos} positions. These masks are disjoint, and padding is excluded from all language-modeling losses. The four losses introduced earlier in
\hyperref[sec:method]{\underline{\textcolor{blue}{Sec.~\ref*{sec:method}}}}
are combined as follows:

\begin{equation}
\footnotesize
\begin{aligned}
\mathcal{L}={}&
\lambda_{\mathrm{CE}}\mathcal{L}_{\mathrm{CE}}
+
\lambda_{\mathrm{out}}\mathcal{L}_{\mathrm{focalLM}}\\[-1pt]
&+
\lambda_{\mathrm{cls}}\mathcal{L}_{\mathrm{focalCls}}
+
\lambda_{\mathrm{hist}}\mathcal{L}_{\mathrm{histCls}}.
\end{aligned}
\label{eq:total_loss}
\end{equation}$\mathcal{L}_{\mathrm{CE}}$ is next-token cross-entropy over prompt-prefix positions $\mathcal{I}_{\mathrm{CE}}$. $\mathcal{L}_{\mathrm{focalLM}}$ emphasizes difficult predictions over output positions $\mathcal{I}_{\mathrm{out}}$. $\mathcal{L}_{\mathrm{focalCls}}$ is focal cross-entropy computed from $\mathbf{p}_{\mathrm{cls}}$ in \hyperref[eq:global_classifier]{\underline{\textcolor{blue}{Eq.~\ref*{eq:global_classifier}}}}. $\mathcal{L}_{\mathrm{histCls}}$ applies an auxiliary focal loss to the pooled history representation $\mathbf{h}_{H}$ in \hyperref[eq:history_pooling]{\underline{\textcolor{blue}{Eq.~\ref*{eq:history_pooling}}}}. These objectives preserve instruction following, prioritize target prediction, and support heterogeneous label spaces. Stages are optimised sequentially (\hyperref[sec:curriculum_ift]{\underline{\textcolor{blue}{Sec.~\ref*{sec:curriculum_ift}}}}).

 \newcommand{\sig}[1]{\textsuperscript{#1}}
\begin{table*}[t]
\centering
\scriptsize
\setlength{\tabcolsep}{2.2pt}
\renewcommand{\arraystretch}{0.96}

\resizebox{\textwidth}{!}{%
\begin{tabular}{l|ccc|ccc|ccc||ccc|ccc}
\toprule

\textbf{Dataset}
& \multicolumn{3}{c|}{\cellcolor{green!10}\textbf{AnnoMI}}
& \multicolumn{3}{c|}{\cellcolor{green!10}\textbf{LRS}}
& \multicolumn{3}{c||}{\cellcolor{green!10}\textbf{TalkLife}}
& \multicolumn{3}{c|}{\cellcolor{red!10}\textbf{Reddit (OOD)}}
& \multicolumn{3}{c}{\cellcolor{red!10}\textbf{CMV (OOD)}} \\

\textbf{Shots}
& \cellcolor{green!10}\textbf{0}
& \cellcolor{green!10}\textbf{1}
& \cellcolor{green!10}\textbf{3}
& \cellcolor{green!10}\textbf{0}
& \cellcolor{green!10}\textbf{1}
& \cellcolor{green!10}\textbf{3}
& \cellcolor{green!10}\textbf{0}
& \cellcolor{green!10}\textbf{1}
& \cellcolor{green!10}\textbf{3}
& \cellcolor{red!10}\textbf{0}
& \cellcolor{red!10}\textbf{1}
& \cellcolor{red!10}\textbf{3}
& \cellcolor{red!10}\textbf{0}
& \cellcolor{red!10}\textbf{1}
& \cellcolor{red!10}\textbf{3} \\

\midrule

\textbf{OLMo-1B (IFT)}
& .18 & .18 & .18
& .35 & .36 & .37
& .23 & .22 & .29
& .14 & .27 & .15
& .36 & .36 & .37 \\

\rowcolor{teal!8}
\textbf{OLMo-1B (LiFT)}
& \textbf{.29}\sig{***}
& \textbf{.34}\sig{***}
& \textbf{.36}\sig{***}
& \textbf{.45}\sig{***}
& \textbf{.47}\sig{***}
& \textbf{.53}\sig{***}
& .22
& \textbf{.25}\sig{***}
& \textbf{.32}\sig{*}
& \textbf{.30}\sig{***}
& \textbf{.32}\sig{***}
& \textbf{.32}\sig{***}
& \textbf{.40}\sig{***}
& \textbf{.38}\sig{**}
& \textbf{.39}\sig{***} \\

\midrule

\textbf{OLMo-7B (IFT)}
& .21 & .38 & .45
& .12 & .13 & .13
& .03 & .01 & .06
& .23 & .22 & .30
& .48 & .47 & .49 \\

\rowcolor{teal!8}
\textbf{OLMo-7B (LiFT)}
& \textbf{.34}\sig{***}
& \textbf{.44}
& \textbf{.45}
& \textbf{.38}
& \textbf{.58}\sig{***}
& \textbf{.59}\sig{***}
& \textbf{.30}\sig{***}
& \textbf{.31}\sig{***}
& \textbf{.32}\sig{***}
& \textbf{.30}
& \textbf{.30}\sig{***}
& \textbf{.32}
& \textbf{.60}\sig{***}
& \textbf{.61}\sig{***}
& \textbf{.62}\sig{***} \\

\midrule

\textbf{LLaMA-8B (IFT)}
& .21 & .35 & .40
& .44 & .46 & .43
& .07 & .07 & .18
& .29 & .29 & .29
& .47 & .47 & .46 \\

\rowcolor{teal!8}
\textbf{LLaMA-8B (LiFT)}
& .21
& \textbf{.38}
& \textbf{.46}
& \textbf{.48}\sig{**}
& \textbf{.51}\sig{***}
& \textbf{.51}\sig{***}
& \textbf{.15}\sig{***}
& \textbf{.18}\sig{***}
& \textbf{.23}\sig{***}
& \textbf{.30}
& \textbf{.32}
& \textbf{.38}\sig{**}
& \textbf{.56}\sig{***}
& \textbf{.57}\sig{***}
& \textbf{.58}\sig{***} \\
\midrule

\textbf{Qwen-14B (IFT)}
& .36 & .27 & .20
& .38 & .41 & .40
& .18 & .03 & .09
& .11 & .17 & .13
& .51 & .55 & .55 \\

\rowcolor{teal!8}
\textbf{Qwen-14B (LiFT)}
& \textbf{.51}\sig{***}
& \textbf{.63}\sig{***}
& \textbf{.69}\sig{***}
& \textbf{.45}\sig{***}
& \textbf{.47}\sig{***}
& \textbf{.50}\sig{***}
& \textbf{.32}\sig{***}
& \textbf{.32}\sig{***}
& \textbf{.34}\sig{***}
& \textbf{.26}\sig{***}
& \textbf{.38}\sig{***}
& \textbf{.38}\sig{***}
& .48\sig{$\dagger$}
& \textbf{.58}\sig{***}
& \textbf{.62}\sig{***} \\

\midrule

\textbf{Qwen-32B (IFT)}
& .51 & .52 & .56
& .36 & .34 & .31
& .29 & .30 & .30
& .22 & .31 & .39
& .65 & .70 & .71 \\

\rowcolor{teal!8}
\textbf{Qwen-32B (LiFT)}
& \textbf{.59}\sig{***}
& \textbf{.61}\sig{***}
& \textbf{.63}\sig{***}
& \textbf{.38}\sig{***}
& \textbf{.39}\sig{**}
& \textbf{.40}\sig{**}
& \textbf{.35}
& \textbf{.32}
& \textbf{.38}
& \textbf{.29}\sig{***}
& \textbf{.38}\sig{***}
& \textbf{.40}\sig{***}
& \textbf{.66}
& \textbf{.71}
& \textbf{.74} \\

\bottomrule
\end{tabular}%
}

\caption{Red-shaded datasets are OOD. Bold indicates LiFT improvements based on the Macro-F1 scores. Superscripts report BH-corrected paired-bootstrap significance ($^{*}p<.05$, $^{**}p<.01$, $^{***}p<.001$); $^{\dagger}$ indicates a significant decrease. Three-seed results are provided in \hyperref[tab:sota_comparison_detailed]{\underline{\textcolor{blue}{Table~\ref*{tab:sota_comparison_detailed}}}}. \hyperref[sec:icl_amplification]{\underline{\textcolor{blue}{Section~\ref*{sec:icl_amplification}}}} and \hyperref[tab:icl_amplification]{\underline{\textcolor{blue}{Table~\ref*{tab:icl_amplification}}}} report shot-dependent gains and in-context amplification.}
\label{tab:sota_comparison}
\end{table*}

\begin{figure*}[!h]
    \centering
    \includegraphics[
        width=\linewidth,
        trim=7 5 5 5,
        clip
    ]{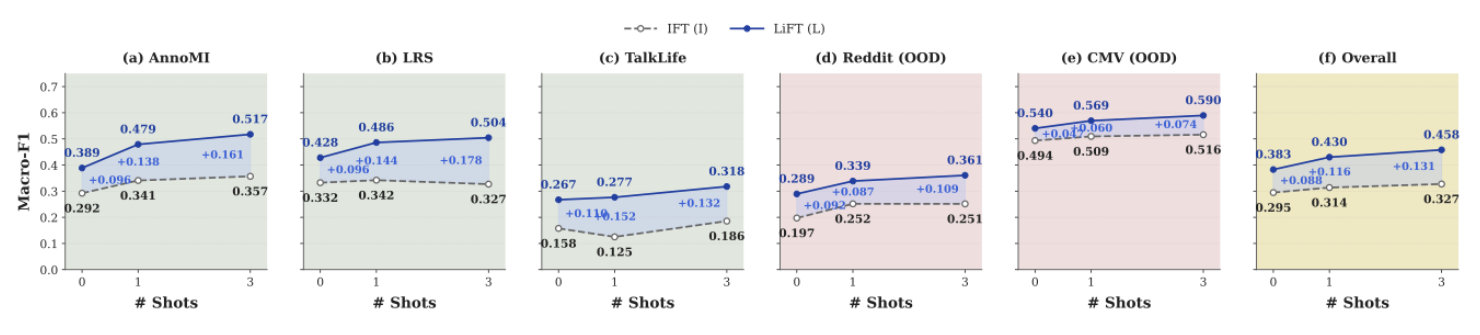}
   
\caption{LiFT vs.\ IFT: Panels (a--f) report Macro-F1 averaged across model families for 0-, 1-, and 3-shot prompting. Per-model gains, aggregation details, and aggregated values are provided in \hyperref[fig:plot2]{\underline{\textcolor{blue}{Figure~\ref*{fig:plot2}}}}, \hyperref[sec:fig]{\underline{\textcolor{blue}{Section~\ref*{sec:fig}}}}, and 
\hyperref[tab:aggregation]{\underline{\textcolor{blue}{Table~\ref*{tab:aggregation}}}}.}

\label{fig:plot1}
\end{figure*}
\section{Experiments}
\label{sec:experimentation}
\subsection{Tasks and Datasets}
\label{sec:datasets}
We evaluate our proposed LiFT framework on five longitudinal datasets (three in-domain, two OOD):

\noindent \textbf{AnnoMI}~\citep{wu2022anno} is a corpus of expert annotated motivational interviewing dialogues. Utterances are labeled as \textit{Change}, \textit{Sustain}, or \textit{Neutral}.\\
\noindent \textbf{Longitudinal Rumour Stance (LRS)}~\citep{tseriotou-etal-2024-sig} contains stance-annotated, timestamped and ordered Twitter timelines on rumourous claims. Posts are labeled as \textit{Switch} and \textit{No Switch} based on the
numbers of supporting tweets versus denying and
questioning ones.\\
\noindent \textbf{TalkLife Moments of Change (MoC)}~\citep{tsakalidis2022identifying} contains sequences of chronologically ordered mental health posts labeled as IS (\textit{In-Switch}), IE (\textit{In-Escalation}), or O (\textit{No change}).\\
\noindent \textbf{Reddit MoC} \cite{tsakalidis-etal-2022-overview} captures mood transitions across Reddit user timelines. The schema is the same as TalkLife MoC. \\%Consecutive post transitions are labeled as \textbf{S} (\textit{Shift}), \textbf{E} (\textit{Escalation}), or \textbf{0} (\textit{No change}).
\noindent\textbf{Change My View (CMV)} \cite{zhang2025prime} is a dataset collected from the corresponding Reddit forum. We use the forum's \textit{delta} mechanism to mark change-in-view replies and capture if a given reply changes the author's view in the flattened thread: 1 (\textit{change}) and 0 (\textit{no change}).\\
We use LRS, TalkLife, and AnnoMI for instruction fine-tuning and in-domain evaluation, applying an 80/20 training–test split. Reddit and CMV are used exclusively for OOD evaluation (\hyperref[tab:dataset_stats]{\underline{\textcolor{blue}{Table~\ref*{tab:dataset_stats}}}}, \hyperref[tab:dataset_overview_presentation]{\underline{\textcolor{blue}{Table~\ref*{tab:dataset_overview_presentation}}}}).

\subsection{Baselines and Experimental Setup}
\label{sec:experiments}
We compare instruction-tuned models with their LiFT counterparts under identical prompting conditions across datasets and shot settings. We evaluate open-weight models across architectures and scales: \textit{OLMo (1B, 7B), LLaMA (8B), and Qwen (14B, 32B)} \footnote{Model repository links are in
\hyperref[app:model_checkpoints]{\underline{\textcolor{blue}{Appendix~\ref*{app:model_checkpoints}}}}}
\hyperref[tab:sota_comparison] %{\underline{\textcolor{blue}{Table~\ref*{tab:sota_comparison}}}} 
and report performance using macro-F1.  We denote as \textit{IFT} the vanilla instruction-tuned
checkpoint before longitudinal adaptation, as \textit{SFT} the
standard fine-tuning baseline, and as \textit{LiFT} the
corresponding model after applying our longitudinal instruction fine-tuning framework. For the OOD and in-domain datasets in-context examples were drawn from a held-out subset excluded from testing %, preventing test-set leakage
and from the training splits respectively.

\noindent \textbf{IFT + ICL:}
We evaluate each vanilla instruction-tuned model using prompt-based
in-context learning with $k \in \{0,1,3\}$ demonstrations.
% Demonstrations
% are sampled from the training data and inserted into a prompt
% template.\\
% No parameter updates are performed at inference time.

\noindent \textbf{LiFT + ICL:}
We evaluate our LiFT models under the
same ICL conditions as the corresponding instruction-tuned models, using identical demonstrations,
prompt formatting, hyperparameters and decoding strategy. 
% This controlled comparison
% ensures that performance differences can be attributed to instruction
% fine-tuning rather than prompt engineering.
\subsection{Statistical Testing}
\label{sec:statistical_testing}

We assess if LiFT's gains are statistically reliable across seeds, datasets, and shots. For paired IFT--LiFT comparisons, we resample matched test examples and use paired bootstrap tests to estimate confidence intervals and significance for macro-F1, macro-recall, accuracy, and minority-class metrics. We use exact McNemar tests to measure whether LiFT fixes more IFT-model errors than it introduces, and sign tests to assess whether gains are directionally consistent across dataset--shot conditions. Finally, we use difference-in-differences to test whether demonstrations amplify LiFT more than instruction-tuned models. Full details and results are in  \hyperref[app:statistical_tests]{\underline{\textcolor{blue}{Appendix~\ref*{app:statistical_tests}}}}.

\section{Results and Discussion}
\label{sec:results}
\vspace{-0.3mm}
LLMs struggle when longitudinal signals are subtle, sparse, or distributed across observations. This is particularly evident in \hyperref[tab:sota_comparison]{\underline{\textcolor{blue}{Table~\ref*{tab:sota_comparison}}}} on TalkLife and Reddit: OLMo-7B IFT achieves only .029/.008/.064 macro-F1 on TalkLife, while Qwen-14B IFT obtains .11/.17/.13 on Reddit under 0-/1-/3-shot prompting. By contrast, LiFT improves macro-F1 in model--dataset--shot configurations reported in \hyperref[tab:sota_comparison]{\underline{\textcolor{blue}{Table~\ref*{tab:sota_comparison}}}}. For example, Qwen-14B LiFT improves over Qwen-14B IFT on AnnoMI from .36/.27/.20 to .51/.63/.69, while OLMo-7B LiFT improves over OLMo-7B IFT on CMV from .48/.47/.49 to .60/.61/.62. The incremental analysis in \hyperref[sec:icl_amplification]{\underline{\textcolor{blue}{Section~\ref*{sec:icl_amplification}}}} further shows positive average amplification across the 0--1--3-shot, indicating that LiFT strengthens both zero-shot performance and the effective use of in-context demonstrations.

\begin{figure}[t]
    \centering
    \includegraphics[
        width=\columnwidth,
        trim=10 6 8 10,
        clip
    ]{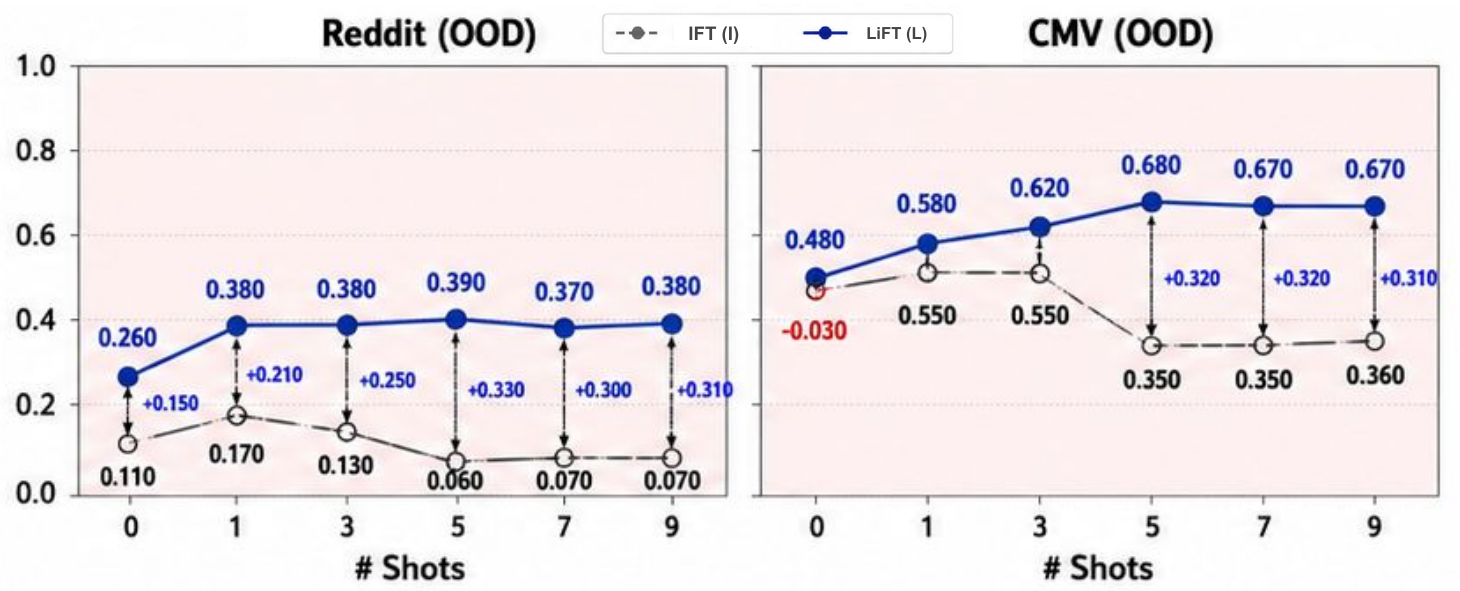}
    \caption{\textbf{LiFT/IFT under longer contexts.} Macro-F1 on OOD datasets using Qwen-14B for 0--9 shots (\hyperref[sec:additionalanalysis]{\underline{\textcolor{blue}{Appendix~\ref*{sec:additionalanalysis}}}}).}
    \label{fig:plot3}
\end{figure}

LiFT also improves performance on \textit{OOD datasets and task formulations}. This transfer suggests that LiFT learns reusable patterns of temporal persistence and change rather than only memorising dataset-specific labels. Minority-class improvements are strongest and most consistent for Qwen-14B and OLMo-7B. For smaller models, gains in macro-F1 more often result from improved balance across classes rather than uniformly higher minority-class recall. Therefore LiFT reduces majority-class bias in many sparse-change settings. It also reduces dependence on model scale through temporal structure that is absent in generic parameter scaling. Larger models retain advantages on some datasets, but for sparse longitudinal change detection, temporal inductive bias, available supervision, and effective adaptation capacity may be more important than raw parameter count.  The \textit{Additional Analysis} shown in
\hyperref[fig:plot3]{\underline{\textcolor{blue}{Fig~\ref*{fig:plot3}}}}, \hyperref[fig:plot5]{\underline{\textcolor{blue}{Fig~\ref*{fig:plot5}}}} extends the OOD evaluation 
to longer in-context settings for 
Qwen-14B and OLMo-7B. While LiFT generalises with more shots especially for larger models retaining its longitudinal capabilities, IFT is lower and often degrades at 5--9 shots.   The \textit{statistical analyses} (\hyperref[tab:stat_summary_main]{\underline{\textcolor{blue}{Table~\ref*{tab:stat_summary_main}}}}) provide further evidence that the improvements are systematic rather than driven by a small number of configurations. LiFT improves Macro-F1 in 72 of the 75 model--dataset--shot comparisons, with one tie and two decreases; 57 gains remain significant after Benjamini--Hochberg correction of the paired-bootstrap tests.

\begin{table*}[!t]
\centering
\scriptsize
\renewcommand{\arraystretch}{0.92}
\setlength{\tabcolsep}{1.2pt}
\setlength{\aboverulesep}{0.3pt}
\setlength{\belowrulesep}{0.3pt}

\newcolumntype{Y}{>{\centering\arraybackslash}X}

% Short row-level separators, matching the reference table
\newcommand{\shortvrule}{%
  \hspace{1pt}%
  \vrule width 0.4pt height 1.7ex depth 0.4ex%
  \hspace{1pt}%
}

\newcommand{\shortdoublevrule}{%
  \hspace{1pt}%
  \vrule width 0.4pt height 1.7ex depth 0.4ex%
  \hspace{0.8pt}%
  \vrule width 0.4pt height 1.7ex depth 0.4ex%
  \hspace{1pt}%
}

\begin{tabularx}{\textwidth}{
@{}l
!{\shortvrule}*{3}{Y}
!{\shortvrule}*{3}{Y}
!{\shortvrule}*{3}{Y}
!{\shortdoublevrule}*{3}{Y}
!{\shortvrule}*{3}{Y}
@{}
}
\toprule
&
\multicolumn{3}{c!{\shortvrule}}
{\cellcolor{green!10}\textbf{AnnoMI}} &
\multicolumn{3}{c!{\shortvrule}}
{\cellcolor{green!10}\textbf{LRS}} &
\multicolumn{3}{c!{\shortdoublevrule}}
{\cellcolor{green!10}\textbf{TalkLife}} &
\multicolumn{3}{c!{\shortvrule}}
{\cellcolor{red!10}\textbf{Reddit (OOD)}} &
\multicolumn{3}{c}
{\cellcolor{red!10}\textbf{CMV (OOD)}} \\

\cmidrule(lr){2-4}
\cmidrule(lr){5-7}
\cmidrule(lr){8-10}
\cmidrule(lr){11-13}
\cmidrule(lr){14-16}

\textbf{Shots}
& \cellcolor{green!10}\textbf{0}
& \cellcolor{green!10}\textbf{1}
& \cellcolor{green!10}\textbf{3}
& \cellcolor{green!10}\textbf{0}
& \cellcolor{green!10}\textbf{1}
& \cellcolor{green!10}\textbf{3}
& \cellcolor{green!10}\textbf{0}
& \cellcolor{green!10}\textbf{1}
& \cellcolor{green!10}\textbf{3}
& \cellcolor{red!10}\textbf{0}
& \cellcolor{red!10}\textbf{1}
& \cellcolor{red!10}\textbf{3}
& \cellcolor{red!10}\textbf{0}
& \cellcolor{red!10}\textbf{1}
& \cellcolor{red!10}\textbf{3} \\
\midrule

\textbf{IFT}
& .21 & .38 & .45
& .12 & .13 & .13
& .03 & .01 & .06
& .23 & .22 & .30
& .48 & .47 & .49 \\
\midrule

\rowcolor{orange!8}
SFT
& .29 & .31 & .33
& .33 & .35 & .38
& .07 & .05 & .06
& .27 & .24 & .28
& .50 & .47 & .51 \\

\rowcolor{orange!8}
LiFT w/o Temporal
& .28 & .32 & .33
& .31 & .35 & .38
& .06 & .05 & .06
& .27 & .24 & .29
& .48 & .48 & .50 \\

\rowcolor{orange!8}
LiFT w/o Curriculum
& .28 & .32 & .33
& .31 & .35 & .38
& .06 & .05 & .06
& .26 & .24 & .29
& .49 & .48 & .51 \\

\rowcolor{orange!8}
Reverse 1--2--3
& .29 & .33 & .34
& .33 & .36 & .39
& .07 & .06 & .06
& .27 & .25 & .30
& .51 & .49 & .53 \\

\rowcolor{orange!8}
Reverse 3--2--1
& .29 & .32 & .34
& .32 & .36 & .38
& .07 & .05 & .06
& .27 & .25 & .29
& .50 & .49 & .52 \\

\rowcolor{orange!8}
Mixed-task
& .27 & .30 & .32
& .28 & .33 & .36
& .05 & .05 & .05
& .25 & .22 & .27
& .47 & .46 & .49 \\

\rowcolor{orange!8}
LiFT w/o History Loss
& .28 & .31 & .34
& .29 & .36 & .38
& .06 & .05 & .06
& .26 & .25 & .29
& .49 & .49 & .52 \\

\rowcolor{teal!8}
\textbf{LiFT}
& \textbf{.34} & \textbf{.44} & \textbf{.45}
& \textbf{.38} & \textbf{.58} & \textbf{.59}
& \textbf{.30} & \textbf{.31} & \textbf{.32}
& \textbf{.30} & \textbf{.30} & \textbf{.32}
& \textbf{.60} & \textbf{.61} & \textbf{.62} \\
\bottomrule
\end{tabularx}

\caption{\textit{OLMo-7B ablation:} Each variant removes or modifies one LiFT component while keeping all other components unchanged. Full results are provided in \hyperref[tab:ablation_all]{\underline{\textcolor{blue}{Table~\ref*{tab:ablation_all}}}} and \hyperref[fig:ablation_absolute]{\underline{\textcolor{blue}{Figure~\ref*{fig:ablation_absolute}}}}.}
\label{tab:ablation_olmo7b}
\end{table*}

% \section{Results and Discussion}
% \label{sec:results}
% \hyperref[tab:sota_comparison]{\underline{\textcolor{blue}{Table~\ref*{tab:sota_comparison}}}} reports macro-F1 across five datasets under 0-, 1-, and 3-shot prompting, with significance markers on LiFT rows denoting paired-bootstrap LiFT--Base differences. Base models struggle on longitudinal change detection, with performance particularly weak on datasets containing subtle or sparse change signals such as TalkLife and Reddit. For example, OLMo-7B (IFT) achieves only .029/.008/.064 on TalkLife, while Qwen-14B (IFT) obtains .107/.167/.127 on Reddit across 0/1/3-shot settings. LiFT improves macro-F1 across  datasets and model sizes. Qwen-14B improves from .357/.271/.204 to .513/.626/.687 on AnnoMI, while OLMo-7B increases from .482/.470/.486 to .601/.606/.618 on CMV. \hyperref[tab:stat_summary_main]{\underline{\textcolor{blue}{Table~\ref*{tab:stat_summary_main}}}} summarizes the statistical evidence. LiFT improves macro-F1 in 56/60 dataset--shot settings, with 49 significant paired-bootstrap gains after Benjamini--Hochberg correction. McNemar tests further show that LiFT usually fixes more Base errors than it introduces (47/60 significant tests; median OR=1.89). The gains are not purely a function of model size: Qwen-14B shows the strongest average improvement, but OLMo-7B also gains substantially, while LLaMA-8B improves more modestly. This suggests that LiFT's effect depends on model family and baseline ICL behavior, not parameter count alone. Few-shot prompting provides additional but dataset-dependent gains: 

% ------------------------------------------------------------
% TABLE 3: Statistical summary aligned to Tables 12 and 15
% ------------------------------------------------------------
\begin{table}[t]
\centering
\scriptsize
\setlength{\tabcolsep}{3.0pt}
\renewcommand{\arraystretch}{1.08}
\resizebox{\columnwidth}{!}{%
\begin{tabular}{@{}lcccc@{}}
\toprule
\textbf{Family} & \textbf{$\Delta$F1} & \textbf{Wins/Sig.}
& \textbf{McN. OR} & \textbf{DID} \\
\midrule
\rowcolor{orange!8}

OLMo-1B  & $+.090/.095$ & $14/15;14\,(+1\,\mathrm{dec.})$ & $1.65;12$ & $+.016;5$ \\
\rowcolor{orange!8}

OLMo-7B  & $+.184/.132$ & $15/15;10$ & $10.12;14$ & $+.003;3$ \\
\rowcolor{orange!8}

LLaMA-8B & $+.062/.057$ & $14/15\,(1\,\mathrm{tie});10$ & $1.15;9$ & $+.018;5$ \\
\rowcolor{orange!8}

Qwen-14B & $+.172/.153$ & $14/15;14\,(+1\,\mathrm{dec.})$ & $2.14;12$ & $+.066;8$ \\
\rowcolor{orange!8}

Qwen-32B & $+.050/.064$ & $15/15;9$ & $3.14;12$ & $+.004;9$ \\
\midrule
\rowcolor{teal!8}

\textbf{Overall}  & \textbf{$+.112/.081$} & \textbf{$72/75;57\,(+2\,\mathrm{dec.})$ }& \textbf{$2.14;59$} &\textbf{ $+.021;30$} \\
\bottomrule
\end{tabular}%
}
\caption{\textit{Statistical summary of LiFT over IFT.}
$\Delta$F1 reports the mean/median LiFT--IFT macro-F1 across
dataset--shot settings. Wins/Sig.\ reports positive wins and
BH-adjusted significant paired-bootstrap gains. McN.\ OR reports the
median McNemar odds ratio and number of significant tests. DID reports
the mean few-shot difference-in-differences and number of significant
positive effects.}
\label{tab:stat_summary_main}
\end{table}

% difference-in-differences tests show positive few-shot amplification in 42/60 settings, with 21 significant effects. Minority-class tests show a more nuanced pattern: LiFT improves minority F1 most consistently for Qwen-14B and OLMo-7B, while smaller models often gain macro-F1 through improved class balance rather than uniformly higher minority recall. Improvements transfer to held-out Reddit and CMV, indicating that LiFT learns transferable longitudinal reasoning patterns rather than dataset-specific heuristics. Full confidence intervals, minority-class analyses, McNemar statistics, and DID tests are reported in \hyperref[app:statistical_tests]{\underline{\textcolor{blue}{Appendix~\ref*{app:statistical_tests}}}}.

\section{Ablation Study}
\label{sec:ablation}
We conduct ablations on OLMo-7B, modifying one LiFT component at a time. Results are in~\hyperref[tab:ablation_olmo7b]{\underline{\textcolor{blue}{Table~\ref*{tab:ablation_olmo7b}}}} (more details in\hyperref[sec:ablation_app]{\underline{\textcolor{blue}{~Appendix~\ref*{sec:ablation_app}}}} and \hyperref[sec:controlled]{\underline{\textcolor{blue}{~\ref*{sec:controlled}}}}).

\noindent{\textbf{SFT:}}
Standard SFT improves but remains below LiFT, especially on TalkLife and CMV. It is trained on the same datasets and few-shot prompts as LiFT, but excludes its components of curriculum ordering and temporal–label conditioning.\\
\noindent{\textbf{Temporal conditioning:}}
Removing this component reduces performance across most settings, particularly on LRS, TalkLife, and CMV, demonstrating the importance of explicit position, recency, and timestep signals for longitudinal modelling.\\
\noindent\textbf{Curriculum:}
Removing the AnnoMI ($\rightarrow$) LRS ($\rightarrow$) TalkLife curriculum consistently reduces performance. Reversing either the (k)-order or the dataset order also underperforms the original progression, indicating that LiFT benefits from aligning increasing dataset complexity with more shots.

\noindent\textbf{Mixed-task training:}
Joint training on all three datasets without a curriculum yields the weakest performance, indicating that ordered progression supports cross-task transfer effectively than simultaneous exposure to longitudinal tasks.\\
\noindent{\textbf{History-aware loss:}}
Removing history lowers performance, suggesting that direct supervision of the historical representation encourages the model to use temporally distributed evidence.\\
\noindent\textbf{Full LiFT:} performs best across all datasets and shot settings, demonstrating that its gains arise from the coordinated contribution of temporal conditioning, curriculum learning, sequential training, and history-aware supervision.

\section{Mechanistic Interpretability Protocol}
\label{sec:mechanistic}

\begin{figure*}[t]
    \centering
    \includegraphics[
        width=0.92\textwidth,
        height=0.30\textheight,
        keepaspectratio,
        trim=15 5 5 8,
        clip
    ]{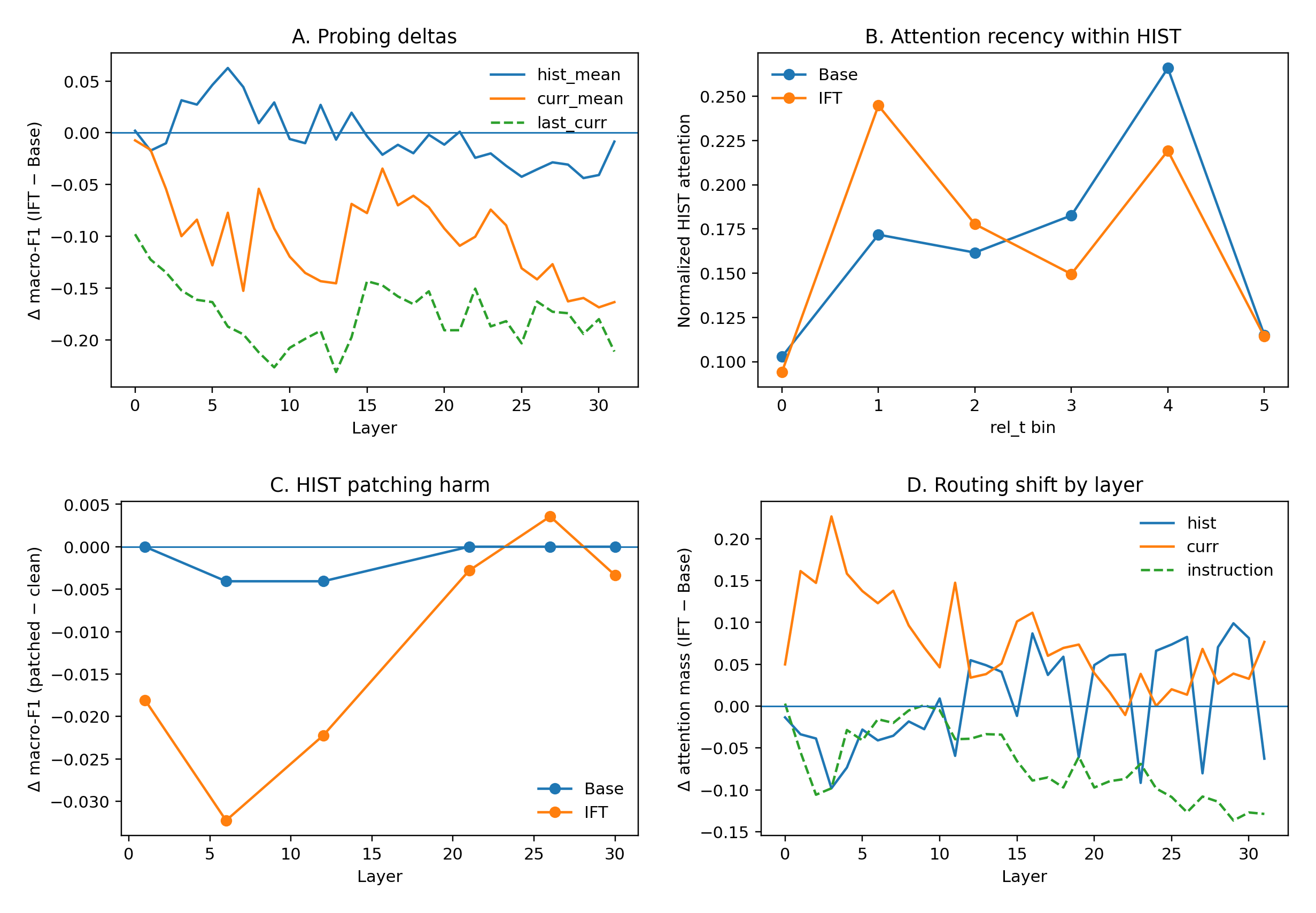}
    \caption{\textit{Mechanistic analysis of LiFT versus IFT using OLMo-7B.}
    \textbf{(A) Probing deltas} measure the change in label decodability, with positive values indicating that LiFT representations encode more label-relevant information than IFT.
    \textbf{(B) Normalized history-recency attention} measures the distribution of attention within the history, where larger values at smaller \texttt{rel\_t} indicate a stronger focus on recent observations.
    \textbf{(C) Activation-patching harm} reports the reduction in Macro-F1 when corrupted \texttt{<HIST>} activations are inserted into an otherwise clean forward pass, indicating causal dependence on the learned history representations.
    \textbf{(D) Routing shift by layer} shows how attention from the prediction position is redistributed across input regions under LiFT relative to IFT.
    Together, these analyses connect improved representation of label-relevant information \textbf{(A)}, temporal attention and routing \textbf{(B,D)}, and causal reliance on history representations \textbf{(C)}.}
    \label{fig:mechanistic}
\end{figure*}
We conduct four OLMo-7B analyses on TalkLife to test whether gains in \hyperref[sec:experiments]{\underline{\textcolor{blue}{Sec.~\ref*{sec:experiments}}}} stem from temporal modeling, few-shot context, or capacity. 

\noindent \textbf{Probing:} We perform layer-wise probing to measure where task-relevant information becomes linearly decodable. For each layer, we extract representations from four prompt regions: history (\texttt{hist\_mean}), few-shot demonstrations (\texttt{fewshot\_mean}), the current post (\texttt{curr\_mean}), and the final decision-facing token (\texttt{last\_curr}). Linear probes (StandardScaler + Logistic Regression with class balancing) are trained with 5-fold stratified cross-validation to predict the gold label. This measures \emph{linear recoverability} of label information from different prompt regions. Differences between models are shown in \hyperref[fig:mechanistic]{\underline{\textcolor{blue}{Fig.~\ref*{fig:mechanistic}A}}}\\
\noindent \textbf{Attention analysis:} We analyze how the prediction position routes attention across prompt regions (\texttt{instruction}, \texttt{fewshot}, \texttt{hist}, \texttt{curr}, \texttt{other}). We further compute normalized within-history recency profiles using relative temporal bins (\texttt{rel\_t}) to test whether models preferentially attend to more recent history items. \hyperref[fig:mechanistic]{\underline{\textcolor{blue}{Fig.~\ref*{fig:mechanistic}B}}},~\hyperref[fig:mechanistic]{\underline{\textcolor{blue}{~\ref*{fig:mechanistic}D}}} show normalised recency profiles and layer-wise routing shifts. \\
%The normalized recency profiles and layer-wise routing shifts are shown in \hyperref[fig:mechanistic]{\underline{\textcolor{blue}{Fig.~\ref*{fig:mechanistic}B}}}, and
\noindent \textbf{Activation patching on \texttt{<HIST>}:} We perform activation patching on the history span to test causal dependence on historical context. For 150 stratified test examples, we create corrupted variants by shuffling history posts while preserving the \texttt{<HIST>} span. Corrupted hidden states are patched to clean runs at selected layers, and we measure changes in macro-F1, logits, margins, and prediction flips. %It gives causal evidence for the model’s reliance on temporally structured history. 
This provides causal evidence of the model's reliance on temporally structured history, as per activation-patching harm results (\hyperref[fig:mechanistic]{\underline{\textcolor{blue}{Fig.~\ref*{fig:mechanistic}C}}}).

\noindent \textbf{Mechanistic Interpretation of IFT Gains}
\label{sec:mechanistic_results}

\noindent \hyperref[fig:mechanistic]{\underline{\textcolor{blue}{Fig.~\ref*{fig:mechanistic}}}} 
explains the gains reported in 
\hyperref[sec:experiments]{\underline{\textcolor{blue}{Sec.~\ref*{sec:experiments}}}}.
The key question is whether LiFT improves longitudinal modeling by encouraging use of temporally ordered history, rather than strengthening local current-token classification. %In 
%\hyperref[fig:mechanistic]{\underline{\textcolor{blue}{Fig.~\ref*{fig:mechanistic}}}}, 
A \emph{positive} delta in Panels A and D indicates larger values for LiFT than the IFT model, while in Panel C a \emph{more negative} delta indicates stronger performance degradation when corrupted history activations are patched, implying greater causal reliance on history. \textit{Probing results} in 
\hyperref[fig:mechanistic]{\underline{\textcolor{blue}{Fig.~\ref*{fig:mechanistic}A}}} 
show higher linear decodability in \texttt{hist\_mean} representations, particularly in early-to-mid layers, and reduced decodability in \texttt{curr\_mean} and \texttt{last\_curr}, suggesting LiFT shifts task-relevant information toward history representations rather than strengthening the current token as a local classifier. \textit{Attention patterns} in 
\hyperref[fig:mechanistic]{\underline{\textcolor{blue}{Fig.~\ref*{fig:mechanistic}B}}} %,~\hyperref[fig:mechanistic]{\underline{\textcolor{blue}{~\ref*{fig:mechanistic}D}}} 
%support this interpretation, with
%\hyperref[fig:mechanistic]{\underline{\textcolor{blue}{~\ref*{fig:mechanistic}B}}} 
show increased normalized attention to smaller \texttt{rel\_t} bins, indicating stronger focus on recent history, and \hyperref[fig:mechanistic]{\underline{\textcolor{blue}{~\ref*{fig:mechanistic}D}}} shows LiFT shifts prediction-time attention away from instruction scaffolding toward evidence-bearing regions, especially \texttt{hist} and \texttt{curr}, indicating temporal context is actively used during decision-time computation. \textit{Activation patching }
(\hyperref[fig:mechanistic]{\underline{\textcolor{blue}{Fig.~\ref*{fig:mechanistic}C}}}) 
provides causal evidence. Temporally shuffling history and patching corrupted \texttt{<HIST>} activations into clean runs harms LiFT substantially more than the IFT model, particularly in early-to-mid layers (peaking around layers 6 and 12). This confirms that temporally structured history representations are causally necessary for prediction under LiFT.

\section{Conclusion}
% \vspace{-1.5mm}
We present LiFT, a longitudinal instruction fine-tuning framework that improves \textit{in-context learning} for longitudinal modelling by LLMs. Across datasets, domains, and model scales, LiFT outperforms IFT in evaluated configurations under identical prompts and demonstration budgets, with strong macro-F1 gains in OOD and imbalanced longitudinal settings. Minority-class improvements are strongest and most consistent for Qwen-14B and OLMo-7B. Interpretability analyses show that these improvements arise from greater reliance on temporally structured history rather than stronger local classification. Probing, attention analysis, and activation patching reveal that LiFT makes historical context more informative, selectively attended, and causally necessary for prediction.% \section{Conclusion}

\section{Limitations}
\label{sec:limitations}
Although our interpretability analyses provide evidence that LiFT
increases the use of temporally structured history, they do not fully
explain the internal mechanisms underlying the observed improvements.
Moreover, these analyses are restricted to OLMo-7B on TalkLife, with
activation patching conducted on 150 examples. 

Each model configuration is trained once, while the reported
three-seed results correspond to inference runs with fixed
demonstrations and deterministic decoding. These runs measure
inference stability rather than variation across independently trained
models. Consequently, the statistical analyses do not capture
optimization variance, and future work should repeat training across
multiple initialization and data-order seeds. Minority-class effects
also vary across model families, indicating that improvements in
macro-F1 do not always translate into uniform gains for rare classes.

The models are evaluated in controlled experimental settings and have
not been tested in real-world applications. Future work should examine
broader populations, distribution shifts over time, calibration,
fairness, robustness, and human oversight before considering deployment
in sensitive domains.

We evaluate only open-weight models. This choice is motivated by the
privacy, governance, and licensing constraints associated with
sensitive longitudinal mental-health data, as well as LiFT's requirement
for access to model weights and internal representations. All training and evaluation are therefore conducted
locally on secure infrastructure using open-weight models, which also
supports methodological control, transparency, and reproducibility.
\section{Ethical Considerations}
\label{sec:ethics}

This work involves datasets derived from conversational and mental health--related timelines. Although several datasets used in our experiments are publicly available and anonymized, research involving mental health signals requires careful ethical consideration. The models developed in this study are intended strictly for research purposes and should not be used for clinical diagnosis, mental health assessment, or decision-making without appropriate professional oversight. The TalkLife dataset contains sensitive and personal user-generated content. Appropriate ethics approval was obtained from the Institutional Review Board (IRB) prior to accessing and processing the data. Access to the dataset was granted through a research licensing agreement with TalkLife following approval of the submitted research proposal. In accordance with these requirements, all data were anonymized and handled using appropriate sensitive data management procedures. To further protect user privacy, all examples reported in this paper are paraphrased rather than reproduced verbatim. Model training and data processing were conducted on secure servers with authorized user-only access. Due to the sensitive nature of the data and the potential risks of misuse, the labeled TalkLife dataset and the trained models are not intended for public release. Furthermore, the Terms of Service of TalkLife state that user data may be shared for research purposes under appropriate safeguards. This study was conducted in compliance with those terms, and access to the TalkLife dataset was obtained through a formal research license agreement. All other datasets used in this work are publicly available and were accessed in accordance with their respective usage policies. Finally, large language models may inadvertently learn biases present in training data. When applied to sensitive domains such as mental health, these biases could lead to misleading or harmful interpretations. Any future deployment of such systems should therefore involve rigorous ethical review, fairness evaluation, and appropriate safeguards to protect user well-being and privacy.

\bibliography{custom}
\appendix
\section{Acronyms and dataset abbreviations}
See table \hyperref[tab:acronyms]{\underline{\textcolor{blue}{Table~\ref*{tab:acronyms}}}} for detailed Acronyms table.
% Required in the preamble:
% \usepackage[table]{xcolor}

\begin{table*}[!t]
\centering
\scriptsize

\rowcolors{2}{gray!8}{white}
\begin{tabularx}{\textwidth}{
    >{\raggedright\arraybackslash}p{0.13\textwidth}
    >{\raggedright\arraybackslash}X}
\toprule
\rowcolor{gray!18}
\textbf{Acronym} & \textbf{Definition} \\
\midrule

NLP
& Natural Language Processing. \\

LLM
& Large Language Model. \\

ICL
& In-Context Learning, where a model performs a task using demonstrations
provided in the input prompt without updating its parameters. \\

LiFT
& Longitudinal Instruction Fine-Tuning, the proposed framework for
improving longitudinal reasoning and in-context learning. \\

IFT
& Vanilla instruction-tuned checkpoint. \\

SFT
& Standard supervised fine-tuning baseline. \\

OOD
& Out-of-Distribution; a dataset or task that is excluded from LiFT
training and used to evaluate transfer to unseen longitudinal tasks. \\

EHR
& Electronic Health Record. \\

LoRA
& Low-Rank Adaptation, a parameter-efficient fine-tuning method. \\

QLoRA
& Quantized Low-Rank Adaptation, which applies LoRA to a quantized IFT
model. \\

LM
& Language Model. \\

MLP
& Multilayer Perceptron. \\

CE
& Cross-Entropy. \\

EOS
& End-of-Sequence token. \\

MI
& Motivational Interviewing. \\

AnnoMI
& Motivational Interviewing dataset containing therapist--client
dialogues annotated with Change, Sustain, and Neutral labels. \\

LRS
& Longitudinal Rumour Stance dataset. \\

CMV
& Change My View, a Reddit-based opinion-change dataset. \\

Sw
& Switch label in LRS, indicating a change in stance. \\

N-Sw
& No-Switch label in LRS, indicating no change in stance. \\

IS
& Switch label in TalkLife and Reddit, indicating a change in emotional
state. \\

IE
& Escalation label in TalkLife and Reddit, indicating an increase in the
intensity of the existing emotional state. \\

O
& No-change label in TalkLife and Reddit. \\

BH
& Benjamini--Hochberg multiple-comparison correction. \\

OR
& Odds Ratio, used to summarize the direction and magnitude of McNemar
test discordances. \\

DID
& Difference-in-Differences, used to test whether additional in-context
demonstrations improve LiFT more than the corresponding IFT model. \\

IRB
& Institutional Review Board. \\

GPU
& Graphics Processing Unit. \\

\bottomrule
\end{tabularx}
\caption{Acronyms and abbreviations used throughout the paper.}
\label{tab:acronyms}

\rowcolors{2}{white}{white}
\end{table*}

\section{Mathematical and Methodological Notation}
See table \hyperref[tab:notation]{\underline{\textcolor{blue}{Table~\ref*{tab:notation}}}}, \hyperref[tab:interpretability-notation]{\underline{\textcolor{blue}{Table~\ref*{tab:interpretability-notation}}}}, and \hyperref[tab:statistical-notation]{\underline{\textcolor{blue}{Table~\ref*{tab:statistical-notation}}}}
for detailed mathematical notation used in the LiFT formulation, prompt
construction, training objective, and statistical analysis.

% Required in the preamble:
% \usepackage[table]{xcolor}

\begin{table*}[!t]
\centering
\scriptsize

\rowcolors{2}{gray!8}{white}
\begin{tabularx}{\textwidth}{
    >{\raggedright\arraybackslash}p{0.18\textwidth}
    >{\raggedright\arraybackslash}X}
\toprule

\rowcolor{gray!18}
\textbf{Symbol} & \textbf{Definition} \\
\midrule

$H$
& Temporally ordered history associated with the current prediction
instance. \\

$H=\{(x_1,t_1),\ldots,(x_m,t_m)\}$
& Sequence of $m$ historical observations and their corresponding
timestamps, ordered such that $t_1\leq\cdots\leq t_m$. \\

$x_i$
& The $i$-th observation in a longitudinal sequence. \\

$t_i$
& Timestamp or temporal position associated with observation $x_i$. \\

$m$
& Number of observations included in the historical context. \\

$(x_{m+1},t_{m+1})$
& Current observation and its timestamp, for which a prediction is
required. \\

$\hat{y}$
& Textual output predicted by the generative formulation. \\

$f(H,x_{m+1})$
& Generative prediction function conditioned on the history and current
observation. \\

$\hat{z}$
& Predicted class label in the classification formulation. \\

$g(H,x_{m+1})$
& Classification function conditioned on the history and current
observation. \\

$\oplus$
& Concatenation of prompt components or token sequences. \\

$I$
& Natural-language task instruction included in the
\texttt{<instruction>} region. \\

$D_k$
& Set of $k$ in-context demonstrations included in the
\texttt{<few-shot>} region. \\

$k$
& Number of in-context demonstrations, or the shot setting. \\

$X$
& Query containing the historical context and current observation. \\

$Y$
& Target response or output label corresponding to query $X$. \\

$\texttt{<HIST>}$
& Structured prompt span containing the temporally ordered historical
observations. \\

$\texttt{<CURR>}$
& Structured prompt span containing the current observation. \\

$\texttt{MAX\_TOKENS}$
& Maximum token budget allowed for the constructed input prompt. \\

$p_t$
& Absolute token-position identifier for token $t$. \\

$r_t$
& Relative position of token $t$ with respect to the end of the
sequence. \\

$z_t$
& Global label identifier associated with token $t$. \\

$\tau_t$
& Absolute timestep identifier associated with token $t$. \\

$e_t^{\mathrm{abs}}$
& Embedding of the absolute token position $p_t$. \\

$e_t^{\mathrm{rel}}$
& Embedding of the relative position-to-end $r_t$. \\

$e_t^{\mathrm{lbl}}$
& Embedding of the global label identifier $z_t$. \\

$e_t^{\mathrm{time}}$
& Embedding of the absolute timestep identifier $\tau_t$. \\

$e_t$
& Concatenated temporal--label conditioning representation:
\[
e_t =
\left(
e_t^{\mathrm{abs}},
e_t^{\mathrm{rel}},
e_t^{\mathrm{lbl}},
e_t^{\mathrm{time}}
\right).
\] \\

$\operatorname{MLP}(\cdot)$
& Multilayer perceptron used to project the concatenated conditioning
embeddings. \\

$u_t$
& Projected conditioning representation,
$u_t=\operatorname{MLP}(e_t)$. \\

$x_t$
& Original language-model token embedding at position $t$. \\

$x'_t$
& Temporally conditioned token embedding,
$x'_t=x_t+Wu_t$. \\

$W$
& Learned projection matrix mapping $u_t$ into the language-model
embedding space. \\

$\operatorname{tok}(\cdot)$
& Tokenization operation applied to a text sequence. \\

$\texttt{input\_ids}$
& Tokenized training sequence formed by concatenating the prompt, target
response, and end-of-sequence token. \\

$[\texttt{eos}]$
& End-of-sequence token appended to each training instance. \\

$\mathcal{L}$
& Total LiFT training objective. \\

$\mathcal{L}_{\mathrm{CE}}$
& Next-token cross-entropy loss computed on prompt tokens outside the
\texttt{<HIST>} span. \\

$\mathcal{L}_{\mathrm{focalLM}}$
& Focal language-modeling loss applied to the output tokens. \\

$\mathcal{L}_{\mathrm{focalCls}}$
& Focal classification loss over the shared global label space. \\

$\mathcal{L}_{\mathrm{histCls}}$
& Auxiliary history-only classification loss computed from a pooled
representation of the \texttt{<HIST>} tokens. \\

$\lambda_{\mathrm{CE}}$
& Weight assigned to the prompt-level cross-entropy loss. \\

$\lambda_{\mathrm{out}}$
& Weight assigned to the focal output language-modeling loss. \\

$\lambda_{\mathrm{cls}}$
& Weight assigned to the global focal classification loss. \\

$\lambda_{\mathrm{hist}}$
& Weight assigned to the auxiliary history-only classification loss. \\

\midrule
\rowcolor{gray!18}
\multicolumn{2}{l}{\textit{Curriculum and algorithm notation}} \\
\midrule

$T$
& Tokenizer used to construct model inputs. \\

$B$
& Maximum context or token budget used by the curriculum builder. \\

$M$
& IFT language model adapted using LiFT. \\

$S=\{1,2,3\}$
& Ordered set of curriculum stages. \\

$s$
& Current curriculum-stage index. \\

$\mathcal{C}_s$
& Set of training instances assigned to curriculum stage $s$. \\

$H_x$
& Rolling historical context constructed for example $x$. \\

$\operatorname{TOKLEN}(\cdot)$
& Function returning the token length of an input sequence or prompt. \\

$b$
& Training batch sampled from curriculum stage $\mathcal{C}_s$. \\

$r$
& Rank of the Low-Rank Adaptation matrices. \\

$\alpha$
& Low-Rank Adaptation scaling parameter. \\

$\gamma$
& Focusing parameter used in the focal-loss objectives. \\

\bottomrule
\end{tabularx}
\caption{Mathematical notation used in the LiFT formulation, prompt
construction, training objective, and curriculum algorithm.}
\label{tab:notation}

\rowcolors{2}{white}{white}
\end{table*}

% Required in the preamble:
% \usepackage[table]{xcolor}

\begin{table*}[!t]
\centering
\small

\rowcolors{2}{gray!8}{white}
\begin{tabularx}{\textwidth}{
    >{\raggedright\arraybackslash}p{0.20\textwidth}
    >{\raggedright\arraybackslash}X}
\toprule
\rowcolor{gray!18}
\textbf{Term or symbol} & \textbf{Definition} \\
\midrule

$\texttt{hist\_mean}$
& Mean hidden-state representation computed over tokens in the
\texttt{<HIST>} span. It measures how much task-relevant information is
encoded in the historical context. \\

$\texttt{fewshot\_mean}$
& Mean hidden-state representation computed over tokens belonging to the
few-shot demonstrations. \\

$\texttt{curr\_mean}$
& Mean hidden-state representation computed over tokens in the current
observation. \\

$\texttt{last\_curr}$
& Hidden-state representation of the final decision-facing token in the
current-observation span. \\

$\mathrm{rel\_t}$
& Relative temporal bin assigned to an item within the history. Smaller
values represent items closer to the current observation. \\

$\Delta_{\mathrm{probe}}$
& Difference in probing performance between LiFT and the corresponding
IFT model for a given layer and prompt region. \\

$\Delta_{\mathrm{attn}}$
& Difference in attention allocation between LiFT and the corresponding
IFT model. \\

$\Delta_{\mathrm{patch}}$
& Change in model performance when corrupted history activations are
patched into a clean forward pass. \\

$\texttt{instruction}$
& Prompt region containing the natural-language task instruction. \\

$\texttt{fewshot}$
& Prompt region containing in-context demonstrations. \\

$\texttt{hist}$
& Prompt region containing the temporally ordered history. \\

$\texttt{curr}$
& Prompt region containing the current observation. \\

$\texttt{other}$
& Tokens that do not belong to the instruction, few-shot, history, or
current-observation regions. \\

$h_t^{(\ell)}$
& Hidden-state representation of token $t$ at model layer $\ell$. \\

$\ell$
& Transformer-layer index. \\

$\mathcal{H}$
& Set of token positions belonging to the \texttt{<HIST>} region. \\

$\bar{h}_{\mathrm{hist}}^{(\ell)}$
& Mean history representation at layer $\ell$:
\[
\bar{h}_{\mathrm{hist}}^{(\ell)}
=
\frac{1}{|\mathcal{H}|}
\sum_{t\in\mathcal{H}} h_t^{(\ell)}.
\] \\

$A_{\mathrm{region}}^{(\ell)}$
& Attention routed from the prediction position to a specified prompt
region at layer $\ell$. \\

$A_{\mathrm{hist},j}^{(\ell)}$
& Attention routed from the prediction position to history item $j$ at
layer $\ell$. \\

$\widetilde{A}_{\mathrm{hist},j}^{(\ell)}$
& Attention to history item $j$, normalized over all tokens within the
history region. \\

$H_{\mathrm{clean}}$
& Original temporally ordered history used in the clean forward pass. \\

$H_{\mathrm{corrupt}}$
& Corrupted history produced by shuffling the temporal order of the
historical observations. \\

$h_{\mathrm{clean}}^{(\ell)}$
& Hidden-state activation produced by the clean history at layer
$\ell$. \\

$h_{\mathrm{corrupt}}^{(\ell)}$
& Hidden-state activation produced by the corrupted history at layer
$\ell$. \\

$\Delta\mathrm{F1}_{\mathrm{patch}}$
& Change in macro-F1 after replacing clean history activations with
corrupted history activations at a selected layer. \\

$\Delta\mathrm{logit}$
& Change in the predicted class logit after activation patching. \\

$\Delta\mathrm{margin}$
& Change in the difference between the highest and second-highest class
logits after activation patching. \\

$\mathrm{FlipRate}$
& Proportion of examples for which activation patching changes the
predicted class. \\

\bottomrule
\end{tabularx}
\caption{Notation used in the mechanistic interpretability analyses.}
\label{tab:interpretability-notation}

\rowcolors{2}{white}{white}
\end{table*}

\begin{table}[!t]
\centering
\scriptsize

\rowcolors{2}{gray!8}{white}
\begin{tabularx}{\columnwidth}{
    >{\raggedright\arraybackslash}p{0.23\columnwidth}
    >{\raggedright\arraybackslash}X}
\toprule
\rowcolor{gray!18}
\textbf{Symbol} & \textbf{Definition} \\
\midrule

$\Delta F1$
& Difference in macro-F1 between LiFT and the corresponding IFT model. \\

$L_i$
& Indicator that the IFT model predicts example $i$ correctly. \\

$L_i$
& Indicator that the LiFT model predicts example $i$ correctly. \\

$\mathbb{I}[\cdot]$
& Indicator function. \\

$b_{01}$
& Number of examples incorrect under IFT but correct under LiFT. \\

$b_{10}$
& Number of examples correct under IFT but incorrect under LiFT. \\

$\mathrm{OR}$
& McNemar odds ratio:
\[
\mathrm{OR}
=
\frac{b_{01}+0.5}{b_{10}+0.5}.
\] \\

$m_k^{L}$
& LiFT performance under the $k$-shot setting. \\

$m_k^{I}$
& IFT performance under the $k$-shot setting. \\

$\operatorname{DID}_{a\rightarrow b}$
& Few-shot difference-in-differences:
\[
\operatorname{DID}_{a\rightarrow b}
=
(m_b^{L}-m_a^{L})
-
(m_b^{I}-m_a^{I}).
\] \\

$p$
& Statistical-test probability value. \\

\bottomrule
\end{tabularx}
\caption{Notation used in the statistical analyses.}
\label{tab:statistical-notation}

\rowcolors{2}{white}{white}
\end{table}
\section{Algorithm: LiFT: Longitudinal Curriculum Construction and Joint Fine-Tuning}

The LiFT pipeline is divided into two components.
\hyperref[alg:lift_builder]{\underline{\textcolor{blue}{Algorithm~\ref*{alg:lift_builder}}}}
describes the LiFT Builder, which converts heterogeneous temporal datasets into sequence-aware, stage-specific curriculum examples with rolling histories and few-shot demonstrations.
\hyperref[alg:lift_trainer]{\underline{\textcolor{blue}{Algorithm~\ref*{alg:lift_trainer}}}}
presents the LiFT Trainer, which sequentially optimizes the LoRA adapters, temporal--label conditioning module, and classification heads using the joint longitudinal training objective.

% -----------------------------
% Algorithm 1: LiFT Builder
% -----------------------------
\begin{algorithm}[!t]
\caption{LiFT Builder: Longitudinal Curriculum Construction}
\label{alg:lift_builder}
\scriptsize

\begin{algorithmic}[1]

\Require Training datasets
$\{\mathcal{D}_{\mathrm{AnnoMI}},
\mathcal{D}_{\mathrm{LRS}},
\mathcal{D}_{\mathrm{TalkLife}}\}$,
tokenizer $\mathcal{T}$, context budget $B$
\Require Ordered curriculum stages $\mathcal{S}=(1,2,3)$
and stage-specific demonstration counts $\{k_s\}$
\Ensure Curriculum datasets $\{\mathcal{C}_s\}_{s\in\mathcal{S}}$

\ForAll{stage $s \in \mathcal{S}$}
    \State Initialize $\mathcal{C}_s \gets \varnothing$
    \State Select the dataset $\mathcal{D}_s$ assigned to stage $s$

    \ForAll{sequence $Q \in \mathcal{D}_s$}
        \State Sort the observations in $Q$ chronologically

        \For{$j \gets 1$ to $|Q|$}
            \State Set the current item to $(x_j,t_j)$
            \State Construct the rolling history
            $\mathcal{H}_j \gets \{(x_i,t_i)\mid i<j\}$

            \While{
                $\Call{TokLen}{
                \mathcal{T},
                \texttt{<hist>}~\mathcal{H}_j~\texttt{</hist>}
                \oplus
                \texttt{<curr>}~x_j~\texttt{</curr}}
                > B$
                \textbf{and}
                $\mathcal{H}_j \neq \varnothing$
            }
                \State Remove the oldest observation from $\mathcal{H}_j$
            \EndWhile

            \State Assign relative temporal indices to
            $\mathcal{H}_j$, e.g., \texttt{t-3}, \texttt{t-2},
            and \texttt{t-1}
            \State Construct query $X_j$ from $\mathcal{H}_j$ and $x_j$
            \State Sample up to $k_s$ demonstrations
            $\mathcal{D}_{k_s}$ from the stage training data
            \State Construct prompt $P_{k_s}$ using
            Equation~\ref{eq:prompt_template}

            \While{
                $\Call{TokLen}{
                \mathcal{T},
                P_{k_s}\oplus Y_j\oplus[\texttt{eos}]}
                > B$
                \textbf{and}
                $|\mathcal{D}_{k_s}|>0$
            }
                \State Remove one demonstration from $\mathcal{D}_{k_s}$
                \State Reconstruct $P_{k_s}$
            \EndWhile

            \If{
                $\Call{TokLen}{
                \mathcal{T},
                P_{k_s}\oplus Y_j\oplus[\texttt{eos}]}
                \leq B$
            }
                \State Add
                $(P_{k_s},Y_j,\mathcal{H}_j,t_j,z_j)$
                to $\mathcal{C}_s$
            \EndIf
        \EndFor
    \EndFor
\EndFor

\State \Return $\{\mathcal{C}_s\}_{s\in\mathcal{S}}$

\end{algorithmic}
\end{algorithm}

% -----------------------------
% Algorithm 2: LiFT Trainer
% -----------------------------
\begin{algorithm}[!t]
\caption{LiFT Trainer: Joint Longitudinal Fine-Tuning}
\label{alg:lift_trainer}
\scriptsize

\begin{algorithmic}[1]

\Require Curriculum datasets $\{\mathcal{C}_s\}_{s\in\mathcal{S}}$,
tokenizer $\mathcal{T}$, IFT model $M$
\Require Ordered stages $\mathcal{S}=(1,2,3)$,
LoRA configurations $\{r_s,\eta_s\}$,
and loss weights $\boldsymbol{\lambda}$
\Ensure Trained LoRA adapters, temporal--label module,
and classification heads

\State Extend $\mathcal{T}$ with the LiFT structural tokens
\State Freeze the parameters of the IFT model $M$
\State Initialize LoRA adapters, the temporal--label module,
and the global and history classification heads

\ForAll{stage $s \in \mathcal{S}$ in curriculum order}
    \If{$s>1$}
        \State Load the trainable components from stage $s-1$
    \EndIf

    \State Configure the stage-specific LoRA rank $r_s$
    and learning rate $\eta_s$

    \ForAll{batch $b \in \mathcal{C}_s$}
        \State Construct $\texttt{input\_ids}$ using
        Equation~\ref{eq:training_input}
        \State Derive the prompt, history, output,
        and classification masks
        \State Derive $p_t$, $r_t$, $\tau_t$, and
        training label identifiers $z_t$

        \State Construct conditioned token representations
        $\mathbf{x}'_t$ using
        Equations~\ref{eq:temporal_embeddings}
        and~\ref{eq:conditioned_embeddings}

        \State Perform a forward pass through $M$
        \State Compute the pooled history representation
        $\mathbf{h}_{H}$ using Equation~\ref{eq:history_pooling}

        \State Compute
        $\mathcal{L}_{\mathrm{CE}}$,
        $\mathcal{L}_{\mathrm{focalLM}}$,
        $\mathcal{L}_{\mathrm{focalCls}}$, and
        $\mathcal{L}_{\mathrm{histCls}}$

        \State Compute the joint objective
        \[
        \mathcal{L}
        =
        \lambda_{\mathrm{CE}}\mathcal{L}_{\mathrm{CE}}
        +
        \lambda_{\mathrm{out}}\mathcal{L}_{\mathrm{focalLM}}
        +
        \lambda_{\mathrm{cls}}\mathcal{L}_{\mathrm{focalCls}}
        +
        \lambda_{\mathrm{hist}}\mathcal{L}_{\mathrm{histCls}}
        \]

        \State Update the LoRA adapters, temporal--label module,
        and classification heads
    \EndFor

    \State Save the stage-$s$ checkpoint
\EndFor

\State \Return trained LoRA adapters, temporal--label module,
and classification heads

\end{algorithmic}
\end{algorithm}
\section{Controlled comparison protocol}
\label{sec:controlled}
 All fine-tuned baselines and ablation variants are initialized from the same instruction-tuned checkpoint and use the same training examples, instruction schema, demonstration format and pools, context budget, total number of optimization steps, LoRA configuration, learning-rate schedule, random seeds, and decoding procedure. Each ablation changes only the component named in the corresponding row, while all other training and evaluation conditions remain fixed. IFT and LiFT are evaluated using identical query instances, demonstrations, prompt formatting, context histories, and decoding settings.
 
 \textbf{SFT + ICL:} The SFT baseline is trained on the same longitudinal instruction examples and with the same optimization budget as LiFT, but excludes curriculum ordering, temporal–label conditioning, the shared classification objective, and the history-only auxiliary objective. At evaluation, SFT uses the same 0-, 1-, and 3-shot prompts as IFT and LiFT.

\textbf{Training--inference separation:}
The global label identifier is never supplied as part of the query.
During supervised training, label identifiers are available only for
the designated target positions. At inference,
no gold label, target-derived identifier, or outcome information is
provided. All query and generated-output positions use a designated
null-label identifier $z_{\varnothing}$. Predictions are therefore
conditioned only on the instruction, demonstrations, historical
observations, current observation, and temporal information.
\section{Dataset Statistics}
This section summarizes key statistics of the datasets used for training and evaluation, including timeline structure, token distributions, and event sparsity \hyperref[tab:dataset_stats]{\underline{\textcolor{blue}{Table~\ref*{tab:dataset_stats}}}}, \hyperref[tab:dataset_overview_presentation]{\underline{\textcolor{blue}{Table~\ref*{tab:dataset_overview_presentation}}}} .

\begin{table*}[t]
\centering
\scriptsize
\setlength{\tabcolsep}{3.2pt}
\renewcommand{\arraystretch}{1.25}

\begin{tabularx}{\textwidth}{
    >{\raggedright\arraybackslash}p{1.35cm}
    >{\raggedright\arraybackslash}p{2.35cm}
    >{\raggedright\arraybackslash}X
    >{\raggedright\arraybackslash}p{2.25cm}
    >{\raggedright\arraybackslash}p{3.0cm}
    >{\centering\arraybackslash}p{1.35cm}
}
\toprule
\textbf{Dataset} &
\textbf{Domain and Sequence} &
\textbf{Task Definition} &
\textbf{Labels} &
\textbf{Key Characteristics} &
\textbf{Role} \\
\midrule

\rowcolor{green!5}
\textbf{AnnoMI} &
Motivational-interviewing dialogue &
Given the preceding therapist--client dialogue, classify the motivational
language expressed in the current client utterance. &
\textit{Change}, \textit{Sustain}, \textit{Neutral} &
44 topics; 6,725 utterances; structured and relatively short contexts
(mean: 142 tokens). &
Train  \\

\rowcolor{green!5}
\textbf{LRS} &
Twitter rumour timelines &
Given the chronological history of a rumour conversation, predict whether
the stance changes at the current post. &
\textit{Switch}, \textit{Non-Switch} &
274 timelines; 4,239 posts; medium-length, noisy social-media sequences
(mean: 216 tokens). &
Train  \\

\rowcolor{green!5}
\textbf{TalkLife} &
Mental-health user timelines &
Given a user's previous posts and current post, identify whether their
emotional or mental state changes or escalates. &
\textit{IS}, \textit{IE}, \textit{O} &
500 timelines; 18,702 posts; long contexts with sparse and often implicit
changes (mean: 1,049 tokens). &
Train  \\

\rowcolor{red!5}
\textbf{Reddit MoC} &
Reddit mental-health timelines &
Detect switches, escalations, or stability in a user's mental state across
chronologically ordered Reddit posts. &
\textit{IS}, \textit{IE}, \textit{O} &
255 timelines; 6,195 posts; longest and noisier user histories
(mean: 2,583 tokens). &
OOD only \\

\rowcolor{red!5}
\textbf{CMV} &
Reddit opinion-change discussions &
Given a reply and the author's previous same-topic contributions, predict
whether the reply changes the author's view. &
\textit{1: Change}, \textit{0: No change} &
9,456 author-topic timelines; 48,570 replies; binary and highly sparse
change events (mean: 786 tokens). &
OOD only \\

\bottomrule
\end{tabularx}

\caption{\textbf{Overview of the longitudinal datasets.}
Green rows indicate datasets used for LiFT training and in-domain (ID)
evaluation with an 80/20 split. Red rows indicate datasets used exclusively
for out-of-distribution (OOD) evaluation.}
\label{tab:dataset_overview_presentation}
\end{table*}
\begin{table}[!t]
\centering
\scriptsize
\setlength{\tabcolsep}{1.2pt}
\renewcommand{\arraystretch}{0.95}

\begin{tabular}{lccccc}
\toprule
\rowcolor{gray!8}
 & \textbf{LRS} & \textbf{TalkLife} & \textbf{AnnoMI} & \textbf{Reddit} & \textbf{CMV} \\
\midrule

\rowcolor{sectiongray}
\multicolumn{6}{c}{\textbf{Dataset Overview}} \\

\rowcolor{gray!8}
Split Type & timelines & timelines & topics & timelines & timeline \\
Total Timelines & 274 & 500 & 44 & 255 & 9,456 \\
\rowcolor{gray!8}
Train Timelines & 219 & 400 & 35 & -- & -- \\
Test Timelines & 55 & 100 & 9 & Full Set & Full Set \\
\rowcolor{gray!8}
Total Posts & 4,239 & 18,702 & 6,725 & 6,195 & 48,570 \\

\midrule
\rowcolor{sectiongray}
\multicolumn{6}{c}{\textbf{Timeline and Post Statistics}} \\

\rowcolor{gray!8}
Mean Posts per Timeline & 15.5 & 37.40 & 10 & 24.29 & 5.14 \\
Mean Tokens per Timeline & 216 & 1,049 & 142 & 2,583 & 786 \\
\rowcolor{gray!8}
Mean Tokens per Post & 14.0 & 22.2 & 15.52 & 106.3 & 153 \\

\midrule
\rowcolor{sectiongray}
\multicolumn{6}{c}{\textbf{Event Statistics}} \\

\rowcolor{gray!8}
Mean Event Frequency & 0.50 & 0.34 & 0.367 & 0.33 & 0.15 \\
Minority Events per Timeline & 6.5 & 2.9 & 4 & 5.5 & 0.15 \\
\rowcolor{gray!8}
Mean Posts Between Events & 0.32 & 1.99 & 0.73 & 1.97 & 34.6 \\
Mean Tokens Between Events & 3.70 & 63.27 & 42.32 & 207.89 & 5,307 \\

\bottomrule
\end{tabular}

\caption{
Comprehensive dataset statistics across all datasets.
}
\label{tab:dataset_stats}
\end{table}

\begin{table}[!t]
\centering
\scriptsize
\setlength{\tabcolsep}{3pt}
\renewcommand{\arraystretch}{1.1}

\begin{tabular}{llccc}
\toprule

\rowcolor{headergray}
\textbf{Model} & \textbf{Context} & \textbf{0} & \textbf{1} & \textbf{3} \\

\midrule

\multirow{6}{*}{\textbf{OLMo-7B}}
& IFT (Author-all) & .487 & .478 & .513 \\
& LiFT (Author-all) & \textbf{.543} & \textbf{.545} & \textbf{.564} \\

& IFT (Conv.) & .482 & .471 & .506 \\
& LiFT (Conv.) & \textbf{.488} & \textbf{.479} & \textbf{.514} \\

& IFT (Author-topic) & .482 & .470 & .486 \\
& LiFT (Author-topic) & \textbf{.601} & \textbf{.606} & \textbf{.618} \\

\midrule

\multirow{6}{*}{\textbf{Qwen-14B}}
& IFT (Author-all) & .512 & .524 & .538 \\
& LiFT (Author-all) & \textbf{.549} & \textbf{.562} & \textbf{.577} \\

& IFT (Conv.) & .506 & .517 & .531 \\
& LiFT (Conv.) & \textbf{.511} & \textbf{.523} & \textbf{.537} \\

& IFT (Author-topic) & .509 & .547 & .553 \\
& LiFT (Author-topic) & .479 & \textbf{.584} & \textbf{.624} \\

\bottomrule
\end{tabular}

\caption{Macro-F1 on CMV with different context sources.}
\label{tab:cmv_context_lift}

\end{table}

\section{Model Checkpoints}
\label{app:model_checkpoints}

We use the following publicly available pretrained model checkpoints:

\begin{itemize}[leftmargin=*,nosep]
    \item \textbf{allenai/OLMo-2-0425-1B-Instruct:}
    \url{allenai/OLMo-2-0425-1B-Instruct}

    \item \textbf{allenai/Olmo-3-7B-Instruct:}
    \url{https://huggingface.co/allenai/Olmo-3-7B-Instruct}
 \item \textbf{allenai/Olmo-3-1025-7B:}
    \url{https://huggingface.co/allenai/Olmo-3-1025-7B}

    \item \textbf{LLaMA-3.1-8B-Instruct:}
    \url{https://huggingface.co/meta-llama/Llama-3.1-8B-Instruct}

    \item \textbf{Qwen2.5-14B-Instruct:}
    \url{https://huggingface.co/Qwen/Qwen2.5-14B-Instruct}

    \item  \textbf{Qwen/Qwen2.5-32B-Instruct}
    \url{https://huggingface.co/Qwen/Qwen2.5-32B-Instruct}
\end{itemize}

\section{Figure-Level Aggregation}
\label{sec:fig}

\begin{figure*}[!h]
    \centering
    \includegraphics[
        width=\linewidth,
        trim=5 5 5 5,
        clip
    ]{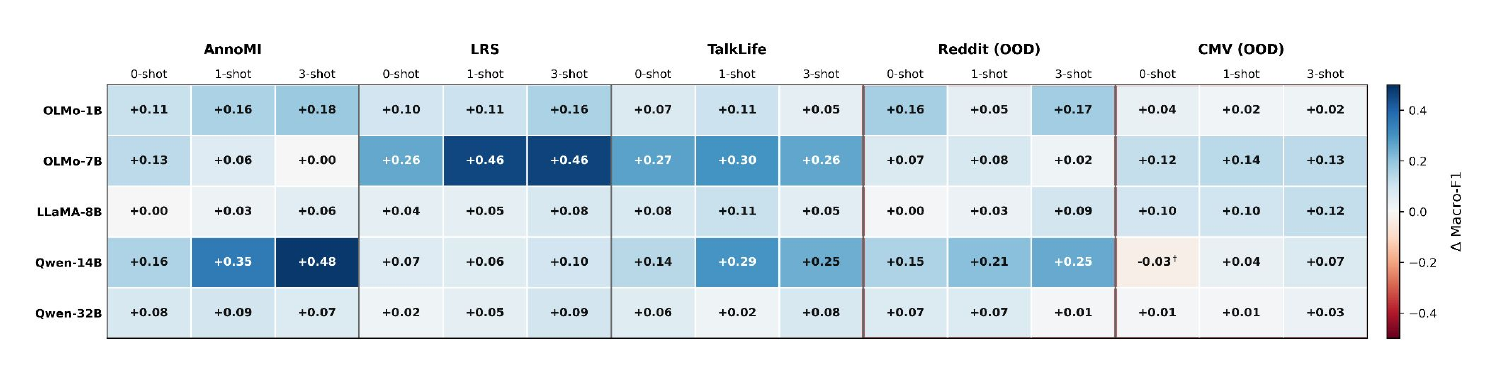}
\caption{Per-model Macro-F1 gains of LiFT over IFT across datasets and shot settings. 
Each cell reports $\Delta_{m,d,s}=F^{L}_{m,d,s}-F^{I}_{m,d,s}$, where positive values indicate improved performance with LiFT.}
 \label{fig:plot2}
\end{figure*}

The results used to construct the dataset-level curves in
\hyperref[fig:plot1]{\underline{\textcolor{blue}{Figure~\ref*{fig:plot1}}}}
are derived from the Macro-F1 scores reported in
\hyperref[tab:sota_comparison]{\underline{\textcolor{blue}{Table~\ref*{tab:sota_comparison}}}}.
The resulting aggregated values are reported in
\hyperref[tab:aggregation]{\underline{\textcolor{blue}{Table~\ref*{tab:aggregation}}}}.
For each dataset \(d\) and shot setting \(s \in \{0,1,3\}\), we average the corresponding IFT and LiFT scores across the \(M=5\) model families:

\begin{equation}
\scriptsize
\overline{F}^{\,I}_{d,s}
=
\frac{1}{M}
\sum_{m=1}^{M}
F^{I}_{m,d,s},
\label{eq:dataset_ift_mean}
\end{equation}
\begin{equation}
\scriptsize
\overline{F}^{\,L}_{d,s}
=
\frac{1}{M}
\sum_{m=1}^{M}
F^{L}_{m,d,s},
\label{eq:dataset_lift_mean}
\end{equation}

where \(F^{I}_{m,d,s}\) and \(F^{L}_{m,d,s}\) denote the Macro-F1 scores reported in \hyperref[tab:sota_comparison]{\underline{\textcolor{blue}{Table~\ref*{tab:sota_comparison}}}}
for the IFT and LiFT variants of model \(m\), respectively. The performance gain of LiFT over its corresponding IFT model is
computed for each model, dataset, and shot setting as:

\begin{equation}
\scriptsize
\Delta_{m,d,s}
=
F^{L}_{m,d,s}
-
F^{I}_{m,d,s}.
\label{eq:model_gain}
\end{equation}
These per-model gains are visualised in
\hyperref[fig:plot2]{\underline{\textcolor{blue}{Figure~\ref*{fig:plot2}}}}.
Positive values indicate that LiFT outperforms IFT, whereas negative values indicate a decrease. The dataset-level gain annotated in the line plots of
\hyperref[fig:plot1]{\underline{\textcolor{blue}{Figure~\ref*{fig:plot1}}}} is:

\begin{equation}
\scriptsize
\Delta_{d,s}
=
\overline{F}^{\,L}_{d,s}
-
\overline{F}^{\,I}_{d,s}.
\label{eq:dataset_gain}
\end{equation}

For the overall panel, scores are averaged across all \(M=5\) model families and \(D=5\) datasets:

\begin{equation}
\scriptsize
\overline{F}^{\,I}_{s}
=
\frac{1}{MD}
\sum_{m=1}^{M}
\sum_{d=1}^{D}
F^{I}_{m,d,s},
\label{eq:overall_ift_mean}
\end{equation}

\begin{equation}
\scriptsize
\overline{F}^{\,L}_{s}
=
\frac{1}{MD}
\sum_{m=1}^{M}
\sum_{d=1}^{D}
F^{L}_{m,d,s}.
\label{eq:overall_lift_mean}
\end{equation}
The corresponding overall LiFT gain is:
\begin{equation}
\scriptsize
\Delta_s
=
\overline{F}^{\,L}_{s}
-
\overline{F}^{\,I}_{s}.
\label{eq:overall_gain}
\end{equation}

\begin{table*}[!t]
\centering
\scriptsize
\setlength{\tabcolsep}{4.2pt}
\renewcommand{\arraystretch}{1.08}

\resizebox{\textwidth}{!}{%
\begin{tabular}{l|ccc|ccc|ccc}
\toprule
& \multicolumn{3}{c|}{\textbf{IFT}}
& \multicolumn{3}{c|}{\textbf{LiFT}}
& \multicolumn{3}{c}{\textbf{Gain}} \\
\cmidrule(lr){2-4}
\cmidrule(lr){5-7}
\cmidrule(lr){8-10}

\textbf{Dataset}
& \textbf{0-shot}
& \textbf{1-shot}
& \textbf{3-shot}
& \textbf{0-shot}
& \textbf{1-shot}
& \textbf{3-shot}
& \textbf{0-shot}
& \textbf{1-shot}
& \textbf{3-shot} \\
\midrule

\rowcolor{green!8}
AnnoMI
& .292 & .341 & .357
& .389 & .479 & .517
& +.096 & +.138 & +.161 \\

\rowcolor{green!8}
LRS
& .332 & .342 & .327
& .428 & .486 & .504
& +.096 & +.144 & +.178 \\

\rowcolor{green!8}
TalkLife
& .158 & .125 & .186
& .267 & .277 & .318
& +.110 & +.152 & +.132 \\

\rowcolor{red!8}
Reddit (OOD)
& .197 & .252 & .251
& .289 & .339 & .361
& +.092 & +.087 & +.109 \\

\rowcolor{red!8}
CMV (OOD)
& .494 & .509 & .516
& .540 & .569 & .590
& +.047 & +.060 & +.074 \\

\midrule
\rowcolor{gray!12}
\textbf{Overall}
& \textbf{.295}
& \textbf{.314}
& \textbf{.327}
& \textbf{.383}
& \textbf{.430}
& \textbf{.458}
& \textbf{+.088}
& \textbf{+.116}
& \textbf{+.131} \\

\bottomrule
\end{tabular}%
}

\caption{\textbf{Aggregated Macro-F1 values used to construct
Figure~\ref{fig:plot2}.}
For each dataset and shot setting, IFT and LiFT scores are
averaged across the five model families reported in
Table~\ref{tab:sota_comparison}.
The gain is computed as
$\Delta_{d,s}=F^{\mathrm{LiFT}}_{d,s}
-F^{\mathrm{IFT}}_{d,s}$.}
\label{tab:aggregation}
\end{table*}

\section{Detailed Results}
\label{sec:appendix_results}

The macro-F1 trends for each model are
visualized in 
\hyperref[fig:macro_f1_all_models_onepage]{\underline{\textcolor{blue}{Figure~\ref*{fig:macro_f1_all_models_onepage}}}}. 
% \hyperref[tab:combined_f1_all_models]{\underline{\textcolor{blue}{Table~\ref*{tab:combined_f1_all_models}}}}
% present per-class precision, recall, and F1 scores along with macro-F1 across
% 0-, 1-, and 3-shot settings. 

\subsection{Incremental In-Context Learning Improvement}
\label{sec:icl_amplification}

To distinguish general fine-tuning gains from improved use of
in-context demonstrations, we measure the incremental change in
Macro-F1 from zero- to one-shot prompting and from one- to three-shot
prompting. For variant $v\in\{I,L\}$, where $I$ and $L$ denote IFT
and LiFT, respectively, the improvement over each transition is

\begin{equation}
\scriptsize
\Delta^{v}_{0\rightarrow1}
=
F^{v}_{1}-F^{v}_{0},
\qquad
\Delta^{v}_{1\rightarrow3}
=
F^{v}_{3}-F^{v}_{1},
\label{eq:incremental_icl}
\end{equation}

where $F^{v}_{k}$ is the Macro-F1 obtained by variant $v$ with $k$
demonstrations. We define LiFT's additional incremental benefit over
IFT as

\begin{equation}
\scriptsize
\operatorname{Amp}_{a\rightarrow b}
=
\left(F^{L}_{b}-F^{L}_{a}\right)
-
\left(F^{I}_{b}-F^{I}_{a}\right),
\label{eq:icl_amplification}
\end{equation}

where $(a,b)\in\{(0,1),(1,3)\}$. Positive amplification indicates
that LiFT benefits more than IFT from the added demonstrations,
separating improved demonstration utilisation from a fixed gain
introduced by fine-tuning.

As shown in
\hyperref[tab:icl_amplification]{\underline{\textcolor{blue}{Table~\ref*{tab:icl_amplification}}}},
LiFT achieves an average amplification of $+.028$ from zero to one
demonstration. Amplification is positive for OLMo-7B, LLaMA-8B, and
Qwen-14B, while OLMo-1B shows a small negative value and Qwen-32B is
approximately neutral ($-.001$). Qwen-14B exhibits the strongest
amplification ($+.094$), followed by OLMo-7B ($+.037$), indicating
that LiFT particularly improves these models' use of the first
demonstration.

From one to three demonstrations, the average amplification is
smaller but remains positive at $+.014$. OLMo-1B, LLaMA-8B,
Qwen-14B, and Qwen-32B retain positive amplification, whereas
OLMo-7B exhibits negative amplification. Overall, LiFT most strongly
improves the initial use of in-context supervision, while the benefit
of further demonstrations varies across model families and shot
transitions.
\begin{table}[!t] \centering \scriptsize \setlength{\tabcolsep}{3.4pt} \renewcommand{\arraystretch}{1.05} \begin{tabular}{l|rrr|rrr} \toprule \rowcolor{gray!8} & \multicolumn{3}{c|}{$0\rightarrow1$ shot} & \multicolumn{3}{c}{$1\rightarrow3$ shots} \\ \rowcolor{gray!8} \textbf{Model} & $\Delta^{I}$ & $\Delta^{L}$ & \textbf{Amp.} & $\Delta^{I}$ & $\Delta^{L}$ & \textbf{Amp.} \\ \midrule \rowcolor{orange!8} OLMo-1B & $+.027$ & $+.018$ & $-.009$ & $-.006$ & $+.034$ & \textbf{$+.040$} \\ \rowcolor{orange!8} OLMo-7B & $+.027$ & $+.064$ & \textbf{$+.037$} & $+.043$ & $+.012$ & $-.031$ \\ \rowcolor{orange!8} LLaMA-8B & $+.035$ & $+.054$ & \textbf{$+.019$} & $+.024$ & $+.041$ & \textbf{$+.018$} \\ \rowcolor{orange!8} Qwen-14B & $-.022$ & $+.073$ & \textbf{$+.094$} & $-.010$ & $+.028$ & \textbf{$+.038$} \\ \rowcolor{orange!8} Qwen-32B & $+.028$ & $+.027$ & $-.001$ & $+.018$ & $+.026$ & \textbf{$+.008$} \\ \midrule \rowcolor{teal!8} \textbf{Overall} & \textbf{$+.019$} & \textbf{$+.047$} & \textbf{$+.028$} & \textbf{$+.014$} & \textbf{$+.028$} & \textbf{$+.014$} \\ \bottomrule \end{tabular} \caption{\textit{Incremental in-context learning improvement of LiFT relative to IFT.} $\Delta^{I}$ and $\Delta^{L}$ denote the changes in Macro-F1 for IFT and LiFT, respectively, when moving from zero to one demonstration and from one to three demonstrations. ICL amplification is defined as $\operatorname{Amp}=\Delta^{L}-\Delta^{I}$. A positive value indicates that LiFT obtains a greater incremental benefit from additional demonstrations than IFT. Values are averaged across the five datasets.} \label{tab:icl_amplification} \end{table}g

\subsection{Additional Analysis with 5, 7, and 9 Shots}
\label{sec:additionalanalysis}
\begin{figure}[t]
    \centering
    \includegraphics[
        width=\columnwidth,
        trim=10 6 8 10,
        clip
    ]{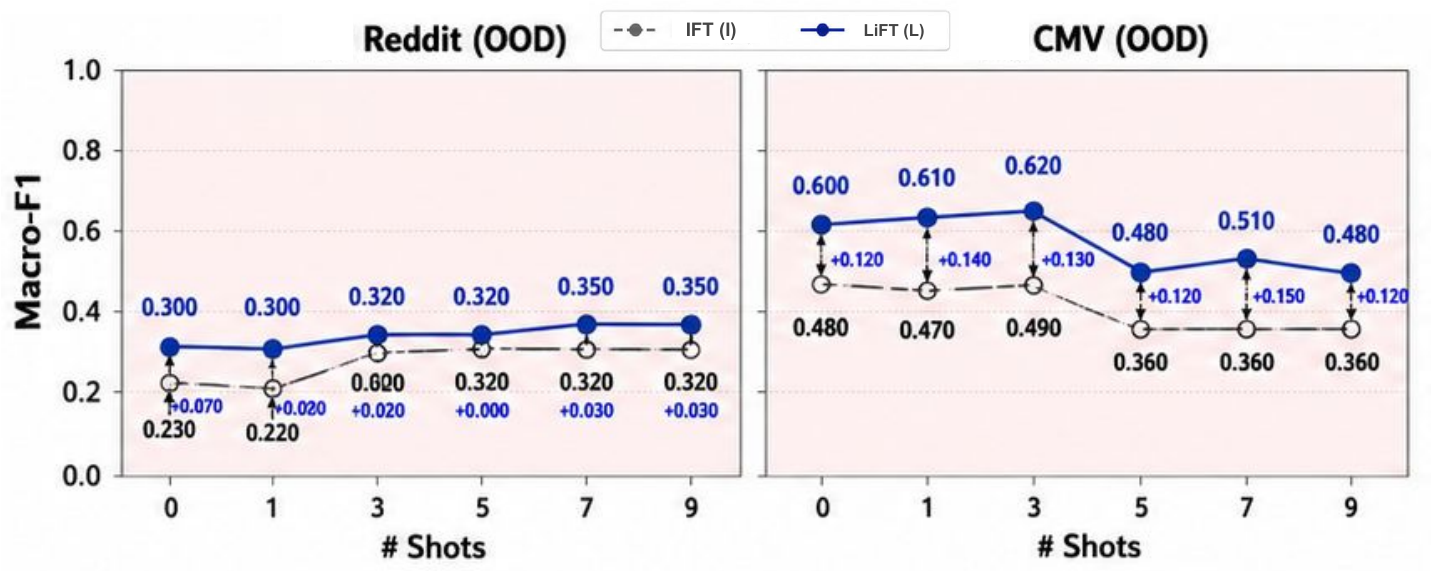}
    \caption{\textbf{OLMo-7B under longer OOD contexts.}
Macro-F1 of IFT and LiFT on Reddit and CMV across 0--9 shots. LiFT consistently outperforms Base on both datasets. On Reddit, performance remains stable and improves slightly at 7--9 shots, whereas on CMV both methods decline beyond 3 shots, but LiFT retains a clear advantage.}
    \label{fig:plot5}
\end{figure}

\hyperref[fig:plot3]{\underline{\textcolor{blue}{Figure~\ref*{fig:plot3}}}} 
and \hyperref[fig:plot5]{\underline{\textcolor{blue}{Figure~\ref*{fig:plot5}}}} extend the OOD evaluation to longer in-context settings for OLMo-7B and Qwen-14B, respectively. Overall, increasing the number of demonstrations does not consistently benefit the IFT models, whereas LiFT is substantially more robust to longer contexts.

For Qwen-14B (\hyperref[fig:plot3]{\underline{\textcolor{blue}{Figure~\ref*{fig:plot3}}}}), LiFT consistently outperforms IFT on Reddit, increasing from $0.26$ at zero-shot to $0.38$--$0.39$ with additional demonstrations. In contrast, IFT peaks at $0.17$ in the one-shot setting and declines to $0.06$--$0.07$ at 5--9 shots. On CMV, LiFT improves from $0.48$ to $0.68$ at five shots and remains stable at $0.67$ with seven and nine shots. IFT reaches $0.55$ at one and three shots but drops sharply to approximately $0.35$ thereafter.

A similar, although less pronounced, pattern is observed for OLMo-7B (\hyperref[fig:plot5]{\underline{\textcolor{blue}{Figure~\ref*{fig:plot5}}}} ). On Reddit, LiFT remains stable or improves slightly at longer shot settings, reaching $0.35$ at seven and nine shots compared with $0.32$ for IFT. On CMV, both methods decline after three shots; however, LiFT maintains a clear advantage, achieving $0.48$--$0.51$ at 5--9 shots compared with $0.36$ for IFT. These results suggest that simply adding more demonstrations can introduce saturation or interference, particularly for the IFT models, while LiFT better preserves useful information as the context grows.

\begin{figure*}[p]
\centering
\scriptsize

\pgfplotsset{
  myaxis/.style={
    width=\linewidth,
    height=2.55cm,
    xlabel={Shots},
    ylabel={Macro-F1},
    xmin=-0.2, xmax=3.2,
    xtick={0,1,3},
    grid=both,
    tick label style={font=\scriptsize},
    label style={font=\scriptsize},
    title style={font=\scriptsize},
    legend style={at={(0.5,1.22)},anchor=south,legend columns=2,font=\scriptsize}
  }
}

% =========================================================
% Row 1: OLMo-1B
% =========================================================
\textbf{OLMo-1B}\par\vspace{-2pt}

\begin{subfigure}{0.19\textwidth}\centering
\begin{tikzpicture}\begin{axis}[myaxis, title={Reddit}, ymin=0.05, ymax=0.35]
\addplot+[mark=o, thick] coordinates {(0,0.136) (1,0.268) (3,0.150)};
\addplot+[mark=square, thick] coordinates {(0,0.301) (1,0.317) (3,0.323)};
\legend{IFT, LiFT}
\end{axis}\end{tikzpicture}
\end{subfigure}\hfill
\begin{subfigure}{0.19\textwidth}\centering
\begin{tikzpicture}\begin{axis}[myaxis, title={TalkLife}, ymin=0.20, ymax=0.36]
\addplot+[mark=o, thick] coordinates {(0,0.225) (1,0.217) (3,0.290)};
\addplot+[mark=square, thick] coordinates {(0,0.296) (1,0.323) (3,0.340)};
\end{axis}\end{tikzpicture}
\end{subfigure}\hfill
\begin{subfigure}{0.19\textwidth}\centering
\begin{tikzpicture}\begin{axis}[myaxis, title={LRS}, ymin=0.34, ymax=0.55]
% OLMo-1B LRS IFT
\addplot+[mark=o, thick] coordinates {(0,0.354) (1,0.364) (3,0.371)};
\addplot+[mark=square, thick] coordinates {(0,0.449) (1,0.473) (3,0.528)};
\end{axis}\end{tikzpicture}
\end{subfigure}\hfill
\begin{subfigure}{0.19\textwidth}\centering
\begin{tikzpicture}\begin{axis}[myaxis, title={AnnoMI}, ymin=0.16, ymax=0.38]
\addplot+[mark=o, thick] coordinates {(0,0.178) (1,0.178) (3,0.178)};
\addplot+[mark=square, thick] coordinates {(0,0.290) (1,0.335) (3,0.363)};
\end{axis}\end{tikzpicture}
\end{subfigure}\hfill
\begin{subfigure}{0.19\textwidth}\centering
\begin{tikzpicture}\begin{axis}[myaxis, title={CMV}, ymin=0.35, ymax=0.42]
\addplot+[mark=o, thick] coordinates {(0,0.361) (1,0.362) (3,0.369)};
\addplot+[mark=square, thick] coordinates {(0,0.400) (1,0.377) (3,0.389)};
\end{axis}\end{tikzpicture}
\end{subfigure}

\vspace{6pt}

% =========================================================
% Row 2: OLMo-7B
% =========================================================
\textbf{OLMo-7B}\par\vspace{-2pt}

\begin{subfigure}{0.19\textwidth}\centering
\begin{tikzpicture}\begin{axis}[myaxis, title={Reddit}, ymin=0.20, ymax=0.34]
\addplot+[mark=o, thick] coordinates {(0,0.227) (1,0.219) (3,0.298)};
\addplot+[mark=square, thick] coordinates {(0,0.298) (1,0.300) (3,0.319)};
\legend{IFT, LiFT}
\end{axis}\end{tikzpicture}
\end{subfigure}\hfill
\begin{subfigure}{0.19\textwidth}\centering
\begin{tikzpicture}\begin{axis}[myaxis, title={TalkLife}, ymin=0, ymax=0.35]
\addplot+[mark=o, thick] coordinates {(0,0.029) (1,0.008) (3,0.064)};
\addplot+[mark=square, thick] coordinates {(0,0.298) (1,0.305) (3,0.324)};
\end{axis}\end{tikzpicture}
\end{subfigure}\hfill
\begin{subfigure}{0.19\textwidth}\centering
\begin{tikzpicture}\begin{axis}[myaxis, title={LRS}, ymin=0.10, ymax=0.60]
\addplot+[mark=o, thick] coordinates {(0,0.124) (1,0.128) (3,0.125)};
\addplot+[mark=square, thick] coordinates {(0,0.380) (1,0.584) (3,0.587)};
\end{axis}\end{tikzpicture}
\end{subfigure}\hfill
\begin{subfigure}{0.19\textwidth}\centering
\begin{tikzpicture}\begin{axis}[myaxis, title={AnnoMI}, ymin=0.20, ymax=0.46]
% OLMo-7B AnnoMI IFT
\addplot+[mark=o, thick] coordinates {(0,0.207) (1,0.379) (3,0.446)};
\addplot+[mark=square, thick] coordinates {(0,0.339) (1,0.443) (3,0.448)};
\end{axis}\end{tikzpicture}
\end{subfigure}\hfill
\begin{subfigure}{0.19\textwidth}\centering
\begin{tikzpicture}\begin{axis}[myaxis, title={CMV}, ymin=0.46, ymax=0.63]
\addplot+[mark=o, thick] coordinates {(0,0.482) (1,0.470) (3,0.486)};
\addplot+[mark=square, thick] coordinates {(0,0.601) (1,0.606) (3,0.618)};
\end{axis}\end{tikzpicture}
\end{subfigure}

\vspace{6pt}

% =========================================================
% Row 3: LLaMA-8B
% =========================================================
\textbf{LLaMA-8B}\par\vspace{-2pt}

\begin{subfigure}{0.19\textwidth}\centering
\begin{tikzpicture}\begin{axis}[myaxis, title={Reddit}, ymin=0.28, ymax=0.40]
\addplot+[mark=o, thick] coordinates {(0,0.291) (1,0.291) (3,0.291)};
\addplot+[mark=square, thick] coordinates {(0,0.296) (1,0.320) (3,0.384)};
\legend{IFT, LiFT}
\end{axis}\end{tikzpicture}
\end{subfigure}\hfill
\begin{subfigure}{0.19\textwidth}\centering
\begin{tikzpicture}\begin{axis}[myaxis, title={TalkLife}, ymin=0.05, ymax=0.24]
\addplot+[mark=o, thick] coordinates {(0,0.067) (1,0.073) (3,0.180)};
\addplot+[mark=square, thick] coordinates {(0,0.145) (1,0.183) (3,0.231)};
\end{axis}\end{tikzpicture}
\end{subfigure}\hfill
\begin{subfigure}{0.19\textwidth}\centering
\begin{tikzpicture}\begin{axis}[myaxis, title={LRS}, ymin=0.42, ymax=0.52]
\addplot+[mark=o, thick] coordinates {(0,0.437) (1,0.460) (3,0.430)};
\addplot+[mark=square, thick] coordinates {(0,0.480) (1,0.507) (3,0.511)};
\end{axis}\end{tikzpicture}
\end{subfigure}\hfill
\begin{subfigure}{0.19\textwidth}\centering
\begin{tikzpicture}\begin{axis}[myaxis, title={AnnoMI}, ymin=0.20, ymax=0.46]
\addplot+[mark=o, thick] coordinates {(0,0.207) (1,0.352) (3,0.400)};
% LLaMA-8B AnnoMI LiFT
\addplot+[mark=square, thick] coordinates {(0,0.207) (1,0.379) (3,0.457)};
\end{axis}\end{tikzpicture}
\end{subfigure}\hfill
\begin{subfigure}{0.19\textwidth}\centering
\begin{tikzpicture}\begin{axis}[myaxis, title={CMV}, ymin=0.45, ymax=0.59]
\addplot+[mark=o, thick] coordinates {(0,0.466) (1,0.468) (3,0.462)};
\addplot+[mark=square, thick] coordinates {(0,0.561) (1,0.568) (3,0.581)};
\end{axis}\end{tikzpicture}
\end{subfigure}

\vspace{6pt}

% =========================================================
% Row 4: Qwen-14B
% =========================================================
\textbf{Qwen-14B}\par\vspace{-2pt}

\begin{subfigure}{0.19\textwidth}\centering
\begin{tikzpicture}\begin{axis}[myaxis, title={Reddit}, ymin=0.05, ymax=0.40]
% Qwen-14B Reddit IFT
\addplot+[mark=o, thick] coordinates {(0,0.107) (1,0.167) (3,0.127)};
\addplot+[mark=square, thick] coordinates {(0,0.260) (1,0.376) (3,0.377)};
\legend{IFT, LiFT}
\end{axis}\end{tikzpicture}
\end{subfigure}\hfill
\begin{subfigure}{0.19\textwidth}\centering
\begin{tikzpicture}\begin{axis}[myaxis, title={TalkLife}, ymin=0, ymax=0.35]
\addplot+[mark=o, thick] coordinates {(0,0.177) (1,0.027) (3,0.090)};
\addplot+[mark=square, thick] coordinates {(0,0.315) (1,0.321) (3,0.335)};
\end{axis}\end{tikzpicture}
\end{subfigure}\hfill
\begin{subfigure}{0.19\textwidth}\centering
\begin{tikzpicture}\begin{axis}[myaxis, title={LRS}, ymin=0.36, ymax=0.52]
\addplot+[mark=o, thick] coordinates {(0,0.384) (1,0.414) (3,0.400)};
% Qwen-14B LRS LiFT
\addplot+[mark=square, thick] coordinates {(0,0.450) (1,0.474) (3,0.496)};
\end{axis}\end{tikzpicture}
\end{subfigure}\hfill
\begin{subfigure}{0.19\textwidth}\centering
\begin{tikzpicture}\begin{axis}[myaxis, title={AnnoMI}, ymin=0.20, ymax=0.70]
\addplot+[mark=o, thick] coordinates {(0,0.357) (1,0.271) (3,0.204)};
\addplot+[mark=square, thick] coordinates {(0,0.513) (1,0.626) (3,0.687)};
\end{axis}\end{tikzpicture}
\end{subfigure}\hfill
\begin{subfigure}{0.19\textwidth}\centering
\begin{tikzpicture}\begin{axis}[myaxis, title={CMV}, ymin=0.47, ymax=0.65]
\addplot+[mark=o, thick] coordinates {(0,0.509) (1,0.547) (3,0.553)};
\addplot+[mark=square, thick] coordinates {(0,0.479) (1,0.584) (3,0.624)};
\end{axis}\end{tikzpicture}
\end{subfigure}
\vspace{6pt}

% =========================================================
% Row 5: Qwen-32B
% =========================================================
\textbf{Qwen-32B}\par\vspace{-2pt}

\begin{subfigure}{0.19\textwidth}\centering
\begin{tikzpicture}
\begin{axis}[
    myaxis,
    title={Reddit},
    ymin=0.20,
    ymax=0.42
]
\addplot+[mark=o, thick]
coordinates {(0,0.22) (1,0.31) (3,0.39)};
\addplot+[mark=square, thick]
coordinates {(0,0.29) (1,0.38) (3,0.40)};
\legend{IFT, LiFT}
\end{axis}
\end{tikzpicture}
\end{subfigure}\hfill
\begin{subfigure}{0.19\textwidth}\centering
\begin{tikzpicture}
\begin{axis}[
    myaxis,
    title={TalkLife},
    ymin=0.28,
    ymax=0.40
]
\addplot+[mark=o, thick]
coordinates {(0,0.29) (1,0.30) (3,0.30)};
\addplot+[mark=square, thick]
coordinates {(0,0.35) (1,0.32) (3,0.38)};
\end{axis}
\end{tikzpicture}
\end{subfigure}\hfill
\begin{subfigure}{0.19\textwidth}\centering
\begin{tikzpicture}
\begin{axis}[
    myaxis,
    title={LRS},
    ymin=0.29,
    ymax=0.42
]
\addplot+[mark=o, thick]
coordinates {(0,0.36) (1,0.34) (3,0.31)};
\addplot+[mark=square, thick]
coordinates {(0,0.38) (1,0.39) (3,0.40)};
\end{axis}
\end{tikzpicture}
\end{subfigure}\hfill
\begin{subfigure}{0.19\textwidth}\centering
\begin{tikzpicture}
\begin{axis}[
    myaxis,
    title={AnnoMI},
    ymin=0.48,
    ymax=0.65
]
\addplot+[mark=o, thick]
coordinates {(0,0.51) (1,0.52) (3,0.56)};
\addplot+[mark=square, thick]
coordinates {(0,0.59) (1,0.61) (3,0.63)};
\end{axis}
\end{tikzpicture}
\end{subfigure}\hfill
\begin{subfigure}{0.19\textwidth}\centering
\begin{tikzpicture}
\begin{axis}[
    myaxis,
    title={CMV},
    ymin=0.64,
    ymax=0.76
]
\addplot+[mark=o, thick]
coordinates {(0,0.65) (1,0.70) (3,0.71)};
\addplot+[mark=square, thick]
coordinates {(0,0.66) (1,0.71) (3,0.74)};
\end{axis}
\end{tikzpicture}
\end{subfigure}

\caption{Macro-F1 for IFT and LiFT models across 0-, 1-, and
3-shot settings on Reddit, TalkLife, LRS, AnnoMI, and CMV for five
backbones: OLMo-1B, OLMo-7B, LLaMA-8B, Qwen-14B, and Qwen-32B.}
\label{fig:macro_f1_all_models_onepage}
\hrule

\vspace{5pt}
% =========================================================
% Compact OLMo-7B ablation on the same float page
% =========================================================
\tikzset{
  basebar/.style={draw=blue!75!black,fill=blue!35},
  sftbar/.style={draw=red!75!black,fill=red!30},
  temporalbar/.style={draw=purple!75!black,fill=purple!45},
  curriculumbar/.style={draw=green!65!black,fill=green!50},
  reverseonebar/.style={draw=cyan!65!black,fill=cyan!45},
  reversethreebar/.style={draw=magenta!70!black,fill=magenta!35},
  mixedbar/.style={draw=black!75,fill=gray!55},
  historybar/.style={draw=yellow!60!black,fill=yellow!70},
  fullbar/.style={draw=orange!85!black,fill=orange!35}
}

\newcommand{\ablationlegenditem}[2]{%
  \tikz[baseline=-0.45ex]{%
    \draw[#1] (0,0) rectangle (0.27,0.10);
  }%
  \hspace{1pt}#2%
}

\newcommand{\addablationbars}[9]{%
  \addplot[basebar,bar shift=-16pt]
    coordinates {(OLMo-7B,#1)};
  \addplot[sftbar,bar shift=-12pt]
    coordinates {(OLMo-7B,#2)};
  \addplot[temporalbar,bar shift=-8pt]
    coordinates {(OLMo-7B,#3)};
  \addplot[curriculumbar,bar shift=-4pt]
    coordinates {(OLMo-7B,#4)};
  \addplot[reverseonebar,bar shift=0pt]
    coordinates {(OLMo-7B,#5)};
  \addplot[reversethreebar,bar shift=4pt]
    coordinates {(OLMo-7B,#6)};
  \addplot[mixedbar,bar shift=8pt]
    coordinates {(OLMo-7B,#7)};
  \addplot[historybar,bar shift=12pt]
    coordinates {(OLMo-7B,#8)};
  \addplot[fullbar,bar shift=16pt]
    coordinates {(OLMo-7B,#9)};
}

\resizebox{0.98\textwidth}{!}{%
\begin{tikzpicture}

\begin{groupplot}[
  group style={
    group size=5 by 1,
    horizontal sep=0.65cm
  },
  ybar,
  width=4.45cm,
  height=3.15cm,
  /pgf/bar width=2.6pt,
  enlarge x limits=0.68,
  symbolic x coords={OLMo-7B},
  xtick=data,
  xticklabel style={font=\scriptsize},
  yticklabel style={
    font=\scriptsize,
    /pgf/number format/fixed,
    /pgf/number format/precision=2
  },
  ylabel style={font=\scriptsize},
  title style={font=\scriptsize},
  tick align=outside,
  scaled y ticks=false
]

% Reddit
\nextgroupplot[
  title={Reddit},
  ylabel={Avg. Macro-F1},
  ymin=0.20,
  ymax=0.33,
  ytick={0.20,0.24,0.28,0.32}
]
\addablationbars
  {0.2733}{0.2633}{0.2667}
  {0.2633}{0.2733}{0.2700}
  {0.2467}{0.2667}{0.3067}

% TalkLife
\nextgroupplot[
  title={TalkLife},
  ymin=0,
  ymax=0.34,
  ytick={0,0.10,0.20,0.30}
]
\addablationbars
  {0.0333}{0.0600}{0.0567}
  {0.0567}{0.0633}{0.0600}
  {0.0500}{0.0567}{0.3100}

% LRS
\nextgroupplot[
  title={LRS},
  ymin=0.10,
  ymax=0.55,
  ytick={0.10,0.20,0.30,0.40,0.50}
]
\addablationbars
  {0.1600}{0.3533}{0.3467}
  {0.3467}{0.3600}{0.3533}
  {0.3233}{0.3433}{0.5267}

% AnnoMI
\nextgroupplot[
  title={AnnoMI},
  ymin=0.28,
  ymax=0.43,
  ytick={0.28,0.32,0.36,0.40}
]
\addablationbars
  {0.3467}{0.3100}{0.3100}
  {0.3100}{0.3200}{0.3167}
  {0.2967}{0.3100}{0.4100}

% CMV
\nextgroupplot[
  title={CMV},
  ymin=0.46,
  ymax=0.63,
  ytick={0.46,0.50,0.54,0.58,0.62}
]
\addablationbars
  {0.4800}{0.4933}{0.4867}
  {0.4933}{0.5100}{0.5033}
  {0.4733}{0.5000}{0.6100}

\end{groupplot}
\end{tikzpicture}%
}

\vspace{4pt}

{\tiny
\setlength{\tabcolsep}{2.8pt}
\renewcommand{\arraystretch}{1.02}

\begin{tabular}{ccccc}
\ablationlegenditem{basebar}{IFT}
&
\ablationlegenditem{sftbar}{SFT}
&
\ablationlegenditem{temporalbar}{w/o Temporal}
&
\ablationlegenditem{curriculumbar}{w/o Curriculum}
&
\ablationlegenditem{reverseonebar}{Reverse 1--2--3}
\\
\ablationlegenditem{reversethreebar}{Reverse 3--2--1}
&
\ablationlegenditem{mixedbar}{Mixed-task}
&
\ablationlegenditem{historybar}{w/o History}
&
\ablationlegenditem{fullbar}{LiFT Full}
&
{}
\end{tabular}
}

\vspace{1pt}

\caption{\textit{OLMo-7B ablation.}
Macro-F1 averaged across 0-, 1-, and 3-shot settings.
Full LiFT outperforms component removals, reversed curricula,
and mixed-task training.}
\label{fig:ablation_olmo7b}
\vspace{5pt}
\hrule

% =========================================================
% Compact cross-backbone ablation
% =========================================================
\begin{minipage}{\textwidth}
\centering

\resizebox{0.99\textwidth}{!}{%
\begin{tikzpicture}

\begin{groupplot}[
  group style={
    group size=5 by 1,
    horizontal sep=0.55cm
  },
  ybar,
  width=4.55cm,
  height=2.55cm,
  /pgf/bar width=2.2pt,
  enlarge x limits=0.18,
  symbolic x coords={
    OLMo-1B,
    LLaMA-8B,
    Qwen-14B,
    Qwen-32B
  },
  xtick=data,
  xticklabels={
    {\shortstack{OLMo\\1B}},
    {\shortstack{LLaMA\\8B}},
    {\shortstack{Qwen\\14B}},
    {\shortstack{Qwen\\32B}}
  },
  xticklabel style={
    font=\tiny,
    align=center,
    anchor=north,
    yshift=-1pt,
    inner sep=0pt
  },
  yticklabel style={
    font=\tiny,
    /pgf/number format/fixed,
    /pgf/number format/precision=2
  },
  ylabel style={font=\tiny},
  title style={font=\scriptsize},
  axis line style={line width=0.35pt},
  tick align=outside,
  scaled y ticks=false
]

% =========================================================
% Reddit
% =========================================================
\nextgroupplot[
  title={Reddit},
  ylabel={Avg. Macro-F1},
  ymin=0,
  ymax=0.40,
  ytick={0,0.2,0.4}
]

\addplot[basebar,bar shift=-7.5pt] coordinates {
  (OLMo-1B,0.1323)
  (LLaMA-8B,0.2910)
  (Qwen-14B,0.1247)
  (Qwen-32B,0.3067)
};

\addplot[sftbar,bar shift=-4.5pt] coordinates {
  (OLMo-1B,0.2380)
  (LLaMA-8B,0.2927)
  (Qwen-14B,0.2833)
  (Qwen-32B,0.3260)
};

\addplot[fullbar,bar shift=-1.5pt] coordinates {
  (OLMo-1B,0.3137)
  (LLaMA-8B,0.3333)
  (Qwen-14B,0.3377)
  (Qwen-32B,0.3567)
};

\addplot[temporalbar,bar shift=1.5pt] coordinates {
  (OLMo-1B,0.2387)
  (LLaMA-8B,0.2903)
  (Qwen-14B,0.2807)
  (Qwen-32B,0.3277)
};

\addplot[curriculumbar,bar shift=4.5pt] coordinates {
  (OLMo-1B,0.2383)
  (LLaMA-8B,0.2900)
  (Qwen-14B,0.2803)
  (Qwen-32B,0.3260)
};

\addplot[historybar,bar shift=7.5pt] coordinates {
  (OLMo-1B,0.2403)
  (LLaMA-8B,0.2920)
  (Qwen-14B,0.2823)
  (Qwen-32B,0.3273)
};

% =========================================================
% TalkLife
% =========================================================
\nextgroupplot[
  title={TalkLife},
  ymin=0,
  ymax=0.37,
  ytick={0,0.15,0.30}
]

\addplot[basebar,bar shift=-7.5pt] coordinates {
  (OLMo-1B,0.2440)
  (LLaMA-8B,0.1067)
  (Qwen-14B,0.0980)
  (Qwen-32B,0.2967)
};

\addplot[sftbar,bar shift=-4.5pt] coordinates {
  (OLMo-1B,0.0457)
  (LLaMA-8B,0.2483)
  (Qwen-14B,0.2323)
  (Qwen-32B,0.3187)
};

\addplot[fullbar,bar shift=-1.5pt] coordinates {
  (OLMo-1B,0.3167)
  (LLaMA-8B,0.1863)
  (Qwen-14B,0.3237)
  (Qwen-32B,0.3500)
};

\addplot[temporalbar,bar shift=1.5pt] coordinates {
  (OLMo-1B,0.0457)
  (LLaMA-8B,0.2497)
  (Qwen-14B,0.2330)
  (Qwen-32B,0.3193)
};

\addplot[curriculumbar,bar shift=4.5pt] coordinates {
  (OLMo-1B,0.0450)
  (LLaMA-8B,0.2490)
  (Qwen-14B,0.2323)
  (Qwen-32B,0.3177)
};

\addplot[historybar,bar shift=7.5pt] coordinates {
  (OLMo-1B,0.0460)
  (LLaMA-8B,0.2500)
  (Qwen-14B,0.2333)
  (Qwen-32B,0.3217)
};

% =========================================================
% LRS
% =========================================================
\nextgroupplot[
  title={LRS},
  ymin=0.30,
  ymax=0.55,
  ytick={0.30,0.40,0.50}
]

\addplot[basebar,bar shift=-7.5pt] coordinates {
  (OLMo-1B,0.3633)
  (LLaMA-8B,0.4423)
  (Qwen-14B,0.3993)
  (Qwen-32B,0.3367)
};

\addplot[sftbar,bar shift=-4.5pt] coordinates {
  (OLMo-1B,0.3460)
  (LLaMA-8B,0.4207)
  (Qwen-14B,0.4033)
  (Qwen-32B,0.3580)
};

\addplot[fullbar,bar shift=-1.5pt] coordinates {
  (OLMo-1B,0.4833)
  (LLaMA-8B,0.4993)
  (Qwen-14B,0.4993)
  (Qwen-32B,0.3900)
};

\addplot[temporalbar,bar shift=1.5pt] coordinates {
  (OLMo-1B,0.3450)
  (LLaMA-8B,0.4190)
  (Qwen-14B,0.4023)
  (Qwen-32B,0.3597)
};

\addplot[curriculumbar,bar shift=4.5pt] coordinates {
  (OLMo-1B,0.3447)
  (LLaMA-8B,0.4187)
  (Qwen-14B,0.4020)
  (Qwen-32B,0.3573)
};

\addplot[historybar,bar shift=7.5pt] coordinates {
  (OLMo-1B,0.3450)
  (LLaMA-8B,0.4190)
  (Qwen-14B,0.4023)
  (Qwen-32B,0.3623)
};

% =========================================================
% AnnoMI
% =========================================================
\nextgroupplot[
  title={AnnoMI},
  ymin=0.15,
  ymax=0.70,
  ytick={0.20,0.40,0.60}
]

\addplot[basebar,bar shift=-7.5pt] coordinates {
  (OLMo-1B,0.1780)
  (LLaMA-8B,0.3197)
  (Qwen-14B,0.2773)
  (Qwen-32B,0.5300)
};

\addplot[sftbar,bar shift=-4.5pt] coordinates {
  (OLMo-1B,0.2517)
  (LLaMA-8B,0.3640)
  (Qwen-14B,0.3473)
  (Qwen-32B,0.5633)
};

\addplot[fullbar,bar shift=-1.5pt] coordinates {
  (OLMo-1B,0.3293)
  (LLaMA-8B,0.3657)
  (Qwen-14B,0.6087)
  (Qwen-32B,0.6100)
};

\addplot[temporalbar,bar shift=1.5pt] coordinates {
  (OLMo-1B,0.2507)
  (LLaMA-8B,0.3640)
  (Qwen-14B,0.3477)
  (Qwen-32B,0.5650)
};

\addplot[curriculumbar,bar shift=4.5pt] coordinates {
  (OLMo-1B,0.2500)
  (LLaMA-8B,0.3633)
  (Qwen-14B,0.3470)
  (Qwen-32B,0.5627)
};

\addplot[historybar,bar shift=7.5pt] coordinates {
  (OLMo-1B,0.2543)
  (LLaMA-8B,0.3677)
  (Qwen-14B,0.3513)
  (Qwen-32B,0.5640)
};

% =========================================================
% CMV
% =========================================================
\nextgroupplot[
  title={CMV},
  ymin=0.35,
  ymax=0.74,
  ytick={0.40,0.55,0.70}
]

\addplot[basebar,bar shift=-7.5pt] coordinates {
  (OLMo-1B,0.3640)
  (LLaMA-8B,0.4653)
  (Qwen-14B,0.5363)
  (Qwen-32B,0.6867)
};

\addplot[sftbar,bar shift=-4.5pt] coordinates {
  (OLMo-1B,0.4493)
  (LLaMA-8B,0.5110)
  (Qwen-14B,0.5270)
  (Qwen-32B,0.6917)
};

\addplot[fullbar,bar shift=-1.5pt] coordinates {
  (OLMo-1B,0.3887)
  (LLaMA-8B,0.5700)
  (Qwen-14B,0.5623)
  (Qwen-32B,0.7033)
};

\addplot[temporalbar,bar shift=1.5pt] coordinates {
  (OLMo-1B,0.4447)
  (LLaMA-8B,0.5080)
  (Qwen-14B,0.5257)
  (Qwen-32B,0.6923)
};

\addplot[curriculumbar,bar shift=4.5pt] coordinates {
  (OLMo-1B,0.4463)
  (LLaMA-8B,0.5107)
  (Qwen-14B,0.5263)
  (Qwen-32B,0.6913)
};

\addplot[historybar,bar shift=7.5pt] coordinates {
  (OLMo-1B,0.4513)
  (LLaMA-8B,0.5143)
  (Qwen-14B,0.5307)
  (Qwen-32B,0.6943)
};

\end{groupplot}
\end{tikzpicture}%
}

\vspace{4pt}

% =========================================================
% Compact one-row legend
% =========================================================
{\tiny
\setlength{\tabcolsep}{4pt}
\renewcommand{\arraystretch}{1.0}

\begin{tabular}{cccccc}
\ablationlegenditem{basebar}{IFT}
&
\ablationlegenditem{sftbar}{SFT}
&
\ablationlegenditem{fullbar}{LiFT Full}
&
\ablationlegenditem{temporalbar}{w/o Temporal}
&
\ablationlegenditem{curriculumbar}{w/o Curriculum}
&
\ablationlegenditem{historybar}{w/o History}
\end{tabular}
}

\vspace{5pt}

\captionof{figure}{\textit{Cross-backbone ablation.}
Absolute Macro-F1 averaged across 0-, 1-, and 3-shot settings for
four backbones.}
\label{fig:ablation_absolute}

\end{minipage}
\end{figure*}

\begin{table*}[!t]
\centering
\scriptsize
\setlength{\tabcolsep}{3.5pt}
\renewcommand{\arraystretch}{1.08}

\resizebox{\textwidth}{!}{%
\begin{tabular}{l|ccc|ccc|ccc||ccc|ccc}
\toprule

& \multicolumn{3}{c|}{\cellcolor{green!10}\textbf{AnnoMI}}
& \multicolumn{3}{c|}{\cellcolor{green!10}\textbf{LRS}}
& \multicolumn{3}{c||}{\cellcolor{green!10}\textbf{TalkLife}}
& \multicolumn{3}{c|}{\cellcolor{red!10}\textbf{Reddit (OOD)}}
& \multicolumn{3}{c}{\cellcolor{red!10}\textbf{CMV (OOD)}} \\

\midrule

\textbf{Shots}
& \cellcolor{green!10}\textbf{0}
& \cellcolor{green!10}\textbf{1}
& \cellcolor{green!10}\textbf{3}
& \cellcolor{green!10}\textbf{0}
& \cellcolor{green!10}\textbf{1}
& \cellcolor{green!10}\textbf{3}
& \cellcolor{green!10}\textbf{0}
& \cellcolor{green!10}\textbf{1}
& \cellcolor{green!10}\textbf{3}
& \cellcolor{red!10}\textbf{0}
& \cellcolor{red!10}\textbf{1}
& \cellcolor{red!10}\textbf{3}
& \cellcolor{red!10}\textbf{0}
& \cellcolor{red!10}\textbf{1}
& \cellcolor{red!10}\textbf{3} \\

\midrule
%%%%%%%%%%%%%%%%%%%%%%%%%%%%%%%%%%%%%%%%%%%%%%%%%%%%
\rowcolor{gray!8}
\textbf{OLMo-1B (IFT)}
& .178 & .178 & .178
& .354 & .364 & .371
& .225 & .217 & .290
& .136 & .268 & .150
& .361 & .362 & .369 \\

\rowcolor{orange!10}
\textbf{SFT}
& .219 & .257 & .279
& .359 & .325 & .354
& .052 & .041 & .044
& .260 & .216 & .238
& .337 & .342 & .369 \\

\rowcolor{orange!10}
\textbf{LiFT w/o Temporal}
& .215 & .259 & .278
& .353 & .326 & .356
& .051 & .042 & .044
& .236 & .219 & .261
& .325 & .351 & .358 \\

\rowcolor{orange!10}
\textbf{LiFT w/o Curriculum}
& .214 & .257 & .279
& .352 & .327 & .355
& .048 & .042 & .045
& .235 & .218 & .262
& .332 & .347 & .360 \\

\rowcolor{orange!10}
\textbf{LiFT w/o History Loss}
& .217 & .255 & .291
& .339 & .345 & .351
& .047 & .041 & .050
& .227 & .260 & .234
& .371 & .344 & .339 \\

\rowcolor{teal!8}
\textbf{LiFT (Full)}
& \textbf{.290} & \textbf{.335} & \textbf{.363}
& \textbf{.449} & \textbf{.473} & \textbf{.528}
&.224 & \textbf{.254} & \textbf{.321}
& \textbf{.301} & \textbf{.317} & \textbf{.323}
& \textbf{.400} & \textbf{.377} & \textbf{.389} \\

\midrule
%%%%%%%%%%%%%%%%%%%%%%%%%%%%%%%%%%%%%%%%%%%%%%%%%%%%
\rowcolor{gray!8}
\textbf{OLMo-7B (IFT)}
& .207 & .379 & .446
& .124 & .128 & .125
& .029 & .008 & .064
& .227 & .219 & .298
& .482 & .470 & .486 \\

\rowcolor{orange!10}
\textbf{SFT}
& .290 & .312 & .330
& .330 & .352 & .378
& .070 & .049 & .055
& .271 & .235 & .281
& .500 & .472 & .506 \\

\rowcolor{orange!10}
\textbf{LiFT w/o Temporal}
& .282 & .318 & .331
& .306 & .345 & .377
& .063 & .054 & .057
& .265 & .239 & .287
& .480 & .481 & .500 \\

\rowcolor{orange!10}
\textbf{LiFT w/o Curriculum}
& .281 & .316 & .332
& .305 & .346 & .376
& .060 & .050 & .058
& .264 & .237 & .288
& .490 & .479 & .512 \\

\rowcolor{orange!10}
\textbf{LiFT w/o History Loss}
& .284 & .314 & .344
& .292 & .364 & .382
& .059 & .053 & .063
& .256 & .254 & .286
& .485 & .487 & .520 \\

\rowcolor{teal!8}
\textbf{LiFT (Full)}
& \textbf{.339} & \textbf{.443} & \textbf{.448}
& \textbf{.380} & \textbf{.584} & \textbf{.587}
& \textbf{.298} & \textbf{.305} & \textbf{.324}
& \textbf{.298} & \textbf{.300} & \textbf{.319}
& \textbf{.601} & \textbf{.606} & \textbf{.618} \\

\midrule
%%%%%%%%%%%%%%%%%%%%%%%%%%%%%%%%%%%%%%%%%%%%%%%%%%%%
\rowcolor{gray!8}
\textbf{LLaMA-8B (IFT)}
& .207 & .352 & .400
& .437 & .460 & .430
& .067 & .073 & .180
& .292 & .291 & .291
& .466 & .468 & .462 \\

\rowcolor{orange!10}
\textbf{SFT}
& .181 & .334 & .289
& .448 & .401 & .426
& .108 & .141 & .186
& .251 & .307 & .284
& .491 & .514 & .536 \\

\rowcolor{orange!10}
\textbf{LiFT w/o Temporal}
& .192 & .268 & .309
& .398 & .372 & .421
& .118 & .091 & .159
& .229 & .298 & .261
& .472 & .491 & .509 \\

\rowcolor{orange!10}
\textbf{LiFT w/o Curriculum}
& .198 & .311 & .361
& .447 & .421 & .476
& .111 & .156 & .121
& .271 & .302 & .344
& .519 & .536 & .553 \\

\rowcolor{orange!10}
\textbf{LiFT w/o History Loss}
& .201 & .327 & .379
& .438 & .471 & .489
& .132 & .169 & .214
& .284 & .280 & .311
& .551 & .538 & .566 \\

\rowcolor{teal!8}
\textbf{LiFT (Full)}
& .207 & \textbf{.379} & \textbf{.457}
& \textbf{.480} & \textbf{.507} & \textbf{.511}
& \textbf{.145} & \textbf{.183} & \textbf{.231}
& \textbf{.296} & \textbf{.320} & \textbf{.384}
& \textbf{.561} & \textbf{.568} & \textbf{.581} \\

\midrule
%%%%%%%%%%%%%%%%%%%%%%%%%%%%%%%%%%%%%%%%%%%%%%%%%%%%
\rowcolor{gray!8}
\textbf{Qwen-14B (IFT)}
& .357 & .271 & .204
& .384 & .414 & .400
& .177 & .027 & .090
& .107 & .167 & .127
& \textbf{.509} & .547 & .553 \\

\rowcolor{orange!10}
\textbf{SFT}
& .253 & .404 & .385
& .419 & .378 & .413
& .220 & .196 & .281
& .219 & .252 & .309
& .458 & .520 & .543 \\

\rowcolor{orange!10}
\textbf{LiFT w/o Temporal}
& .248 & .408 & .387
& .412 & .380 & .415
& .216 & .199 & .284
& .218 & .257 & .307
& .450 & .531 & .536 \\

\rowcolor{orange!10}
\textbf{LiFT w/o Curriculum}
& .247 & .406 & .388
& .411 & .381 & .414
& .213 & .199 & .285
& .217 & .256 & .308
& .454 & .526 & .539 \\

\rowcolor{orange!10}
\textbf{LiFT w/o History Loss}
& .250 & .404 & .400
& .398 & .399 & .410
& .212 & .198 & .290
& .209 & .272 & .306
& .456 & .532 & .544 \\

\rowcolor{teal!8}
\textbf{LiFT (Full)}
& \textbf{.513} & \textbf{.626} & \textbf{.687}
& \textbf{.450} & \textbf{.474} & \textbf{.496}
& \textbf{.315} & \textbf{.321} & \textbf{.335}
& \textbf{.260} & \textbf{.376} & \textbf{.377}
& .479 & \textbf{.584} & \textbf{.624} \\

\midrule
%%%%%%%%%%%%%%%%%%%%%%%%%%%%%%%%%%%%%%%%%%%%%%%%%%%%
\rowcolor{gray!8}
\textbf{Qwen-32B (IFT)}
& .512 & .524 & .556 & .363 & .344 & .308
& .290 & .300 & .304 & .224 & .313 & .391
& .650 & .700 & .710 \\
\rowcolor{orange!10}
\textbf{SFT}
& .545 & .560 & .585
& .366 & .358 & .350
& .315 & .306 & .335
& .250 & .336 & .392
& .653 & .702 & .720 \\

\rowcolor{orange!10}
\textbf{LiFT w/o Temporal}
& .540 & .565 & .590
& .364 & .360 & .355
& .310 & .308 & .340
& .247 & .342 & .394
& .651 & .704 & .722 \\

\rowcolor{orange!10}
\textbf{LiFT w/o Curriculum}
& .538 & .562 & .588
& .363 & .357 & .352
& .308 & .307 & .338
& .245 & .340 & .393
& .652 & .703 & .719 \\

\rowcolor{orange!10}
\textbf{LiFT w/o History Loss}
& .542 & .558 & .592
& .362 & .365 & .360
& .312 & .309 & .344
& .242 & .345 & .395
& .654 & .705 & .724 \\

\rowcolor{teal!8}
\textbf{LiFT (Full)}

& \textbf{.594} & \textbf{.612} & .\textbf{632} & \textbf{.381} & \textbf{.393} & \textbf{.400}
& \textbf{.354 }& \textbf{.320} & \textbf{.378} & \textbf{.292} & \textbf{.380} & \textbf{.400}
&\textbf{ .660 }& \textbf{.712} & .\textbf{736} \\

\bottomrule
\end{tabular}%
}

\caption{
Full ablation study of the \textit{LiFT} framework across five model scales
(\textit{OLMo-1B}, \textit{OLMo-7B}, \textit{LLaMA-8B},
\textit{Qwen-14B}, and \textit{Qwen-32B}).
We compare the IFT model, SFT baseline, full LiFT model, and variants removing
temporal conditioning, curriculum learning, and the history-only auxiliary loss.
Results are reported as Macro-F1 across five datasets and 0-, 1-, and 3-shot
settings. Bold indicates the highest Macro-F1 within each model, dataset, and
shot setting; tied values are not bolded.
}
\label{tab:ablation_all}
\end{table*}

\section{Compute Resources, Training Cost and Training Details}
\label{sec:compute}

In \hyperref[tab:compute]{\underline{\textcolor{blue}{Table~\ref*{tab:compute}}}}, we report the compute resources and training configuration for all LiFT models. 
Training follows a three-stage longitudinal curriculum:
AnnoMI $\rightarrow$ LRS $\rightarrow$ TalkLife. Each stage continues from the
previous stage using LoRA adapters and task-specific longitudinal heads. We use LoRA scaling $\alpha=2r$ with dropout $0.05$. At each higher-rank stage, previous adapter weights are loaded via rank-resize initialization, while earlier rank slices are frozen and only newly added ranks are updated.
% Required packages:
% \usepackage{booktabs}
% \usepackage{tabularx}
% \usepackage[table]{xcolor}
% \usepackage{array}

\begin{table*}[!t]
\centering
\scriptsize
\renewcommand{\arraystretch}{0.94}
\setlength{\tabcolsep}{3.5pt}
\setlength{\aboverulesep}{0.3pt}
\setlength{\belowrulesep}{0.3pt}

\begin{tabularx}{\textwidth}{@{}p{0.235\textwidth}X@{}}
\toprule
\rowcolor{gray!15}
\textbf{Component} & \textbf{Details} \\
\midrule

% ------------------------------------------------------------------
\rowcolor{gray!20}
\multicolumn{2}{@{}l}{\textbf{Models and compute}} \\

Models
& OLMo-1B, OLMo-7B, LLaMA-3.1-8B, Qwen2.5-14B, and
Qwen2.5-32B \\

\rowcolor{gray!8}
Model Checkpoints
& Exact Hugging Face model identifiers and revisions are recorded in
the released configuration files \\

Hardware
& NVIDIA A100 (80GB), H100, and GH200 (120GB) GPUs \\

\rowcolor{gray!8}
\# GPUs
& 2--3 depending on model size; Qwen2.5-32B uses 2 GPUs \\

Training Method
& LoRA-based parameter-efficient fine-tuning; where
backbone parameters remain frozen \\

\rowcolor{gray!8}
Precision
& bfloat16 on CUDA  \\

% ------------------------------------------------------------------
\midrule
\rowcolor{gray!20}
\multicolumn{2}{@{}l}{\textbf{Training and curriculum}} \\

Training Stages
& Three sequential stages:
AnnoMI $\rightarrow$ LRS $\rightarrow$ TalkLife \\

\rowcolor{gray!8}
Continuous Training
& Each stage initializes from the preceding stage's LoRA and
wrapper-head state; the Stage-3 state forms the final LiFT model \\

Training Seed
& Fixed seed 42 for Python and PyTorch; each model configuration is
trained once \\

\rowcolor{gray!8}
Epochs
& 1 epoch per curriculum stage \\

Batch Size (per device)
& 2 chunks \\

\rowcolor{gray!8}
Gradient Accumulation
& 8 steps \\

Effective Batch Size
& $2 \times 8 \times \#\mathrm{GPUs}
=16\times\#\mathrm{GPUs}$ chunks \\

\rowcolor{gray!8}
Optimizer
& AdamW (\texttt{adamw\_torch}) \\

Gradient Clipping
& Maximum gradient norm of 1.0 \\

\rowcolor{gray!8}
Validation Split
& 10\% chunk-level validation split at each stage, constructed with
seed 42 \\

Best Checkpoint
& Checkpoint with the lowest validation loss; best weights are loaded
at the end of each stage \\

\rowcolor{gray!8}
Checkpointing
& Every \texttt{SAVE\_STEPS} optimization steps
(default: 25); at most two checkpoints are retained per stage \\

Resume Strategy
& Stage-level and within-stage recovery from S3 checkpoints, including
model, optimizer, scheduler, RNG, and global-step states \\

% ------------------------------------------------------------------
\midrule
\rowcolor{gray!20}
\multicolumn{2}{@{}l}{\textbf{LoRA configuration}} \\

LoRA Target Modules
& \texttt{q\_proj}, \texttt{k\_proj}, \texttt{v\_proj},
\texttt{o\_proj}, \texttt{gate\_proj}, \texttt{up\_proj}, and
\texttt{down\_proj} \\

\rowcolor{gray!8}
Stage 1
& $r=4$, $\alpha=8$, learning rate $2{\times}10^{-4}$,
cosine schedule \\

Stage 2
& $r=8$, $\alpha=16$, learning rate $1{\times}10^{-4}$,
cosine schedule \\

\rowcolor{gray!8}
Stage 3
& $r=16$, $\alpha=32$, learning rate $5{\times}10^{-5}$,
constant schedule \\

LoRA Dropout
& 0.05 \\

\rowcolor{gray!8}
Cross-Stage Expansion
& Previous LoRA dimensions are transferred when the rank increases;
new rank dimensions remain trainable \\

% ------------------------------------------------------------------
\midrule
\rowcolor{gray!20}
\multicolumn{2}{@{}l}{\textbf{Sequence construction and LiFT components}} \\
Special Tokens
& \texttt{<instruction>}, \texttt{<few-shot>},
\texttt{<query>}, \texttt{<output>}, \texttt{<HIST>},
\texttt{<CURR>}, and task-specific tags \\

Global Label Space
& Shared namespace covering AnnoMI, LRS, and TalkLife labels, with one
additional padding label \\

\rowcolor{gray!8}
Temporal Components
&  absolute-position, relative-position, timestep, and
global-label embeddings \\

Temporal Fusion
& Temporal features are concatenated, transformed by a two-layer MLP,
projected to the model hidden size, and added to token embeddings \\

\rowcolor{gray!8}
Auxiliary Heads
& Global classification head and history-only classification head \\

% ------------------------------------------------------------------
% \midrule
% \rowcolor{gray!20}
% \multicolumn{2}{@{}l}{\textbf{Training objective}} \\

% Prompt CE Loss
% & Next-token cross-entropy outside the \texttt{<HIST>} span, with
% $\lambda_{\mathrm{CE}}=1.0$ \\

% \rowcolor{gray!8}
% Output LM Loss
% & Focal language-model loss over the output and EOS region, with
% $\lambda_{\mathrm{out}}=1.0$ and $\gamma=1.0$ \\

% Global Classification Loss
% & Class-balanced focal loss from the global classification head, with
% $\lambda_{\mathrm{cls}}=1.0$ \\

% \rowcolor{gray!8}
% History Classification Loss
% & Class-balanced focal loss from mean-pooled \texttt{<HIST>}
% representations, with $\lambda_{\mathrm{hist}}=0.2$ \\

% Class Weighting
% & Effective-number class balancing with $\beta=0.99$ \\

% \rowcolor{gray!8}
% Total Loss
% & Weighted sum of prompt CE, output focal LM, global classification,
% and history-only classification losses \\

% ------------------------------------------------------------------
\midrule
\rowcolor{gray!20}
\multicolumn{2}{@{}l}{\textbf{Inference and evaluation}} \\

Shot Settings
& Zero-, one-, and three-shot prompting \\

\rowcolor{gray!8}
Inference Seeds
& 42, 43, and 44; seeds are applied only during inference \\

Inference Determinism
& Python, NumPy, PyTorch, and CUDA are seeded; CuDNN deterministic
mode is enabled\\

\rowcolor{gray!8}
Few-Shot Demonstrations
& Fixed demonstrations are used for one- and three-shot prompting \\

Decoding
& Greedy generation \\

Evaluation
& Per-class precision, recall, F1, support, accuracy, and Macro-F1 \\

\rowcolor{gray!8}
Run Aggregation
& Prediction files and classification reports are retained for each
seed; arithmetic means and sample standard deviations are reported
across the three inference runs \\

Inference Interpretation
& Because decoding and demonstrations are fixed, the seeded runs
measure implementation-level inference stability rather than training
variance \\

Statistical Testing
& Prediction-level paired bootstrap tests with Benjamini--Hochberg
correction over the specified IFT--LiFT comparison family \\

\rowcolor{gray!8}
Released Artifacts
& Training and inference scripts, prompt templates,
and environment specifications \\

% ------------------------------------------------------------------
\midrule
\rowcolor{gray!20}
\multicolumn{2}{@{}l}{\textbf{Resource estimates}} \\

Training Time (per model)
& 4--8 h (1B), 12--18 h (7B), 18--24 h (8B),
24--36 h (14B), and approximately 36--48 h (32B) \\

\rowcolor{gray!8}
GPU Hours (per model)
& 30--80 (1B), 100--200 (7B), 150--300 (8B),
250--500 (14B), and approximately 400--800 (32B) \\

Total Compute
& Approximately 1,500--3,000 GPU hours, including model training and
ablation experiments \\

\rowcolor{gray!8}
Memory Usage
& Approximately 20--35 GB per GPU for 1B--14B models and
80--110 GB per GPU for Qwen2.5-32B \\

Storage
& Approximately 50--150 GB per model and up to approximately 200 GB
for 32B checkpoints, adapters, merged models, predictions, and logs \\

\rowcolor{gray!8}
Carbon Estimate
& Approximately 75--180 kg CO$_2$ across training runs and ablations \\

\bottomrule
\end{tabularx}

\caption{\textit{LiFT compute and reproducibility configuration.}
The table summarizes the principal training, inference, evaluation,
and resource settings.}
\label{tab:compute}
\end{table*}

\section{Ablation Study}
\label{sec:ablation_app}

To analyze the contribution of individual components in the LiFT framework, we perform ablation experiments by removing key modules while keeping the remaining training setup unchanged, and compare against a standard supervised fine-tuning (SFT) baseline trained on same datasets, the training pipeline and the same LoRA-based adaptation. 
\hyperref[tab:ablation_all]{\underline{\textcolor{blue}{Table~\ref*{tab:ablation_all}}}} 
and 
\hyperref[fig:ablation_absolute]{\underline{\textcolor{blue}{Figure~\ref*{fig:ablation_absolute}}}} 
report results for \textit{OLMo-1B, 7B}, \textit{LLaMA-8B}, and \textit{Qwen-14B, 32B}, showing trends across architectures. 
Across models and datasets, the \textit{full LiFT configuration achieves the best macro-F1}, substantially outperforming both the IFT model and the SFT baseline. While SFT generally provides small improvements over the IFT model in some settings, it fails to capture the longitudinal structure of the data. Removing either \textit{temporal conditioning} or \textit{curriculum learning} generally degrades performance, indicating that temporal signals and staged training help capture longitudinal dependencies. In contrast, removing the \textit{history-aware loss} yields mixed effects: performance usually decreases but occasionally improves, suggesting complex interactions with other objectives. Overall, curriculum learning and temporal conditioning provide gains, while the history-aware loss offers complementary but less uniform benefits. For example, on OLMo-1B (Reddit 1-shot), removing temporal conditioning reduces macro-F1 from .285 to .219, with a similar drop (.218) when removing curriculum learning.

\subsection{Context Source Analysis: CMV.}
We compare three context-selection strategies on CMV: Conversation-Level Context, Author History (All), and Author History (Same-Topic), across 0- and 1-shot settings
(\hyperref[tab:cmv_context_lift]{\underline{\textcolor{blue}{Table~\ref*{tab:cmv_context_lift}}}}).
LiFT consistently improves over the corresponding IFT model, with Author History (Same-Topic) achieving the best performance. This highlights the value of topic-aligned historical context for persuasion prediction.
\section{Statistical Testing Details}
\label{app:statistical_tests}

We conduct statistical tests to assess whether LiFT's improvements over IFT ICL are reliable across seeds, datasets, and shot settings, and to determine which empirical claims are supported. Let $y_i$ denote the gold label for example $i$, and let $\hat{y}^{I}_{i}$ and $\hat{y}^{L}_{i}$ denote the paired IFT and LiFT predictions for the same example. For each dataset, model family, shot setting, and seed, we align predictions by example identifier.

\subsection{Paired bootstrap tests.}
For each paired comparison, we resample matched test examples with replacement and recompute the metric difference
{\scriptsize
\[
\Delta_m = m(y, \hat{y}^{L}) - m(y, \hat{y}^{I}),
\]}
where $m$ is macro-F1, macro-recall, accuracy, minority-label recall, or minority-vs-rest F1. We use 1,000 bootstrap samples and report the observed difference, 95\% confidence interval, and two-sided bootstrap $p$-value. We expect LiFT to yield positive $\Delta_m$, especially for macro-F1 on longitudinal datasets.

\subsection{McNemar tests.}
To test whether 
LiFT fixes more IFT errors than it introduces, we compute the discordant counts
where $I_i$ and $L_i$ indicate whether IFT and LiFT are correct on example $i$:

{\scriptsize\[
\begin{aligned}
b_{01} &= \sum_i \mathbb{I}[I_i=0, L_i=1],\\
b_{10} &= \sum_i \mathbb{I}[I_i=1, L_i=0].
\end{aligned}
\]}
We apply the exact McNemar binomial test under $H_0: b_{01}=b_{10}$ and report the odds ratio $(b_{01}+0.5)/(b_{10}+0.5)$. We expect odds ratios above 1 when LiFT improves raw correctness.

\subsection{Directional sign tests.}
To assess whether LiFT gains are directionally consistent across dataset--shot conditions, we count the number of positive macro-F1 deltas out of all tested conditions and apply a two-sided sign test under $p=0.5$. This tests whether improvements are systematic rather than driven by isolated datasets.

\subsection{Few-shot difference-in-differences.}
To test whether demonstrations amplify LiFT more than IFT, we compute Difference-In-Difference (DID), 

{\scriptsize\[
\operatorname{DID}_{a\rightarrow b}
=
\big(m^{L}_{b}-m^{L}_{a}\big)
-
\big(m^{I}_{b}-m^{I}_{a}\big),
\]}
where $a,b\in\{0,1,3\}$ denote shot settings and $m$ are the respective LiFT and IFT models. Positive DID indicates that the gain from additional demonstrations is larger for LiFT than for IFT. For DID, we use a four-way matched sample across IFT/LiFT and both shot settings, and reuse the same bootstrap resamples for all four conditions.

\subsection{Multiple comparisons.}
For each statistical output table, we apply Benjamini-Hochberg correction over the reported $p$-values. In the summary table below, ``sig.'' counts tests with Benjamini-Hochberg adjusted $p<0.05$ within each statistical test family; raw and adjusted $p$-values are provided in the released statistical tables.

\providecommand{\sig}[1]{\textsuperscript{#1}}
\providecommand{\std}[1]{%
\raisebox{0.3ex}{\scalebox{0.55}{($\pm$ #1)}}%
}

\begin{table*}[!t]
\centering
\scriptsize
\setlength{\tabcolsep}{2pt}
\renewcommand{\arraystretch}{1.1}

\resizebox{\textwidth}{!}{%
\begin{tabular}{l|ccc|ccc|ccc||ccc|ccc}
\toprule

& \multicolumn{3}{c|}{\cellcolor{green!10}\textbf{AnnoMI}}
& \multicolumn{3}{c|}{\cellcolor{green!10}\textbf{LRS}}
& \multicolumn{3}{c||}{\cellcolor{green!10}\textbf{TalkLife}}
& \multicolumn{3}{c|}{\cellcolor{red!10}\textbf{Reddit (OOD)}}
& \multicolumn{3}{c}{\cellcolor{red!10}\textbf{CMV (OOD)}} \\
\midrule

\textbf{Shots}
& \cellcolor{green!10}\textbf{0}
& \cellcolor{green!10}\textbf{1}
& \cellcolor{green!10}\textbf{3}
& \cellcolor{green!10}\textbf{0}
& \cellcolor{green!10}\textbf{1}
& \cellcolor{green!10}\textbf{3}
& \cellcolor{green!10}\textbf{0}
& \cellcolor{green!10}\textbf{1}
& \cellcolor{green!10}\textbf{3}
& \cellcolor{red!10}\textbf{0}
& \cellcolor{red!10}\textbf{1}
& \cellcolor{red!10}\textbf{3}
& \cellcolor{red!10}\textbf{0}
& \cellcolor{red!10}\textbf{1}
& \cellcolor{red!10}\textbf{3} \\
\midrule

\textbf{OLMo-1B (IFT)}
& .178 \std{.000}
& .178 \std{.000}
& .178 \std{.000}
& .354 \std{.000}
& .364 \std{.000}
& .371 \std{.000}
& ..225 \std{.000}
& .217 \std{.000}
& .290 \std{.000}
& .136 \std{.024}
& .268 \std{.015}
& .150 \std{.025}
& .361 \std{.000}
& .362 \std{.000}
& .369 \std{.000} \\

\rowcolor{teal!8}
\textbf{OLMo-1B (LiFT)}
& \textbf{.290}\sig{***} \std{.000}
& \textbf{.335}\sig{***} \std{.000}
& \textbf{.363}\sig{***} \std{.000}
& \textbf{.449}\sig{***} \std{.003}
& \textbf{.473}\sig{***} \std{.000}
& \textbf{.528}\sig{***} \std{.000}
& \textbf{.224} \std{.016}
& \textbf{.254}\sig{***} \std{.000}
& \textbf{.321}\sig{*} \std{.000}
& \textbf{.301}\sig{***} \std{.000}
& \textbf{.317}\sig{***} \std{.000}
& \textbf{.323}\sig{***} \std{.000}
& \textbf{.400}\sig{***} \std{.000}
& \textbf{.377}\sig{**} \std{.000}
& \textbf{.389}\sig{***} \std{.000} \\

\midrule

\textbf{OLMo-7B (IFT)}
& .207 \std{.000}
& .379 \std{.000}
& .446 \std{.000}
& .124 \std{.174}
& .128 \std{.000}
& .125 \std{.000}
& .029 \std{.000}
& .008 \std{.000}
& .064 \std{.000}
& .227 \std{.023}
& .219 \std{.019}
& .298 \std{.014}
& .482 \std{.000}
& .470 \std{.000}
& .486 \std{.000} \\

\rowcolor{teal!8}
\textbf{OLMo-7B (LiFT)}
& \textbf{.339}\sig{***} \std{.000}
& \textbf{.443} \std{.000}
& \textbf{.448} \std{.000}
& \textbf{.380} \std{.000}
& \textbf{.584}\sig{***} \std{.000}
& \textbf{.587}\sig{***} \std{.000}
& \textbf{.298}\sig{***} \std{.000}
& \textbf{.305}\sig{***} \std{.000}
& \textbf{.324}\sig{***} \std{.000}
& \textbf{.298} \std{.000}
& \textbf{.300}\sig{***} \std{.000}
& \textbf{.319} \std{.000}
& \textbf{.601}\sig{***} \std{.000}
& \textbf{.606}\sig{***} \std{.000}
& \textbf{.618}\sig{***} \std{.000} \\

\midrule

\textbf{LLaMA-8B (IFT)}
& .207 \std{.000}
& .352 \std{.000}
& .400 \std{.000}
& .437 \std{.000}
& .460 \std{.000}
& .430 \std{.000}
& .067 \std{.000}
& .073 \std{.000}
& .180 \std{.000}
& .292 \std{.000}
& .291 \std{.000}
& .291 \std{.000}
& .466 \std{.000}
& .468 \std{.000}
& .462 \std{.000} \\

\rowcolor{teal!8}
\textbf{LLaMA-8B (LiFT)}
& .207 \std{.000}
& \textbf{.379} \std{.000}
& \textbf{.457} \std{.000}
& \textbf{.480}\sig{**} \std{.000}
& \textbf{.507}\sig{***} \std{.000}
& \textbf{.511}\sig{***} \std{.000}
& \textbf{.145}\sig{***} \std{.000}
& \textbf{.183}\sig{***} \std{.000}
& \textbf{.231}\sig{***} \std{.000}
& \textbf{.296} \std{.016}
& \textbf{.320} \std{.009}
& \textbf{.384}\sig{**} \std{.019}
& \textbf{.561}\sig{***} \std{.001}
& \textbf{.568}\sig{***} \std{.003}
& \textbf{.581}\sig{***} \std{.002} \\

\midrule
\textbf{Qwen-14B (IFT)}
& .357 \std{.000}
& .271 \std{.000}
& .204 \std{.000}
& .384 \std{.000}
& .414 \std{.000}
& .400 \std{.000}
& .177 \std{.000}
& .027 \std{.000}
& .090 \std{.000}
& .107 \std{.012}
& .167 \std{.024}
& .127 \std{.015}
& \textbf{.509} \std{.000}
& .547 \std{.000}
& .553 \std{.000} \\

\rowcolor{teal!8}
\textbf{Qwen-14B (LiFT)}
& \textbf{.513}\sig{***} \std{.000}
& \textbf{.626}\sig{***} \std{.000}
& \textbf{.687}\sig{***} \std{.000}
& \textbf{.450}\sig{***} \std{.000}
& \textbf{.474}\sig{***} \std{.000}
& \textbf{.496}\sig{***} \std{.000}
& \textbf{.315}\sig{***} \std{.000}
& \textbf{.321}\sig{***} \std{.000}
& \textbf{.335}\sig{***} \std{.000}
& \textbf{.260}\sig{***} \std{.022}
& \textbf{.376}\sig{***} \std{.026}
& \textbf{.377}\sig{***} \std{.011}
& .479\sig{$\dagger$} \std{.000}
& \textbf{.584}\sig{***} \std{.000}
& \textbf{.624}\sig{***} \std{.000} \\
\midrule

\textbf{Qwen-32B (IFT)}
& .512\std{.000} & .524  \std{.000}& .556\std{.000}
& .363  \std{.001}& .344\std{.000} & .308 \std{.000}
& .290 \std{.000} & .300\std{.000} & .304 \std{.000}
& .224 \std{.020} & .313 \std{.000} & .391\std{.000}
& .65 \std{.000} & .70 \std{.000} & .71 \std{.012} \\

\rowcolor{teal!8}
\textbf{Qwen-32B (LiFT)}
& \textbf{.594}\sig{***} \std{.004}
& \textbf{.612}\sig{***} \std{.002}
& \textbf{.632}\sig{***} \std{.002}
& \textbf{.381}\sig{**} \std{.000}
& \textbf{.393}\sig{***} \std{.022}
& \textbf{.400}\sig{**} \std{.010}
& \textbf{.354}\std{.000}
& \textbf{.320}\std{.000}
& \textbf{.378}\std{.000}
& \textbf{.292}\sig{***} \std{.000}
& \textbf{.380}\sig{***} \std{.005}
& \textbf{.400}\sig{***} \std{.001}
& \textbf{.660}\std{.000}
& \textbf{.712} \std{.000}
& \textbf{.736} \std{.000}\\
\bottomrule
\end{tabular}%
}

\caption{\textit{Detailed results corresponding to
Table~\ref{tab:sota_comparison}.}
Macro-F1 is reported as mean $\pm$ standard deviation over three seeds.
Values are shown to three decimal places; therefore, some standard
deviations round to .000. Bold indicates the higher Macro-F1 within
each IFT--LiFT comparison. Significance markers are defined in
Table~\ref{tab:sota_comparison}.}
\label{tab:sota_comparison_detailed}
\end{table*}

\begin{table*}[!t]
\centering
\small
\setlength{\tabcolsep}{3.5pt}
\renewcommand{\arraystretch}{1.08}
\resizebox{\textwidth}{!}{%
\begin{tabular}{lcccccc}
\toprule
% \rowcolor{gray!10}

\textbf{Model family} &
\textbf{$\Delta$ Macro-F1} &
\textbf{Macro-F1 tests} &
\textbf{Sign test} &
\textbf{McNemar} &
\textbf{Minority effect} &
\textbf{Few-shot DID} \\
\midrule
\rowcolor{orange!10}

\textbf{OLMo-1B}
& $+.100 / +.106$
& $15/15$ wins; $15$ sig.; adj. $\tilde{p}_{boot}=0.000$
& $15/15$, $p=6.1{\times}10^{-5}$
& $\widetilde{\mathrm{OR}}=1.65$; $12/15$ sig.; adj. $\tilde{p}=6.3{\times}10^{-9}$
& $\Delta$F1 $-.121/-.099$; $1/15$ sig.+; recall $0/15$ sig.+
& $+.011/+.010$; $5/15$ sig.+; adj. $\tilde{p}=.019$ \\
\rowcolor{orange!10}

\textbf{OLMo-7B}
& $+.184 / +.132$
& $15/15$ wins; $10$ sig.; adj. $\tilde{p}_{boot}=0.000$
& $15/15$, $p=6.1{\times}10^{-5}$
& $\widetilde{\mathrm{OR}}=10.12$; $14/15$ sig.; adj. $\tilde{p}=2.0{\times}10^{-34}$
& $\Delta$F1 $+.089/+.098$; $10/15$ sig.+; recall $7/15$ sig.+
& $+.027/+.005$; $3/15$ sig.+; adj. $\tilde{p}=.112$ \\
\rowcolor{orange!10}

\textbf{LLaMA-8B}
& $+.062 / +.057$
& $14/15$ wins (one tie); $10$ sig.; adj. $\tilde{p}_{boot}=0.000$
& $14$ wins and one tie, $p=1.2{\times}10^{-4}$
& $\widetilde{\mathrm{OR}}=1.15$; $9/15$ sig.; adj. $\tilde{p}=9.2{\times}10^{-4}$
& $\Delta$F1 $+.023/+.011$; $7/15$ sig.+; recall $8/15$ sig.+
& $+.022/+.024$; $5/15$ sig.+; adj. $\tilde{p}=.066$ \\
\rowcolor{orange!10}
\textbf{Qwen-14B}
& $+.172 / +.153$
& $14/15$ wins; $14$ sig.; adj. $\tilde{p}_{boot}=0.000$
& $14/15$, $p=9.8{\times}10^{-4}$
& $\widetilde{\mathrm{OR}}=2.14$; $12/15$ sig.; adj. $\tilde{p}=1.5{\times}10^{-16}$
& $\Delta$F1 $+.194/+.218$; $11/15$ sig.+; recall $8/15$ sig.+
& $+.075/+.043$; $8/15$ sig.+; adj. $\tilde{p}=0.000$ \\

\midrule
\rowcolor{teal!8}

\textbf{Overall}
& $+.130 / +.098$
& $58/60$ wins (one tie); $49$ sig.; adj. $\tilde{p}_{boot}=0.000$
& $58$ wins, one tie, and one loss, $p=2.1{\times}10^{-16}$
& $\widetilde{\mathrm{OR}}=1.89$; $47/60$ sig.; adj. $\tilde{p}=2.6{\times}10^{-16}$
& $\Delta$F1 $+.046/+.018$; $29/60$ sig.+; recall $23/60$ sig.+
& $+.034/+.019$; $21/60$ sig.+; adj. $\tilde{p}=.064$ \\

\bottomrule
\end{tabular}%
}
\caption{\textit{Summary of statistical tests across model families.} $\Delta$ Macro-F1 reports mean/median LiFT--IFT differences across dataset--shot conditions. ``Macro-F1 tests'' reports positive LiFT wins, Benjamini--Hochberg significant paired-bootstrap gains, and median adjusted bootstrap $p$-value. McNemar reports the median corrected-vs-broken odds ratio, number of significant tests, and median adjusted $p$-value. Minority effect reports mean/median minority-vs-rest F1 difference, significant positive minority-F1 tests, and significant positive minority-recall tests. DID measures whether few-shot demonstrations improve LiFT more than IFT. ``sig.'' denotes Benjamini--Hochberg adjusted $p<0.05$.}
\label{tab:statistical_summary}
\end{table*}

\section{Prompt Example Appendix}
\label{sec:appendix}
This appendix presents the prompting templates used across all datasets and
in-context learning (ICL) settings, including zero-shot, 1-shot, and 3-shot
variants. TalkLife prompts are shown in
\hyperref[fig:talklife_prompts]{\underline{\textcolor{blue}{Figures~\ref*{fig:talklife_prompts}}}}, 
AnnoMI prompts in
\hyperref[fig:AnnoMI_prompts]{\underline{\textcolor{blue}{Figures~\ref*{fig:AnnoMI_prompts}}}}, 
LRS prompts in
\hyperref[fig:lrs_prompts]{\underline{\textcolor{blue}{Figures~\ref*{fig:lrs_prompts}}}}, Reddit prompts in
\hyperref[fig:reddit_prompts]{\underline{\textcolor{blue}{Figures~\ref*{fig:reddit_prompts}}}} and CMV \hyperref[fig:cmv_prompts]{\underline{\textcolor{blue}{Figures~\ref*{fig:cmv_prompts}}}}.

\begin{figure*}[t]
\centering
\begin{tcolorbox}[
width=\textwidth,
title={Prompting Templates for TalkLife},
colback=gray!8,
colframe=black!60,
coltitle=white,
colbacktitle=black!65,
fonttitle=\bfseries,
boxrule=0.8pt,
arc=6pt,
left=10pt,
right=10pt,
top=8pt,
bottom=8pt
]

\scriptsize

\textbf{Task Description}

You are an expert annotator of mood dynamics in longitudinal mental health timelines.
Given a sequence of posts written by the same user in chronological order,
determine the mood change label for each post relative to its preceding timeline context.

\medskip
\textbf{Label Definitions}

\begin{itemize}
\item \textsc{O}: \textbf{No observable mood change}.  
The emotional state remains consistent with the previous timeline context.

\item \textsc{IS}: \textbf{Mood switch}.  
A clear shift in emotional valence or affective state occurs.

\item \textsc{IE}: \textbf{Mood escalation}.  
Emotion direction remains consistent but intensity, urgency, or distress increases.
\end{itemize}

\textbf{Annotation Guidelines}

\begin{itemize}
\item Posts appear in chronological order.
\item The first post in a timeline is labeled \textsc{O}.
\item Topic changes alone do not imply mood change.
\item Stronger wording alone does not imply escalation unless severity increases.

\end{itemize}

\medskip
\textbf{Prompt Variants}

\textbf{0-shot:} Model receives only the task description and input timeline.

\textbf{1-shot:} Model receives one labeled timeline example before the input.

\begin{quote}\ttfamily
Example Timeline:\\
Post 1: ``Example text'' $\rightarrow$ O\\
Post 2: ``Example text'' $\rightarrow$ O\\
Post 3: ``Example text'' $\rightarrow$ IS\\
Post 4: ``Example text'' $\rightarrow$ IE
\end{quote}

\textbf{3-shot:} Model receives three labeled timeline examples before the input.

\begin{quote}\ttfamily
Example 1:\\
Post 1: ``Example text'' $\rightarrow$ O\\
Post 2: ``Example text'' $\rightarrow$ IS

Example 2:\\
Post 1: ``Example text'' $\rightarrow$ O\\
Post 2: ``Example text'' $\rightarrow$ O\\
Post 3: ``Example text'' $\rightarrow$ IE

Example 3:\\
Post 1: ``Example text'' $\rightarrow$ O\\
Post 2: ``Example text'' $\rightarrow$ IS
\end{quote}

\medskip
\textbf{Input Format}

\begin{quote}\ttfamily
Timeline:\\
Post 1: ``...'' $\rightarrow$ O\\
Post 2: ``...'' $\rightarrow$ O\\
Post 3: ``...'' $\rightarrow$ ?
\end{quote}

\medskip
\textbf{Output Format}

\begin{quote}\ttfamily
O \quad | \quad IS \quad | \quad IE
\end{quote}

Return exactly one label corresponding to the final post in the timeline.

\end{tcolorbox}

\caption{Prompt templates used for TalkLife timeline mood change detection under
0-shot, 1-shot, and 3-shot prompting settings.}
\label{fig:talklife_prompts}

\end{figure*}

\begin{figure*}[t]
\centering
\begin{tcolorbox}[
width=\textwidth,
title={Prompt Templates for AnnoMI},
colback=gray!10,
colframe=black!60,
coltitle=white,
colbacktitle=black!65,
fonttitle=\bfseries,
boxrule=0.8pt,
arc=6pt,
left=10pt,
right=10pt,
top=8pt,
bottom=8pt
]

\scriptsize

\textbf{Task Description}

You are an expert Motivational Interviewing (MI) coder.
Given a \textbf{timeline of client utterances} within the same topic,
classify each utterance according to its motivational orientation.
Labels should be determined relative to the preceding client utterances
in the timeline. Therapist utterances are not labeled.

\medskip
\textbf{Label Definitions}

\begin{itemize}
\item \textsc{CHANGE}: \textbf{Change talk}.  
Language indicating movement toward behavior change.

\item \textsc{SUSTAIN}: \textbf{Sustain talk}.  
Language favoring the status quo or resisting change.

\item \textsc{NEUTRAL}: \textbf{Neutral talk}.  
Language that is neither change nor sustain talk, or is ambiguous.
\end{itemize}

\textbf{Annotation Guidelines}

\begin{itemize}
\item Utterances appear in chronological order.
\item The first utterance may be labeled \textsc{NEUTRAL} if no context exists.
\item Consider the evolving motivational stance across the timeline.
\item If mixed or ambiguous, assign \textsc{NEUTRAL}.
\item Return exactly one label for the final utterance.
\end{itemize}

\medskip
\textbf{Prompt Variants}

\textbf{0-shot:} Model receives only the task description and input timeline.

\textbf{1-shot:} Model receives one labeled timeline example before the input.

\begin{quote}\ttfamily
Example Timeline:\\
Utterance 1: ``Example text'' $\rightarrow$ NEUTRAL\\
Utterance 2: ``Example text'' $\rightarrow$ SUSTAIN\\
Utterance 3: ``Example text'' $\rightarrow$ CHANGE
\end{quote}

\textbf{3-shot:} Model receives three labeled timeline examples before the input.

\begin{quote}\ttfamily
Example 1:\\
Utterance 1: ``Example text'' $\rightarrow$ NEUTRAL\\
Utterance 2: ``Example text'' $\rightarrow$ CHANGE

Example 2:\\
Utterance 1: ``Example text'' $\rightarrow$ SUSTAIN\\
Utterance 2: ``Example text'' $\rightarrow$ SUSTAIN\\
Utterance 3: ``Example text'' $\rightarrow$ CHANGE

Example 3:\\
Utterance 1: ``Example text'' $\rightarrow$ NEUTRAL\\
Utterance 2: ``Example text'' $\rightarrow$ SUSTAIN
\end{quote}

\medskip
\textbf{Input Format}

\begin{quote}\ttfamily
Timeline:\\
Utterance 1: ``...'' $\rightarrow$ NEUTRAL\\
Utterance 2: ``...'' $\rightarrow$ SUSTAIN\\
Utterance 3: ``...'' $\rightarrow$ ?
\end{quote}

\medskip
\textbf{Output Format}

\begin{quote}\ttfamily
CHANGE \quad | \quad SUSTAIN \quad | \quad NEUTRAL
\end{quote}

Return exactly one label corresponding to the final utterance.

\end{tcolorbox}

\caption{Prompt templates used for AnnoMI client-talk classification under
0-shot, 1-shot, and 3-shot prompting settings using topic-level timelines.}
\label{fig:AnnoMI_prompts}

\end{figure*}

\begin{figure*}[t]
\centering
\begin{tcolorbox}[
width=\textwidth,
title={Prompt Templates for LRS },
colback=gray!8,
colframe=black!60,
coltitle=white,
colbacktitle=black!65,
fonttitle=\bfseries,
boxrule=0.8pt,
arc=6pt,
left=10pt,
right=10pt,
top=8pt,
bottom=8pt
]
\scriptsize

\textbf{Task Description}

You are an expert annotator trained to detect \textbf{stance changes} in longitudinal
social media timelines. A \textbf{stance} is the author's position toward a specific
claim, not tone, emotion, or level of detail.

Given a sequence of posts written by the same author in chronological order,
determine whether the stance expressed in each post switches relative to the
preceding timeline context about the same claim.

\medskip
\textbf{Label Definitions}

\begin{itemize}
\item \textsc{1}: \textbf{Stance switch}.  
The current post explicitly reverses or contradicts an earlier stance.

\item \textsc{0}: \textbf{No switch}.  
The current post remains consistent with the prior stance.
\end{itemize}

\textbf{Annotation Guidelines}

\begin{itemize}
\item Posts appear in chronological order.
\item The first post in a timeline is labeled \textsc{0}.
\item Only explicit reversals count as switches.
\item Added details, questions, tone changes, or related topics do not count as switches.

\end{itemize}

\medskip
\textbf{Prompt Variants}

\textbf{0-shot:} Model receives only the task description and input timeline.

\textbf{1-shot:} Model receives one labeled timeline example before the input.

\begin{quote}\ttfamily
Example Timeline:\\
Post 1: ``Example text'' $\rightarrow$ 0\\
Post 2: ``Example text'' $\rightarrow$ 0\\
Post 3: ``Example text'' $\rightarrow$ 1
\end{quote}

\textbf{3-shot:} Model receives three labeled timeline examples before the input.

\begin{quote}\ttfamily
Example 1:\\
Post 1: ``Example text'' $\rightarrow$ 0\\
Post 2: ``Example text'' $\rightarrow$ 1

Example 2:\\
Post 1: ``Example text'' $\rightarrow$ 0\\
Post 2: ``Example text'' $\rightarrow$ 0\\
Post 3: ``Example text'' $\rightarrow$ 1

Example 3:\\
Post 1: ``Example text'' $\rightarrow$ 0\\
Post 2: ``Example text'' $\rightarrow$ 0
\end{quote}

\medskip
\textbf{Input Format}

\begin{quote}\ttfamily
Timeline:\\
Post 1: ``...'' $\rightarrow$ 0\\
Post 2: ``...'' $\rightarrow$ 0\\
Post 3: ``...'' $\rightarrow$ ?
\end{quote}

\medskip
\textbf{Output Format}

\begin{quote}\ttfamily
0 \quad | \quad 1
\end{quote}

Return exactly one label corresponding to the final post in the timeline.

\end{tcolorbox}
\caption{Prompt templates used for LRS stance switch classification under 0-shot, 1-shot, and 3-shot prompting settings.}
\label{fig:lrs_prompts}
\end{figure*}

\begin{figure*}[t]
\centering
\begin{tcolorbox}[
width=\textwidth,
title={Prompt Templates for Reddit},
colback=gray!8,
colframe=black!60,
coltitle=white,
colbacktitle=black!65,
fonttitle=\bfseries,
boxrule=0.8pt,
arc=6pt,
left=10pt,
right=10pt,
top=8pt,
bottom=8pt
]
\scriptsize

\textbf{Task Description}

You are an expert annotator of mood dynamics in longitudinal mental health timelines.
Given a sequence of posts written by the same user in chronological order, determine
whether each post reflects a change in mood relative to the preceding timeline context.

\medskip
\textbf{Label Definitions}

\begin{itemize}
\item \textsc{O}: \textbf{No observable mood change}.  
The emotional state remains similar to previous timeline entries.

\item \textsc{S}: \textbf{Mood switch}.  
A clear shift in emotional valence or overall affect occurs.

\item \textsc{E}: \textbf{Mood escalation}.  
The emotional direction remains consistent but intensity, urgency,
or distress increases meaningfully.
\end{itemize}

\textbf{Annotation Guidelines}

\begin{itemize}
\item Posts appear in chronological order.
\item The first post in a timeline is labeled \textsc{O}.
\item Topic changes alone do not constitute a mood change.
\item Stronger wording alone does not imply escalation unless severity increases.
\end{itemize}

\medskip
\textbf{Prompt Variants}

\textbf{0-shot:} Model receives only the task description and input timeline.

\textbf{1-shot:} Model receives one labeled timeline example before the input.

\begin{quote}\ttfamily
Example Timeline:\\
Post 1: ``Example text'' $\rightarrow$ O\\
Post 2: ``Example text'' $\rightarrow$ O\\
Post 3: ``Example text'' $\rightarrow$ S\\
Post 4: ``Example text'' $\rightarrow$ E
\end{quote}

\textbf{3-shot:} Model receives three labeled timeline examples before the input.

\begin{quote}\ttfamily
Example 1:\\
Post 1: ``Example text'' $\rightarrow$ O\\
Post 2: ``Example text'' $\rightarrow$ S

Example 2:\\
Post 1: ``Example text'' $\rightarrow$ O\\
Post 2: ``Example text'' $\rightarrow$ O\\
Post 3: ``Example text'' $\rightarrow$ E

Example 3:\\
Post 1: ``Example text'' $\rightarrow$ O\\
Post 2: ``Example text'' $\rightarrow$ S
\end{quote}

\medskip
\textbf{Input Format}

\begin{quote}\ttfamily
Timeline:\\
Post 1: ``...'' $\rightarrow$ O\\
Post 2: ``...'' $\rightarrow$ O\\
Post 3: ``...'' $\rightarrow$ ?
\end{quote}

\medskip
\textbf{Output Format}

\begin{quote}\ttfamily
O \quad | \quad S \quad | \quad E
\end{quote}

Return exactly one label corresponding to the final post in the timeline.

\end{tcolorbox}
\caption{Prompt templates used for Reddit timeline mood change classification under
0-shot, 1-shot, and 3-shot prompting settings.}
\label{fig:reddit_prompts}
\end{figure*}

\begin{figure*}[t]
\centering
\begin{tcolorbox}[
width=\textwidth,
title={Prompt Templates for CMV},
colback=gray!8,
colframe=black!60,
coltitle=white,
colbacktitle=black!65,
fonttitle=\bfseries,
boxrule=0.8pt,
arc=6pt,
left=10pt,
right=10pt,
top=8pt,
bottom=8pt
]
\scriptsize

\textbf{Task Description}

You are an expert annotator of persuasion and belief change in longitudinal
Change-My-View (CMV) conversation timelines.
Given a sequence of conversations written by the same user in chronological order,
determine whether the \textbf{final conversation} results in a change in the
original poster's view relative to the earlier timeline context.

\medskip
\textbf{Label Definitions}

\begin{itemize}
\item \textsc{0}: \textbf{No view change}.  
The conversation does not provide evidence that the OP's stance changed.

\item \textsc{1}: \textbf{View changed}.  
The conversation contains evidence that the OP reconsidered or changed their view.
\end{itemize}

\textbf{Annotation Guidelines}

\begin{itemize}
\item Conversations appear in chronological order.
\item Prior conversations provide background context about the user's views.
\item The prediction concerns \textbf{only the final conversation in the timeline}.
\item Evidence such as concessions, acknowledgments, or explicit reconsideration supports \textsc{1}.
\item Disagreement, rebuttal, or insufficient persuasion supports \textsc{0}.
\end{itemize}

\medskip
\textbf{Prompt Variants}

\textbf{0-shot:} Model receives only the task description and input timeline.

\textbf{1-shot:} Model receives one labeled timeline example before the input.

\begin{quote}\ttfamily
Example Timeline:\\
Conversation 1: ``Example CMV discussion'' $\rightarrow$ 0\\
Conversation 2: ``Example CMV discussion'' $\rightarrow$ 0\\
Conversation 3: ``Example CMV discussion'' $\rightarrow$ 1
\end{quote}

\textbf{3-shot:} Model receives three labeled timeline examples before the input.

\begin{quote}\ttfamily
Example 1:\\
Conversation 1: ``Example discussion'' $\rightarrow$ 0\\
Conversation 2: ``Example discussion'' $\rightarrow$ 1

Example 2:\\
Conversation 1: ``Example discussion'' $\rightarrow$ 0\\
Conversation 2: ``Example discussion'' $\rightarrow$ 0\\
Conversation 3: ``Example discussion'' $\rightarrow$ 1

Example 3:\\
Conversation 1: ``Example discussion'' $\rightarrow$ 0\\
Conversation 2: ``Example discussion'' $\rightarrow$ 1
\end{quote}

\medskip
\textbf{Input Format}

\begin{quote}\ttfamily
Timeline:\\
Conversation 1: ``...'' $\rightarrow$ 0\\
Conversation 2: ``...'' $\rightarrow$ 0\\
Conversation 3: ``...'' $\rightarrow$ ?
\end{quote}

\medskip
\textbf{Output Format}

\begin{quote}\ttfamily
0 \quad | \quad 1
\end{quote}

Return exactly one label corresponding to the final conversation in the timeline.

\end{tcolorbox}
\caption{Prompt templates used for CMV conversation timeline view-change classification under
0-shot, 1-shot, and 3-shot prompting settings.}
\label{fig:cmv_prompts}
\end{figure*}

% \nocite{langley00}
% =========================
% Optional: Ethics / Broader Impact (if desired)
% =========================
% \section{Broader Impact}
% Discuss intended use, misuse risks, privacy, and clinical non-substitution.

%%%%%%%%%%%%%%%%%%%%%%%%%%%%%%%%%%%%%%%%%%%%%%%%%%%%%%%%%%%%%%%%%%%%%%%%%%%%%%%
%%%%%%%%%%%%%%%%%%%%%%%%%%%%%%%%%%%%%%%%%%%%%%%%%%%%%%%%%%%%%%%%%%%%%%%%%%%%%%%
% APPENDIX
%%%%%%%%%%%%%%%%%%%%%%%%%%%%%%%%%%%%%%%%%%%%%%%%%%%%%%%%%%%%%%%%%%%%%%%%%%%%%%%
%%%%%%%%%%%%%%%%%%%%%%%%%%%%%%%%%%%%%%%%%%%%%%%%%%%%%%%%%%%%%%%%%%%%%%%%%%%%%%%
\newpage
\appendix
\onecolumn

\end{document}